\documentclass{article}
\pdfoutput=1 

\usepackage{microtype}
\usepackage{graphicx}
\usepackage{enumitem}
\usepackage{color, soul}

\usepackage{hyperref}

\usepackage[accepted]{icml2019}

\usepackage{xspace}
\newcommand*{\eg}{e.g.\@\xspace}

\usepackage{bm}
\usepackage{amsmath}
\usepackage{enumitem}
\usepackage{placeins}
\usepackage{subcaption}
\icmltitlerunning{Explainability Techniques for Graph Convolutional Networks}

\begin{document}

\twocolumn[
\icmltitle{Explainability Techniques for Graph Convolutional Networks}


\icmlsetsymbol{equal}{*}

\begin{icmlauthorlist}
\icmlauthor{Federico Baldassarre}{kth}
\icmlauthor{Hossein Azizpour}{kth}
\end{icmlauthorlist}

\icmlaffiliation{kth}{Division of Robotics, Perception, and Learning, School of Electrical Engineering and Computer Science, KTH (Royal Institute of Technology), Stockholm, Sweden}

\icmlcorrespondingauthor{Federico Baldassarre}{fedbal@kth.se}
\icmlcorrespondingauthor{Hossein Azizpour}{azizpour@kth.se}

\icmlkeywords{Machine Learning, Graph Networks, Explainability, ICML}

\vskip 0.3in
]



\printAffiliationsAndNotice{}  

\begin{abstract}
Graph Networks are used to make decisions in potentially complex scenarios but it is usually not obvious how or why they made them. In this work, we study the explainability of Graph Network decisions using two main classes of techniques, gradient-based and decomposition-based, on a toy dataset and a chemistry task. Our study sets the ground for future development as well as application to real-world problems.
\end{abstract}


\vspace{-.8cm}
\section{Introduction}
\label{sec:intro_short}

Many concepts from chemistry, life science, and physics are naturally represented in the graph domain. Most Machine Learning (ML) methods, however, were devised to operate on Euclidean space. One of the most successful ML techniques, Deep Learning, has been recently generalized to operate on a graph domain. These Graph Networks (GN) have achieved remarkable performances in various applications thanks to their consistency to the native data representation, with examples from biochemistry \cite{duvenaud2015convolutional, kearnes2016molecular, fout2017protein, zitnik2018modeling}, physics \cite{battaglia2016interaction, chang2017compositional, gilmer2017neural, watters2017visual, sanchez2018graph}, visual recognition \cite{qi20173d, qi2018eccv, narasimhan2018out}, and natural language processing \cite{bastings2017emnlp, beck2018acl}. 

 ML algorithms become increasingly trustworthy to humans when the basis for their decisions can be explained in human terms. Interpretability is also useful for diagnosing biases, designing datasets and gaining insight on governing laws.
While several methods have been developed for standard deep networks, there is a lack of study for applicability of such methods on GNs; that is the focus of this work.

\begin{figure}[t!]
    \centering
    \begin{subfigure}{.5\linewidth}
    \centering
    \includegraphics[width=\linewidth]{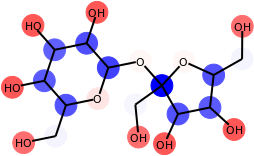}
    \end{subfigure}
    \vspace{-0.2cm}
    \caption{\textbf{Explanation for the solubility of sucrose:} {\small the prediction is decomposed into positive (red) and negative (blue) contributions attributed to the atoms using Layer-wise Relevance Propagation.}}
    \label{fig:sucrose}
\end{figure}

In this work, we assume the general form of GNs as defined in \cite{battaglia2018arxiv}. Regarding explanation algorithms, we consider two main classes: a) gradient-based such as Sensitivity Analysis \cite{baehrens2010jmlr, simonyan2014iclrw} and Guided Backpropagation \cite{springenberg2015ICLRW}, b) decomposition-based such as Layer-wise Relevance Propagation \cite{bach2015plosone} and Taylor decomposition \cite{montavon2017pr}. We base the discussions on a toy dataset and a chemistry task. We hope this work will set the ground for this important topic and suggest future directions and applications.

The contributions of this work can be summarized as:
\begin{enumerate}[noitemsep,nolistsep,topsep=-10pt]
    \item to the best of our knowledge, this is the first work to focus on explainability techniques for GN
    \item we highlight and identify the challenges and future directions for explaining GN
    \item we compare two main classes of explainability methods on graph-based tasks
predictions
\end{enumerate}
\vspace{7pt}
Our PyTorch~\cite{paszke2017automatic} implementation of GNs equipped with different explanation algorithms is available at {\href{https://github.com/baldassarreFe/graph-network-explainability}{github.com/baldassarreFe/graph-network-explainability}}.

\section{Related works}
\label{sec:related_short}
This work is closely related to Graph Networks and explanation techniques developed for standard networks.

\textbf{Graph Networks} Graphs can be embedded in Euclidean spaces using Neural Networks \cite{perozzi2014sigkdd, hamilton2017nips, kipf2016arxiv}. This representation preserves the relational structure of the graph while enjoying the properties of Euclidean space. The embedding can then be processed further, \eg with more interpretable linear models \cite{duvenaud2015convolutional}. On the other hand, DL algorithms that operate end-to-end in the graph domain can leverage the structure to predict at vertex~\cite{hu2018cvpr}, edge~\cite{qi2018eccv}, or global~\cite{zitnik2018modeling} levels.\\
Several variants of GN have been proposed starting from the early work of~\cite{scarselli2009nn}. It was further extended, by gating~\cite{li2016iclr}, convolutions in spectral~\cite{bruna2014iclr} and spatial~\cite{kipf2017semisupervised} domains, skip connections~\cite{rahimi2018acl} and recently attention~\cite{velickovic2018iclr}.
In this work, we focus on the GN as described in \cite{battaglia2018arxiv}, which tries to generalize all variants while remaining conceptually simple.

\textbf{Model interpretation and explanation of predictions} ML methods are ubiquitous in several industries, with deep networks achieving impressive performances. Where humans are involved, the expectation of high levels of safety, reliability, and fairness arises. Thanks to this demand, several techniques have been developed for increasing the transparency of the most successful models. 
At high level these techniques can be divided into those that attempt to \textit{interpret the model as a whole}~\cite{simonyan2014iclrw, nguyen2016arxiv} and those that try to \textit{explain individual predictions} made by a model~\cite{sung1998esa, bach2015plosone}\footnote{For insight on such grouping refer to~\cite{gilpin2018dsaa}}. \\
This paper focuses on the latter, for which several techniques have been developed on standard deep networks. These include works that attempt at variation-based analysis~\cite{baehrens2010jmlr, simonyan2014iclrw, springenberg2015ICLRW, ribeiro2016sigkdd, smilkov2017arxiv, bordes2018midl} and output decomposition~\cite{montavon2017pr, shrikumar2017icml, kindermans2018iclr}. \cite{sundararajan2017icml} combines the two principles and invert the decisions~\cite{zeiler2014eccv, mahendran2015cvpr, carlsson2017iclrw}. Furthermore, \cite{dhurandhar2018nips} uses contrastive explanations, \cite{zhang2018nips} identifies minimal changes in the input to get the desired prediction. In this work, we evaluate variation- and decomposition-based techniques in the context of GNs.

\textbf{Explanation for GNs} To the best of our knowledge, exploring explanation techniques for GNs has not been the focus of any prior work. In \cite{duvenaud2015convolutional}, GNs are used to learn molecular fingerprints and predict their chemical properties. Chemically grounded insights are then extracted from the model by heuristic inspection of their results\footnote{as described in the authors' rebuttal: \cite{duvenaud2015convolutionalReviewURL}}. In our experiments, we show that GNs trained end-to-end on those problems achieve similar performance while making it possible to \textit{explain individual predictions}.

\section{Method}
\subsection{Graph Network}
GN as described in \cite{battaglia2018arxiv} use a message-passing algorithm that aggregates local information similar to convolutions in CNNs. Graphs can contain features on edges $E = \{\bm{e}_k\}$, nodes $V = \{\bm{v}_i\}$ and graph-level $\bm{u}$. At every layer of computation, the graph is updated using three update functions $\phi$ and three aggregation functions $\rho$:
\begin{align}
  \begin{split}
    \bm{e}'_k &= \phi^e\left(\bm{e}_k, \bm{v}_{r_k}, \bm{v}_{s_k}, \bm{u} \right) \\
    \bm{v}'_i &= \phi^v\left(\bm{\bar{e}}'_i, \bm{v}_i, \bm{u}\right) \\
    \bm{u}' &= \phi^u\left(\bm{\bar{e}}', \bm{\bar{v}}', \bm{u}\right)
  \end{split}
  \begin{split}
    \bm{\bar{e}}'_i &= \rho^{e \rightarrow v}\left(E'_i\right) \\
    \bm{\bar{e}}' &= \rho^{e \rightarrow u}\left(E'\right) \\
    \bm{\bar{v}}' &= \rho^{v \rightarrow u}\left(V'\right)   
  \end{split}
  \label{eq:gn-functions}
\end{align}
where $r_k$ and $s_k$ represent the sender and receiver nodes of the $k$-th edge, and the sets $E'_i$, $E'$, $V'$ represent the edges incident to node $i$, all edges updated by $\phi^e$ and all nodes updated by $\phi^n$. Each processing layer leaves the structure of the graph unchanged, updating only the features of the graph and not its topology. 
The mapping $f: (E,V,\bm{u}) \to y$, can represent a single quantity of interest (\eg the solubility of a molecule) or a graph with individual predictions for nodes and edges. 
In this work, all $\phi$ are linear transformations followed by ReLU activations, and all $\rho$ are sum/mean/max pooling operations.
\vspace{-5pt}
\subsection{Explainability}
\paragraph{Sensitivity Analysis (SA)} produces local explanations for the prediction of a differentiable function $f$ using the squared norm of its gradient w.r.t. the inputs $\bm{x}$ \cite{gevrey2003review}: $\bm{S}(\bm{x}) \propto \|\nabla_\mathbf{x} f\|^2$. The saliency map $\bm{S}$ produced with this method describes the extent to which variations in the input would produce a change in the output.
\vspace{-10pt}
\paragraph{Guided Backpropagation (GBP)} also constructs a saliency map using gradients \cite{springenberg2015ICLRW}. Different from SA, negative gradients are clipped during backpropagation, which concentrates the explanation on the features that have an excitatory effect on the output.
\vspace{-10pt}
\paragraph{Layer-wise Relevance Propagation (LRP)} produces relevance maps by decomposing the output signal of every transformation into a combination of its inputs. For certain configuration of weights and activations, LRP can be interpreted as a repeated Taylor decomposition \cite{montavon2017pr} that preserves the amount of total relevance $R$ across layers: $\sum R^{(x)} = \dots = \sum R^{(\ell)} = \dots = f(\bm{x})$. \cite{bach2015plosone} introduces two rules for LRP, the $\alpha\beta$-rule and the $\epsilon$-stabilized rule, both discussed in Appendix \ref{sec:app-explanation-lrp}. We chose the latter for its robustness and simplicity.\\
Different to the previous two methods, LRP identifies which features of the input contribute the most to the final prediction, rather than focusing on its variation. Furthermore, it is capable of handling positive and negative relevance, allowing for a deeper analysis of the contributing factors.
\vspace{-10pt}
\paragraph{Autograd-based implementation}
All three methods rely on a backward pass through the network to propagate gradients/relevance from the output and accumulating it to the input.
Since the computational graph of a GN can become complex and non-sequential, we take advantage of PyTorch's capability to track operations and implement these algorithms on top of its \textit{autograd} module.

\label{sec:method}

\section{Experiments}
\label{sec:experiments}
To evaluate different explainability methods on GNs, we consider a toy graph problem and a chemistry regression task. Task-specific comments can be found in this section, followed by a more general discussion in Section~\ref{sec:discussions}.

\subsection{Infection}

In this toy problem, the input graph represents a group of individuals who are either \textit{sick} or \textit{healthy}, as well as \textit{immune} to a certain disease. Between people are directed edges, representing the relationships they have, characterized as \textit{virtual} or not. The disease spreads according to a simple rule: a sick node infects the neighbors to which it is connected through a non-virtual edge, unless the target node is immune. 
The objective is to predict the state of every node of the graph after one step of the spread, and then evaluate the correctness of the explanations produced with Sensitivity Analysis, Guided Backpropagation and Layer-wise Relevance Propagation against the logical infection dynamics.
Details about the dataset, the network and the training procedure are in Appendix \ref{sec:app-gn-infection}.

\begin{figure}[h!]
\centering
\begin{subfigure}{0.23\linewidth}
    \centering
    \includegraphics[width=\linewidth]{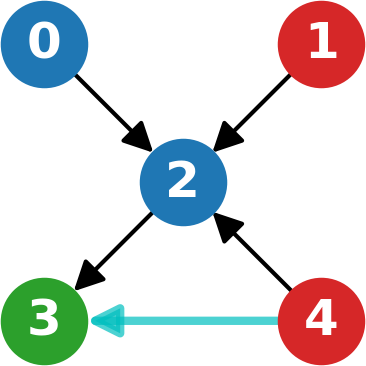}
    \vspace{-10pt}    
    \caption*{\small Input}
\end{subfigure}\hfill\vrule\hfill
\begin{subfigure}{0.23\linewidth}
    \centering
    \includegraphics[width=\linewidth]{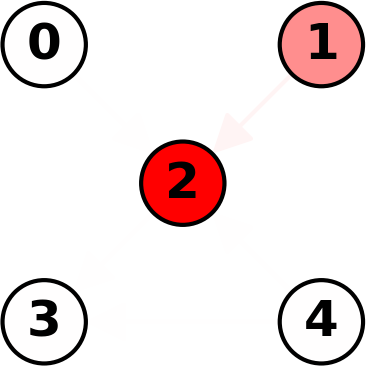}
    \vspace{-10pt}    
    \caption*{\small SA}
\end{subfigure}\hfill
\begin{subfigure}{0.23\linewidth}
    \centering
    \includegraphics[width=\linewidth]{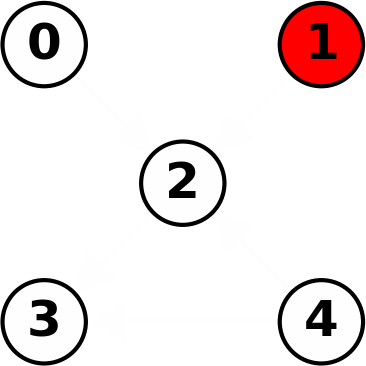}
    \vspace{-10pt}
    \caption*{\small GBP}
\end{subfigure}\hfill
\begin{subfigure}{0.23\linewidth}
    \centering
    \includegraphics[width=\linewidth]{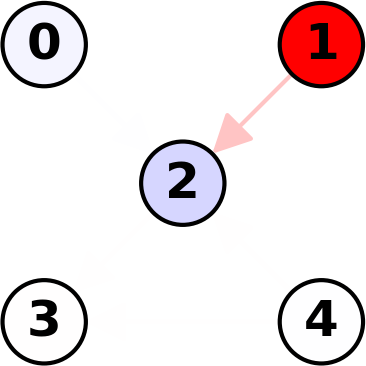}
    \vspace{-10pt}
    \caption*{\small LRP}
\end{subfigure}%
\caption{\textbf{Explaining why node 2 becomes infected.}\\ {\small SA places high relevance on the node itself (if 2 was more sick at the beginning, it would be more infected at the end). GBP correctly identifies node 1 as a source of infection, but very small importance is given to the edge. LRP decomposes the prediction into a negative contribution (blue, node 2 is not sick), and two positive contributions (red, node 1 is sick and $1\to 2$ is not virtual). Node 4 is ignored due to max pooling.}}
\label{fig:small-graph-1}
\end{figure}

When tasked to explain the prediction for a single node, all three techniques identify the relevant nodes/edges of the input graph (Fig.~\ref{fig:small-graph-1}).
We note however that the explanations produced by variation-based methods tend to diverge from how a human would intuitively describe the process in terms of cause and effect, while LRP results are more natural.
Appendix~\ref{sec:app-infection-results} provides a detailed case-based visualization of the explanations, down to the individual features.

\subsection{Solubility}

We train a GN to predict the aqueous solubility of organic compounds from their molecular graph as in \cite{duvenaud2015convolutional}. Our multi-layer GN matches their performances while remaining simple. Details about the dataset, the network and the training procedure are in Appendix \ref{sec:app-gn-infection}.

\begin{figure}[t!]
\centering
\begin{subfigure}{0.50\linewidth}
    \centering
    \includegraphics[width=.5\linewidth]{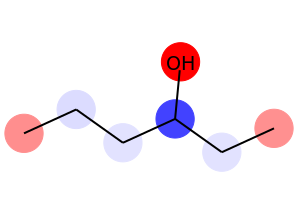}
    \vspace{-5pt}
    \caption*{\small 3-Hexanol (-0.80) \vspace{10pt}}
\end{subfigure}%
\begin{subfigure}{0.50\linewidth}
    \centering
    \includegraphics[width=.5\linewidth]{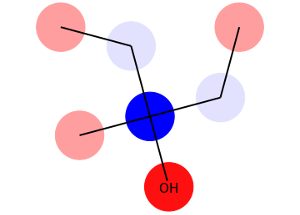}
    \vspace{-5pt}
    \caption*{\small 3-Methyl-3-pentanol (-0.36) \vspace{10pt}}
\end{subfigure}\newline%
\begin{subfigure}{0.50\linewidth}
    \centering
    \includegraphics[width=.7\linewidth]{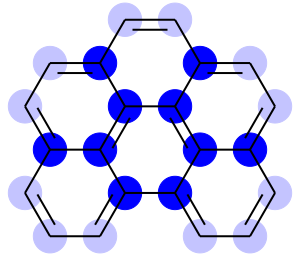}
    \vspace{-4pt}
    \caption*{\small Benzo[ghi]perylene (-9.02) \vspace{2pt}}
\end{subfigure}%
\begin{subfigure}{0.50\linewidth}
    \centering
    \includegraphics[width=.7\linewidth]{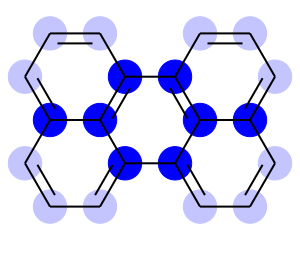}
    \vspace{-4pt}
    \caption*{\small Perylene (-8.80) \vspace{2pt}}
\end{subfigure}%
\caption{\textbf{Predicted solubility (log mol/L) and its explanation produced with LRP.} {\small Positive relevance (red) are on R-OH groups, indicating their positive contribution to the predicted value. Negative relevance (blue) can be found on central carbons and non-polar aromatic rings, indicating they advocate towards lower solubility values. See the appendix for a breakdown of the explanation onto the individual features.}}
\label{fig:molecules}
\end{figure}

When explaining the predictions of the network, LRP attributes positive and negative relevance to features that are known to correlate with solubility, such as the presence of R-OH groups on the outside of the molecule and features that typically indicate low solubility such as repeated non-polar aromatic rings (Fig. \ref{fig:molecules}). Similar observations are made in \cite{duvenaud2015convolutional}, although by manual inspection of high-scoring predictions.

Note that LRP was originally introduced to explain classification predictions, but here it is adopted for a regression task. A discussion on how to interpret these explanations can be found in section~\ref{sec:app-solubility-results} of the appendix.

\section{Discussion}
\label{sec:discussions}




Recent methods for explainability have been developed within image or text domains~\cite{bach2015plosone,springenberg2015ICLRW,ribeiro2016sigkdd}. With the experiments presented in this work, we intend to highlight some key differences of the graph domain that require special consideration to produce meaningful explanations.

\subsection{The role of connections}
\begin{figure}[h!]
\centering
\begin{subfigure}{.9\linewidth}
    \centering
    \includegraphics[width=\linewidth]{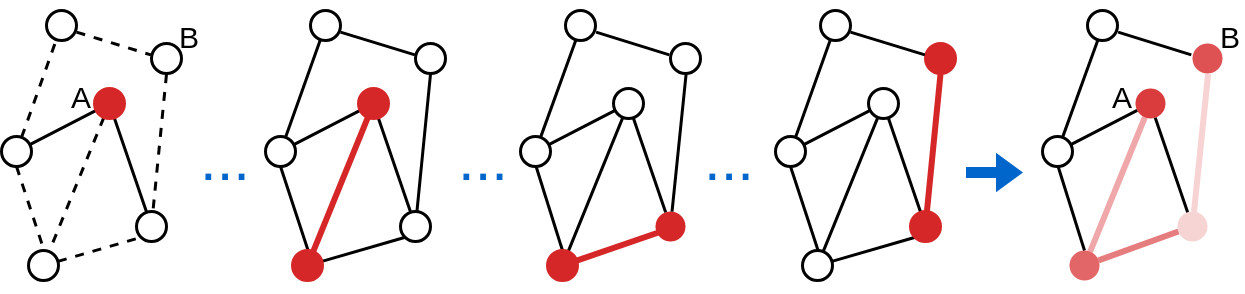}
\end{subfigure}
\vspace{-5pt}
\caption{\textbf{Node A is responsible for a prediction on B.} {\small Even if the input graph does not have features on the edges (represented as dashed lines), the relevant path $A \to B$ is identified by aggregating the relevance at multiple steps of graph convolution.}}
\label{fig:role-of-connections}
\end{figure}

Images can be seen as graphs with a regular grid topology and whose features are attributed only to nodes. In this context, an explanation can take the form of a heatmap over the image, highlighting relevant pixels and, implicitly, their local connections. 
For graphs with irregular connectivity, edges acquire a more preeminent role that can be missed when using image-based explanation techniques. For example, in a graph where edge features are not present or are all identical (not informative), neither gradients nor relevance would be propagated back to these connections, even though the presence itself of a connection between two nodes is a source of information that should be taken into account for explanations.
We propose to take advantage of the structure-preserving property of graph convolution and aggregate explanations at multiple steps of message-passing, arguing that the importance of connections should emerge from the intermediate steps  (Fig.~\ref{fig:role-of-connections}).

\subsection{Pooling}
\textbf{Architectural choice} In standard NN, pooling operations are commonly used to aggregate features. In message-passing GNs, pooling is used to aggregate edge and node features at a local and global level, while not modifying the topology of the network (Eq.~\ref{eq:gn-functions}). 
The choice of pooling function in GN is closely related to the learning problem, \eg sum pooling is best for counting at a global level, while max pooling can be used for identifying local properties.

\begin{figure}[t!]
\centering
\begin{subfigure}{0.23\linewidth}
    \centering
    \includegraphics[width=\linewidth]{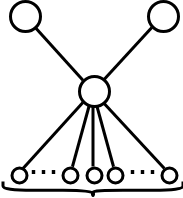}
    \caption*{\small Input}
\end{subfigure}\hfill\vrule\hfill
\begin{subfigure}{0.23\linewidth}
    \centering
    \includegraphics[width=\linewidth]{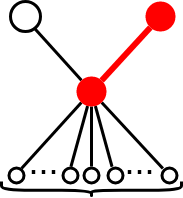}
    \caption*{\small Na{\"i}ve}
\end{subfigure}\hfill
\begin{subfigure}{0.23\linewidth}
    \centering
    \includegraphics[width=\linewidth]{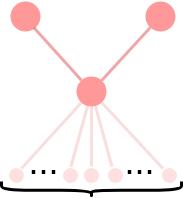}
    \caption*{\small $L_p$-norm}
\end{subfigure}\hfill
\begin{subfigure}{0.23\linewidth}
    \centering
    \includegraphics[width=\linewidth]{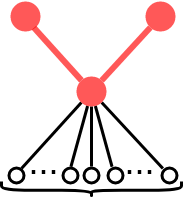}
    \caption*{\small Search-based}
\end{subfigure}%

\caption{\textbf{Propagation rules for max pooling.} {\small The two top nodes are important, the $N$ bottom ones are not. Relevance can be na{\"i}vely propagated from the central node to only one of the top nodes. Approximating max pooling as a $L_p$-norm would result in a more complete explanation, but for large $N$ relevance could disperse to unimportant nodes. A search-based method identifies and propagates relevance only to the relevant nodes.}}
\label{fig:max-vs-sum}
\end{figure}

\textbf{Explanations} The choice of aggregation also influences the explanations obtained for a prediction. Sum and mean pooling propagate gradients/relevance to all their inputs, possibly identifying all sources of signal. Max pooling, instead, considers only one of its inputs and disregards others, regardless of their magnitude, which can lead to incomplete explanations (\eg multiple neighboring sick nodes could explain an infection). To counter this, LRP proposes to approximate max pooling with $L_p$-norm during relevance propagation, but this approach can over-disperse relevance to unimportant inputs. We suggest that the backward pass through max pooling should be approached as a search that only redistributes relevance to those inputs that result in a similar prediction if chosen as maxima (Fig.~\ref{fig:max-vs-sum}).

\subsection{Heterogeneous Graph Features}
Images are usually represented as a matrix of continuous RGB pixel values, while graphs are often employed for domains that require a mixed encoding of continuous, binary and categorical features that are \textit{semantically meaningful} \cite{fout2017protein,kearnes2016molecular,sanchez2018graph}. Thus, rather than aggregating the explanation at the node/edge level, it can be of higher interest to evaluate the importance of individual features. For this reason, visualizations based on graph heatmaps might be insufficient. We suggest a more detailed visualization in Appendix~\ref{sec:app-more-results}.

\subsection{Perturbation-based evaluation} 
Images and graphs can be considered as points in very high-dimensional spaces, belonging to complex and structured manifolds \cite{tenenbaum2000global}. The commonly used representation of an image introduces an elevated degree of redundancy, so that changing the value of a single pixel minimally affects the content and meaning associated with the image.
Under this observation, an explanation can be quantitatively evaluated by progressively "graying out" pixels in order of importance and measuring how it affects the prediction~\cite{bach2015plosone}.
On the other hand, graph representations tend to be less redundant and the structure of the graph is a constituent part of its identity, therefore small alterations of nodes/edges can drastically alter the meaning of the graph. In our chemistry problem, for example, replacing an atom or bond would fundamentally change a molecule or invalidate it. As a viable strategy, one could rely on domain-specific knowledge to perform such changes while remaining semantically close to the original. Alternatively, one can learn a bijective grounding of graphs onto a manifold with meaningful neighborhood to conduct such an evaluation. In Appendix~\ref{sec:app-solubility-results} we present a hand-crafted example of progressively eliminating atoms from a molecule in order of importance using domain knowledge.

\section{Conclusion}
\label{sec:conclusion}
As a expository paper, in this work, we introduced and focused on an important problem with impactful applications: we analyzed the major existing explanation techniques in the context of Graph Networks. We further conducted a case-based analysis on two simple but complementary tasks as well as some important high-level discussions on design choices when explaining a GNs decision. Finally, we provided an implementation of five different explanation techniques using PyTorch autograd which can be readily used for any definition of GN. In tandem with the high-level technical novelty, we hope these contributions open up fruitful discussions at the workshop and pave the road for future development of specific techniques for GN explanation for real-world applications.

\textbf{Acknowledgement.} Federico Baldassarre is partially funded by Swedish Research Council project 2017-04609

\bibliography{ICML19_LRG}
\bibliographystyle{icml2019}

\clearpage
\appendix

\section{Explainability techniques}
\label{sec:app-explanation}

\subsection{Explanations for individual features}
Explanations for image-based tasks usually aggregate the importance of the input features at the pixel level, e.g. by taking an average over the individual color channels. This is done under the reasonable assumption that spatial locations are the smallest unit of input that can still be interpreted by humans.
The tasks considered in this work make use of node/edge features that are heterogeneous and individually interpretable. Therefore, we choose to present the explanations at the feature level, rather than aggregating at node or edge level.
Furthermore, we observed that the sign of the gradients produced with Sensitivity Analysis can provide additional context to the explanation. For this reason, the visualizations in this appendix will make use of the gradient "as is" and not of its squared norm.\\
Overall, we observe that explanations produced by variation-based methods tend to diverge from how a human would intuitively describe the process in terms of causes and effects. Decomposition-based methods result instead in more natural explanations.
We posit that the decomposition of the output signal makes LRP more suitable for the categorical distribution of the relevant features on both nodes and edges.

\subsection{Layer-wise Relevance Propagation rules}
\label{sec:app-explanation-lrp}
Layer-wise Relevance Propagation (LRP) is a signal decomposition method introduced in \cite{bach2015plosone}, where the authors mainly propose two rules. \\
The former, known as $\alpha\beta$-rule:
\vspace{-5pt}\begin{equation*}
    R^{(\ell)}_i = \sum_j \left( \alpha\frac{z_{ij}^+}{\sum_{i'}z_{i'j}^+ + b_j^+} + \beta\frac{z_{ij}^-}{\sum_{i'}z_{i'j}^-  + b_j^-} \right) R^{(\ell+1)}_i,
\end{equation*}\vspace{-5pt}

where $\alpha+\beta=1$, $x$ is the input of the layer, $w$ are its weights, and $z_{ij} = x^{(\ell)}_i w_{ij}$. \\
The latter, known as $\epsilon$-stabilized rule:
\vspace{-5pt}\begin{equation*}
    R^{(\ell)}_i = \sum_j \frac{z_{ij}^+}{\sum_{i'}z_{i'j}^+ + b_j^+ + \epsilon} R^{(\ell+1)}_i,
\end{equation*}\vspace{-5pt}
where $\epsilon$ is a small number to avoid division by zero.

We found the former with $\beta\neq 0$ to be quite unstable in the presence of zeros in the input or in the weights, a situation that occurs often when using one-hot encoding of categorical features and L1 regularization for the weights. Therefore, despite the $\alpha\beta$-rule should allow for more flexibility in tuning the ratio of positive and negative relevance, we chose the simpler $\epsilon$-stabilized rule with $\epsilon=10^{-16}$.

\subsection{LRP for regression}
Layer-wise Relevance Propagation was initially developed as an explanation technique for classification tasks. In the context of our Solubility experiment, we extend its application to a regression task. Since the prediction target is now a continuous variable, the explanations produced by LRP can be interpreted as "How much does this feature of this atom/bond contribute, positively or negatively, to the final predicted value?".\\
Also note, that due to the use of bias terms in our networks, the conservation property of LRP does not hold in full. Some relevance, in fact, will inevitably be attributed to the biases, that are internal parameters of the model and therefore not interpretable.
\vspace{-5pt}
\section{Experiment details}
\label{sec:app-gn}

\subsection{Infection}
\label{sec:app-gn-infection}
\subsubsection*{Feature representation}

\begin{figure}[b!]
    \centering
    \includegraphics[width=.8\columnwidth]{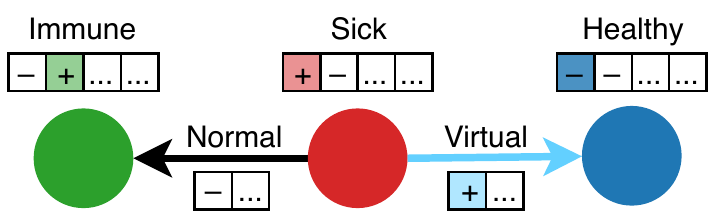}
    \vspace{-5pt}
    \caption{\textbf{Graph features for the Infection problem.}\\ {\small Nodes encode whether they are sick or healthy, immune or at risk, plus two uninformative features. Edges encode whether they are virtual or not, plus a single uninformative feature.}}
    \label{fig:infection-graph-features}
\end{figure}

The feature vectors $\bm{e}_k \in [-1, +1]^2$ and $\bm{n}_i \in [-1, +1]^4$ encode edge and node features respectively. Both include uninformative features that the network should learn to ignore and which should be attributed no importance by the explanation techniques (Fig.~\ref{fig:infection-graph-features}). 
Notably, binary features are encoded as $\{-1,+1\}$ rather than $\{0,1\}$, while this does not affect variation-based models (SA and GPB), it facilitates the propagation of relevance to the input when LRP is used.\\
The synthetic dataset used for training contains $100,000$ with 30 or fewer nodes generated with the Barab\'{a}si-Albert algorithm. The datasets used for validation and testing contain graphs of up to 60 nodes and different percentages of sick and immune nodes.

\subsubsection*{Architecture and training}
\begin{figure*}[h!]
\begin{subfigure}{.6\linewidth}
    \includegraphics[width=1.\linewidth]{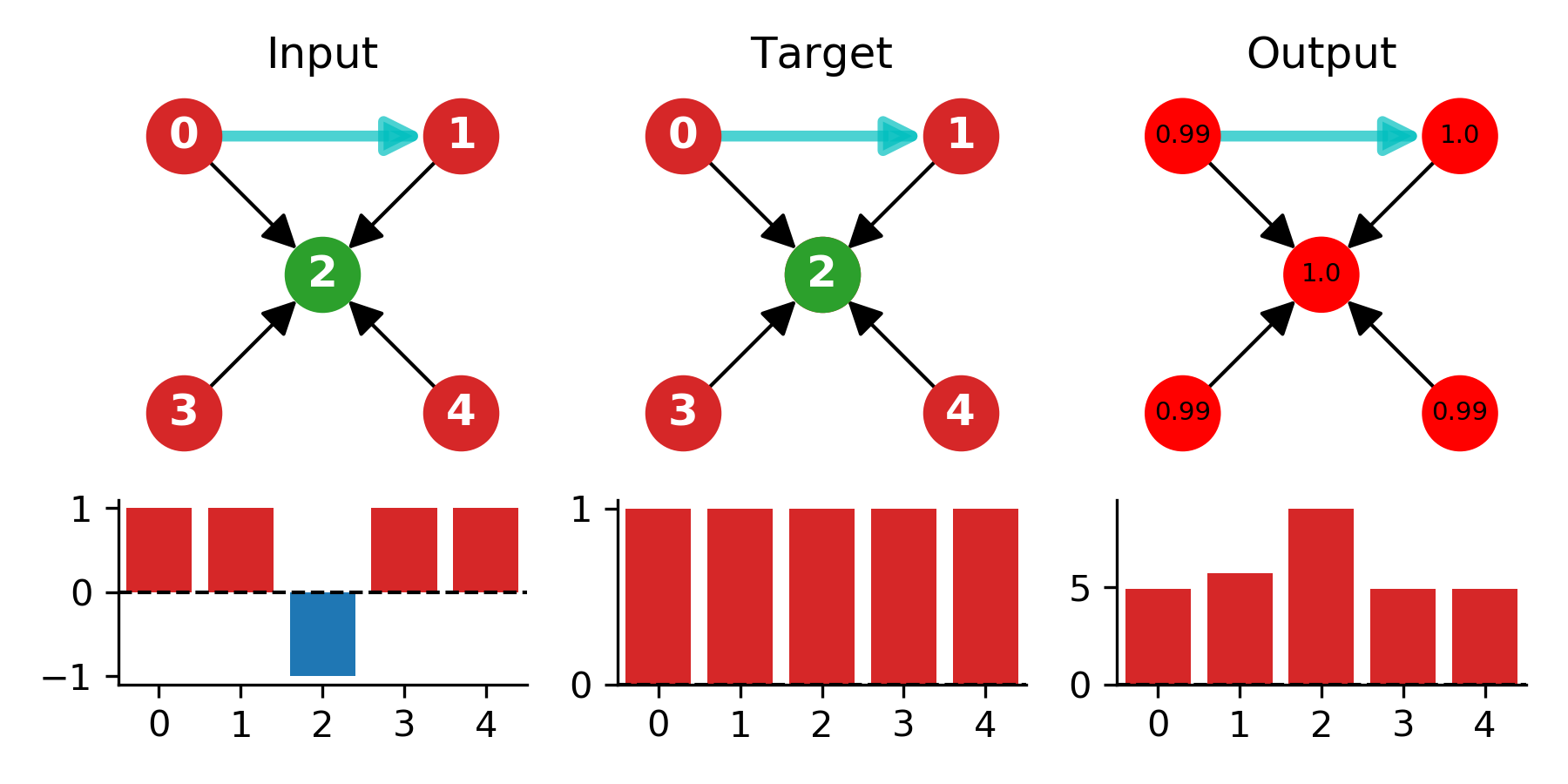}
\end{subfigure}\hfill\newline%
\begin{subfigure}{0.44\linewidth}
    \centering
    \includegraphics[width=1.\linewidth]{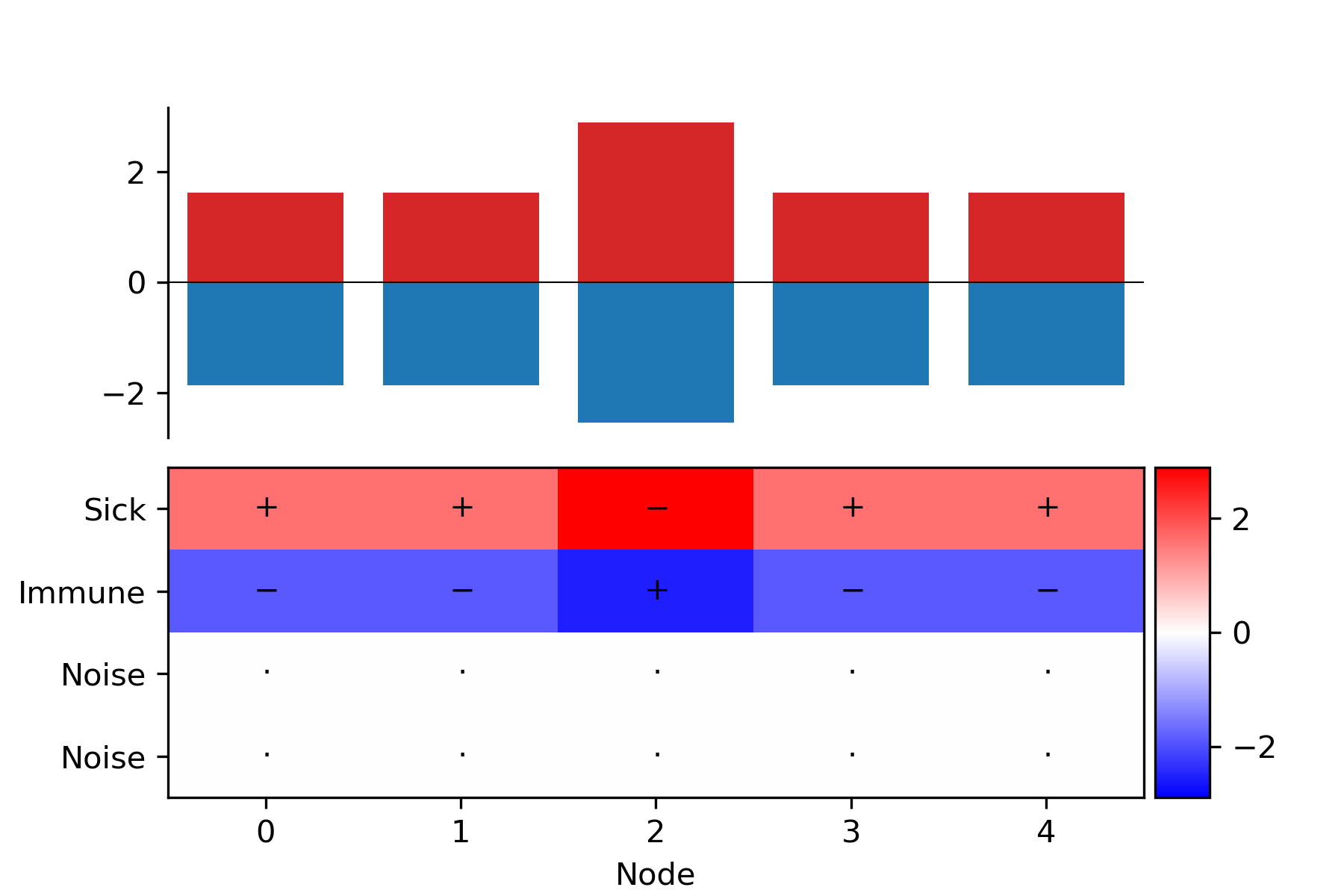}
\end{subfigure}%
\begin{subfigure}{0.44\linewidth}
    \centering
    \includegraphics[width=1.\linewidth]{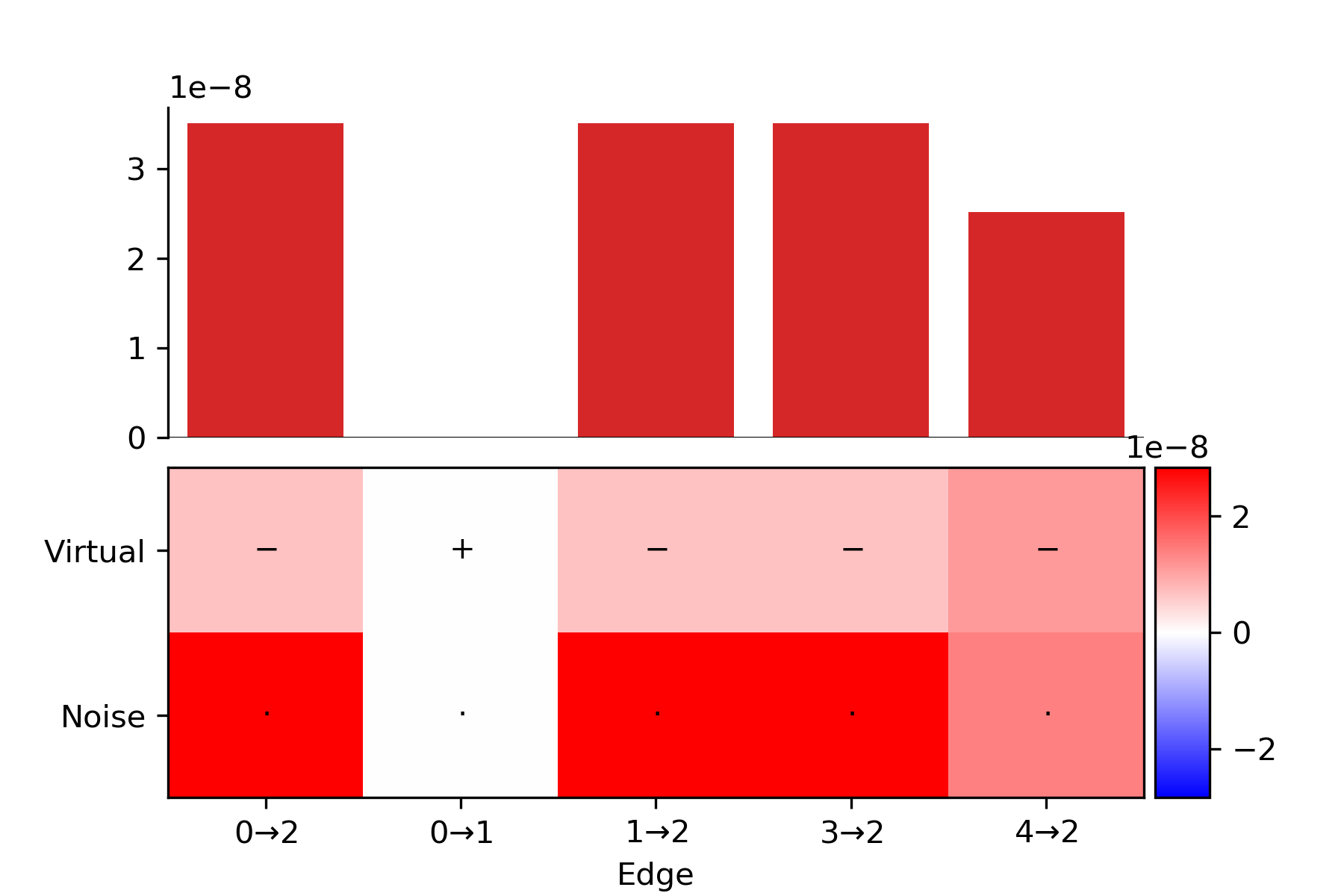}
\end{subfigure}%
\begin{subfigure}{0.11\linewidth}
    \includegraphics[width=1.\linewidth]{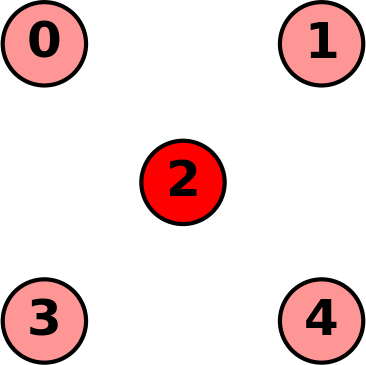}
    \caption*{SA}
\end{subfigure}\newline%
\begin{subfigure}{0.44\linewidth}
    \centering
    \includegraphics[width=1.\linewidth]{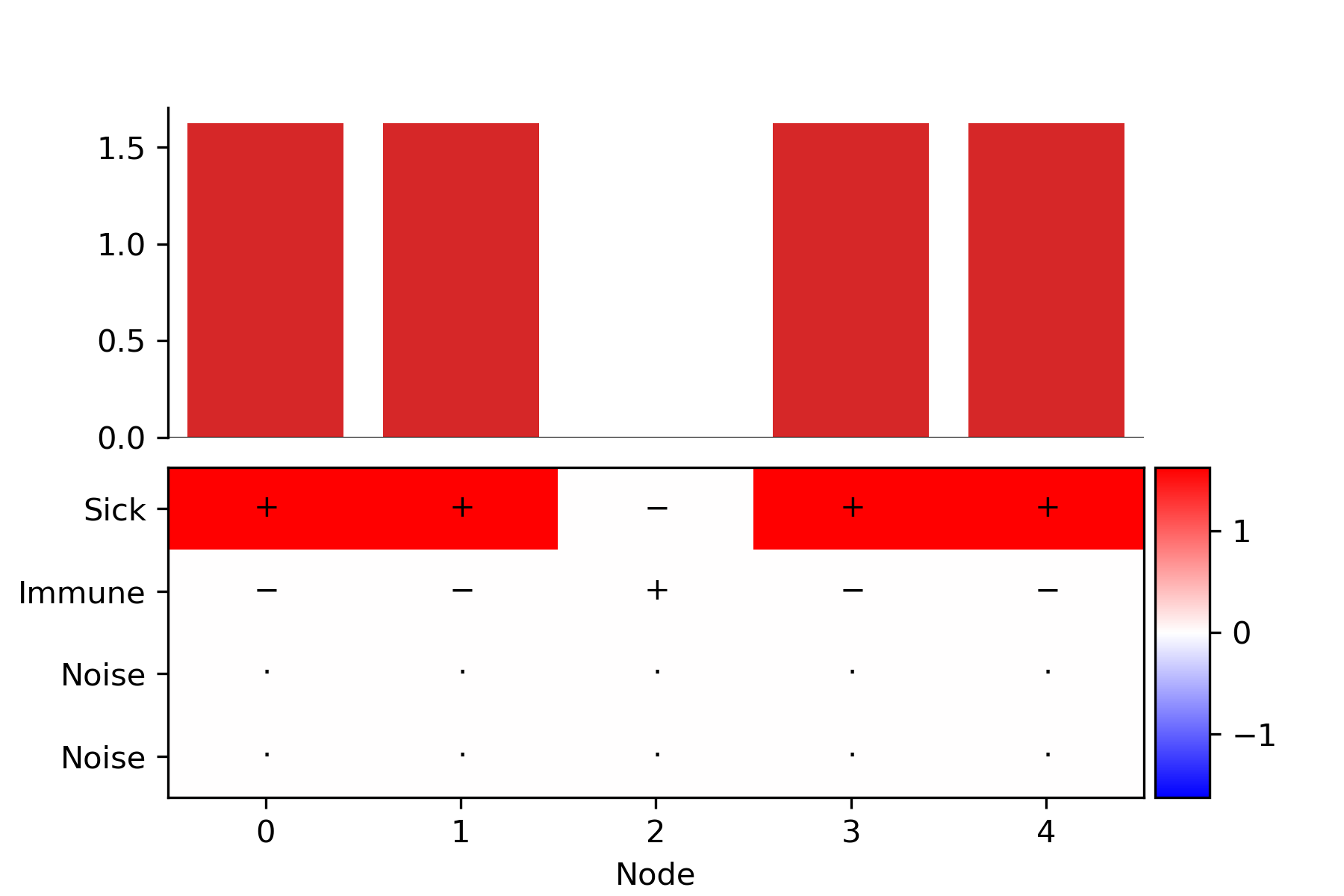}
\end{subfigure}%
\begin{subfigure}{0.44\linewidth}
    \centering
    \includegraphics[width=1.\linewidth]{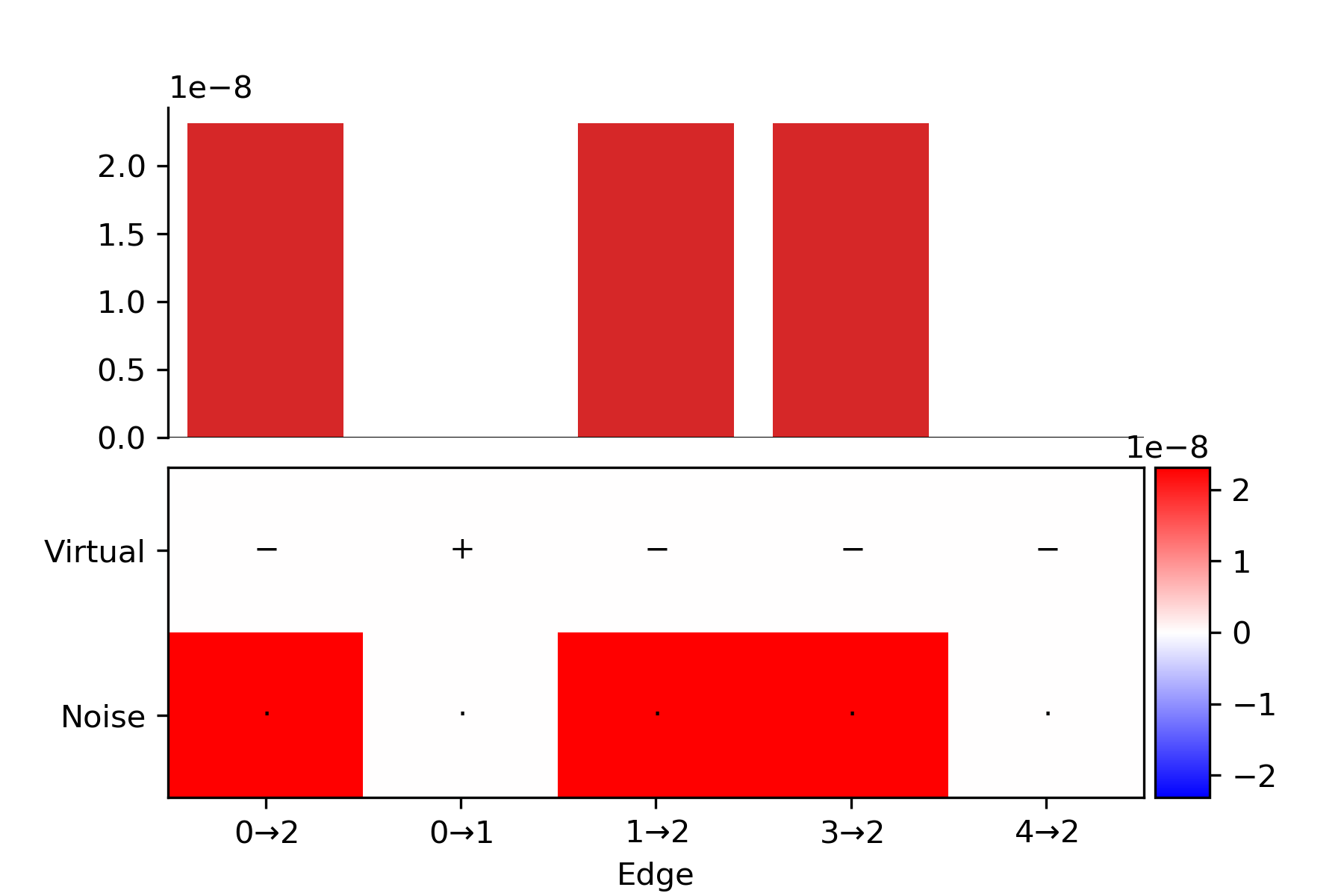}
\end{subfigure}%
\begin{subfigure}{0.11\linewidth}
    \includegraphics[width=1.\linewidth]{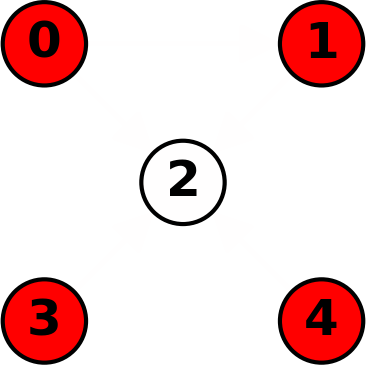}
    \caption*{GBP}
\end{subfigure}\newline%
\begin{subfigure}{0.44\linewidth}
    \centering
    \includegraphics[width=1.\linewidth]{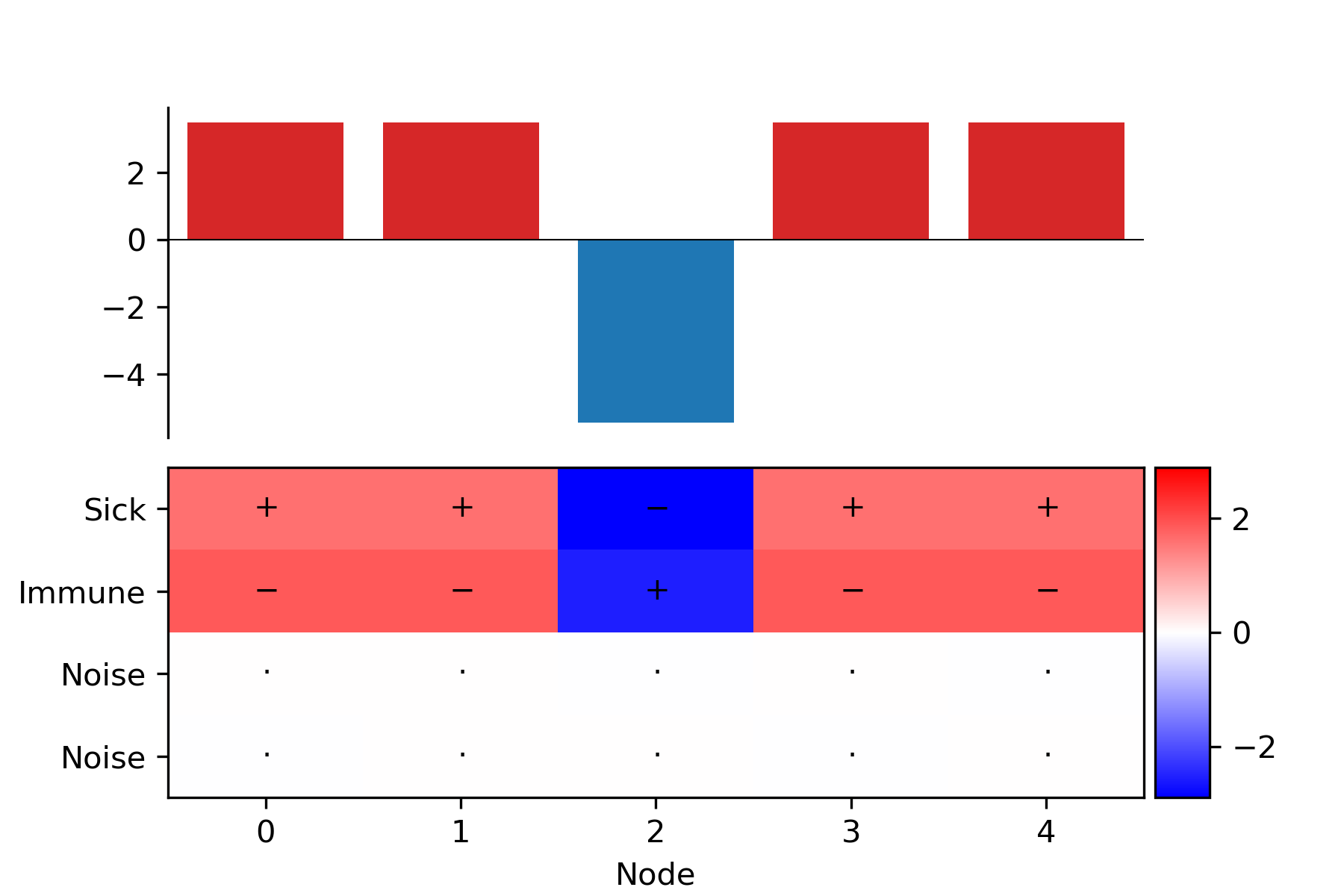}
\end{subfigure}%
\begin{subfigure}{0.44\linewidth}
    \centering
    \includegraphics[width=1.\linewidth]{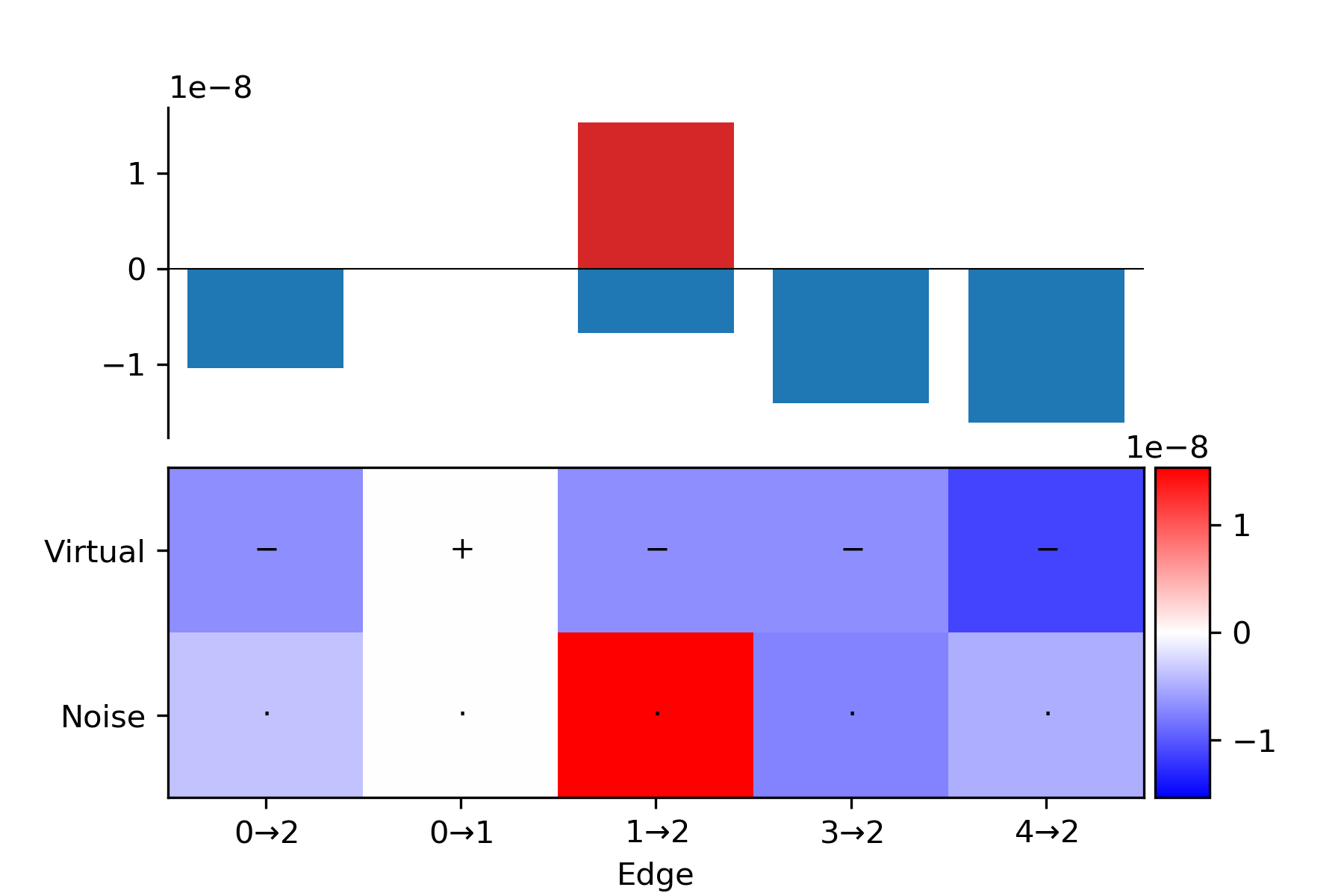}
\end{subfigure}%
\begin{subfigure}{0.11\linewidth}
    \includegraphics[width=1.\linewidth]{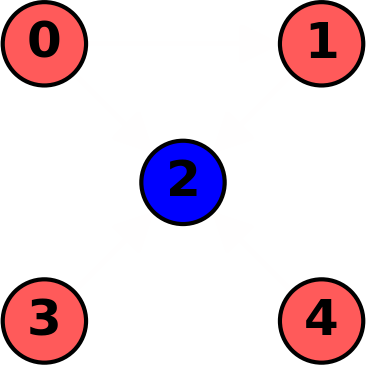}
    \caption*{LRP}
\end{subfigure}%
\caption{\textbf{Graph 4 - Sum pooling.} In this case, the network trained with sum pooling fails to produce the correct prediction. In contrast to the image domain, where the number of pixels in a convolutional neighborhood is fixed, GNs aggregate information over a potentially very large neighborhood, where even small contributions can add up and produce a wrong answer.}
\label{fig:appendix-graph-5-sum}
\end{figure*}

The network used for the Infection task makes use of a single layer of graph processing as in Eq.~\ref{eq:gn-functions}, without graph-level features. 
The update functions for the edges and the nodes are shallow multi-layer perceptrons, with ReLU activations and we use sum/max pooling to aggregate the edges incident to a node.
We use the Adam optimizer \cite{kingma2014adam} to minimize the binary cross-entropy between the per-node predictions and the ground truth.\\
Multiple choices of hyperparameters such as learning rate, number of neurons in the hidden layers and L1 regularization yield similar outcomes.
Both sum and max pooling perform well, but the former fails in some corner cases (Fig.~\ref{fig:appendix-graph-5-sum}).

\subsection{Solubility}

\label{sec:app-gn-solubility}

\subsubsection*{Dataset and features}
The Solubility dataset is the same as \cite{duvenaud2015convolutional}, consisting of around 1000 organic molecules represented as SMILES strings and their measured solubility in water. The molecules are represented as graphs having atoms as nodes (with degree, number of hydrogens, implicit valence and type as features) and bonds as edges (with their type, whether they are conjugate and whether they are in a ring as features). 

\subsubsection*{Architecture and training}
As optimization objective we use the mean squared error between the measured log-solubility and the global features $\bm{u}$ of the output graph of a multi-layer GN, where each layer performs updates the graph as in Eq.~\ref{eq:gn-functions}. Using multiple layers of graph convolution allows the network to aggregate information at progressively larger scales, starting from the local neighborhood and extending to wider groups of atoms. Dropout is applied at the output of every linear transformation, as a technique to counteract overfitting.\\
We tested multiple combinations of hyperparameters and obtain results comparable to \cite{duvenaud2015convolutional} using 3-5 hidden graph layers with a dimensionality of either 64, 128 or 256 and sum/mean pooling for all aggregation operations. Max pooling performed much worse, probably due to the nature of the task.

\section{Additional results}
\label{sec:app-more-results}

\subsection{Infection}
\label{sec:app-infection-results}

\subsubsection*{Example prediction on a medium-sized graph}

\begin{figure*}[hb!]
\begin{subfigure}{1.\linewidth}
    \centering
    \includegraphics[width=1.\linewidth]{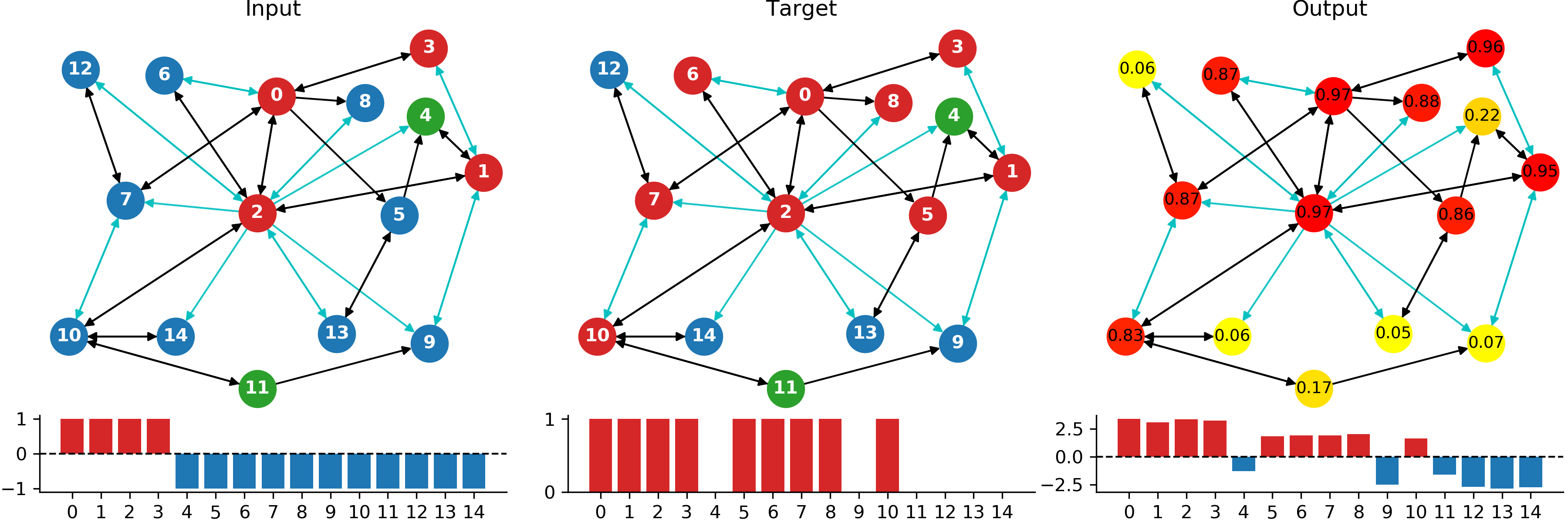}
\end{subfigure}%
\caption{\textbf{Graph 1: nodes 0, 1, 2 and 3 are initially sick; nodes 4 and 11 are immune; the others are healthy.} After one propagation step, the infection reaches nodes 5, 6, 7 and 10. The network predicts the correct label for every node of the graph, following the spread of the infection along non-virtual edges to non-immune nodes. The figures that follow are a visualization of the explanations produced for nodes: 10, 13, 4}
\label{fig:appendix-biggraph-output}
\end{figure*}

\begin{figure*}[h!]
\begin{subfigure}{0.7\linewidth}
    \centering
    \includegraphics[width=1.\linewidth]{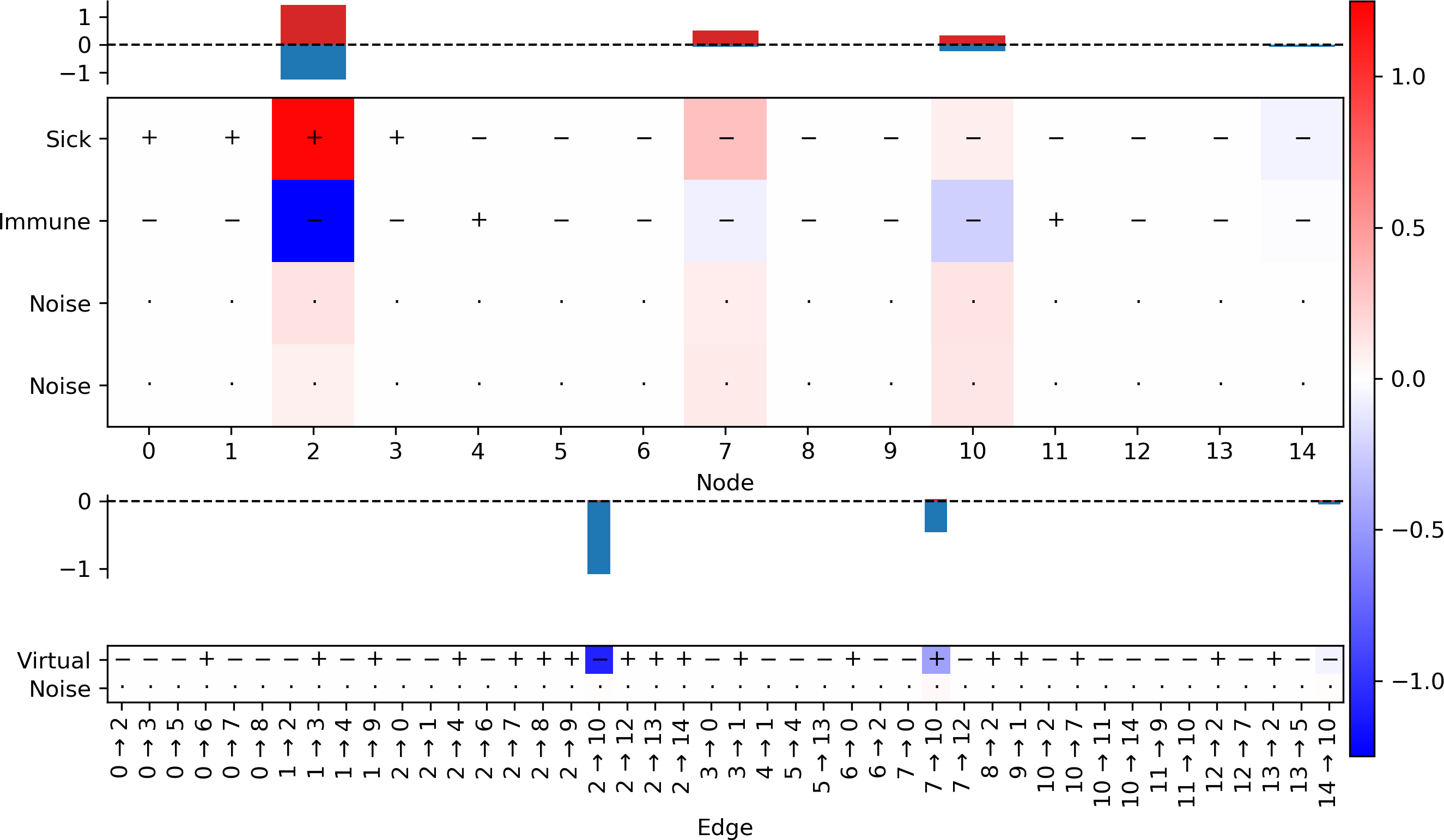}
\end{subfigure}%
\begin{subfigure}{0.3\linewidth}
    \includegraphics[width=1.\linewidth]{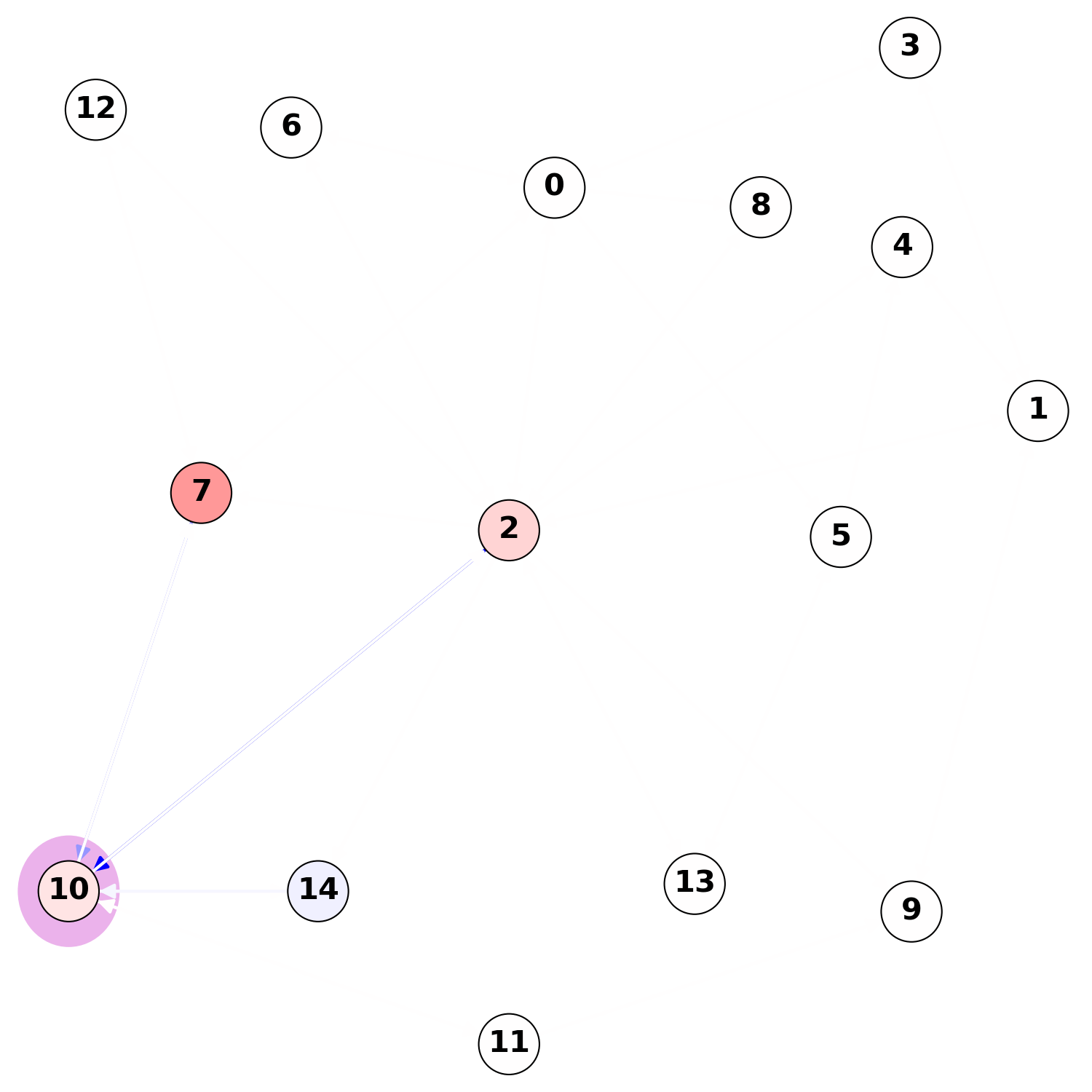}
    \caption*{Sensitivity Analysis (SA)}
\end{subfigure}\newline%
\begin{subfigure}{0.7\linewidth}
    \centering
    \includegraphics[width=1.\linewidth]{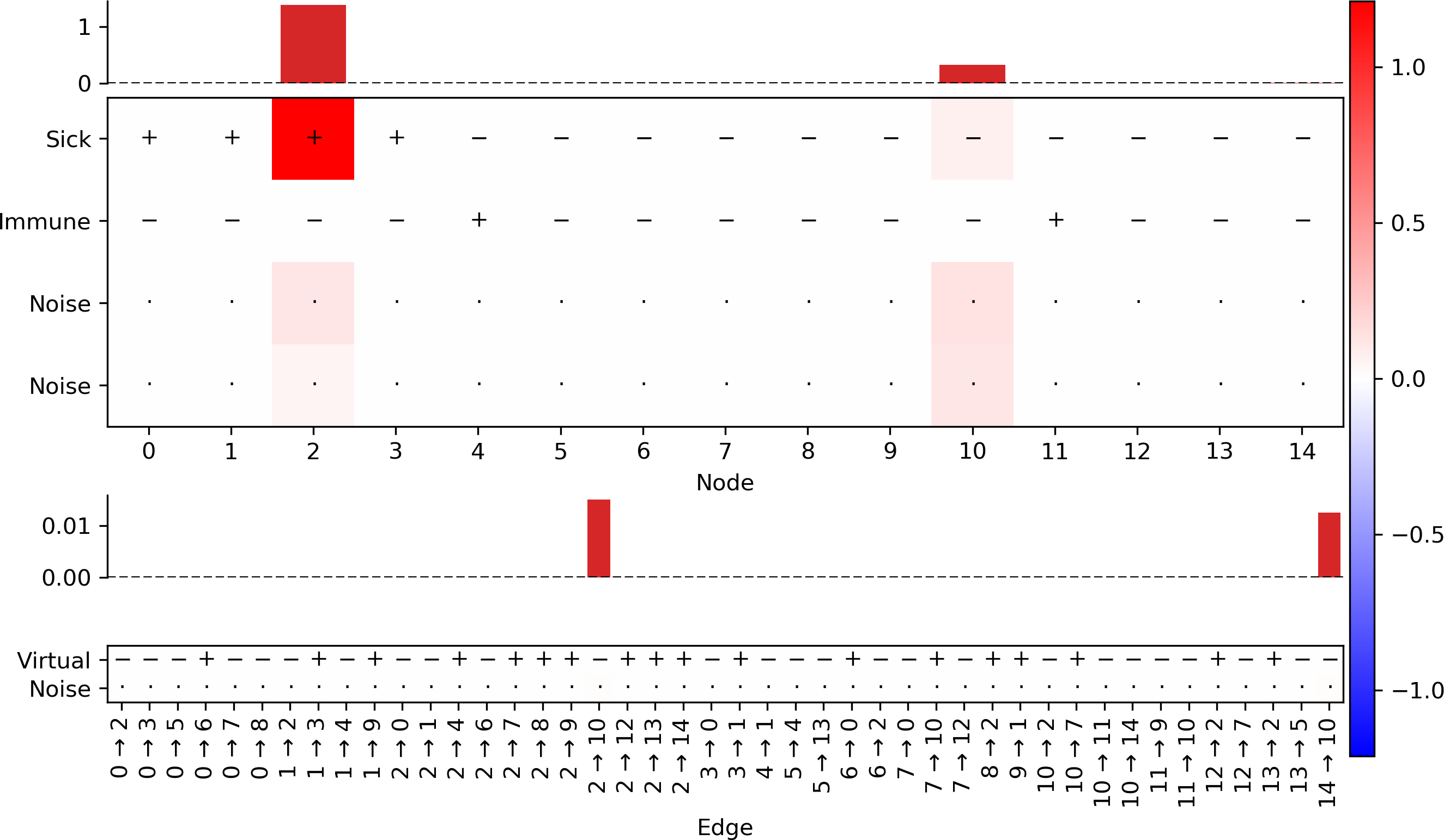}
\end{subfigure}%
\begin{subfigure}{0.3\linewidth}
    \includegraphics[width=1.\linewidth]{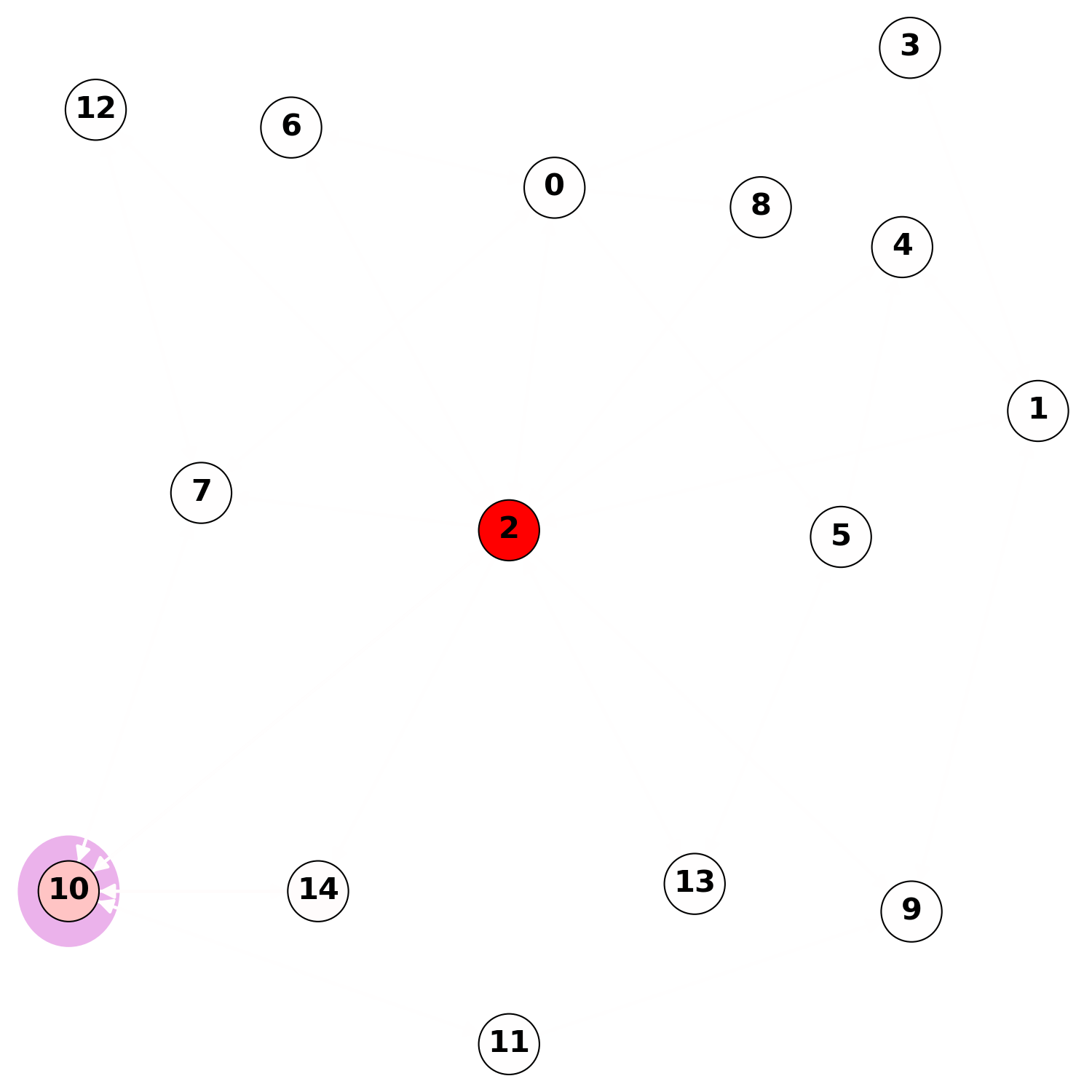}
    \caption*{Guided Backpropagation (GBP)}
\end{subfigure}\newline%
\begin{subfigure}{0.7\linewidth}
    \centering
    \includegraphics[width=1.\linewidth]{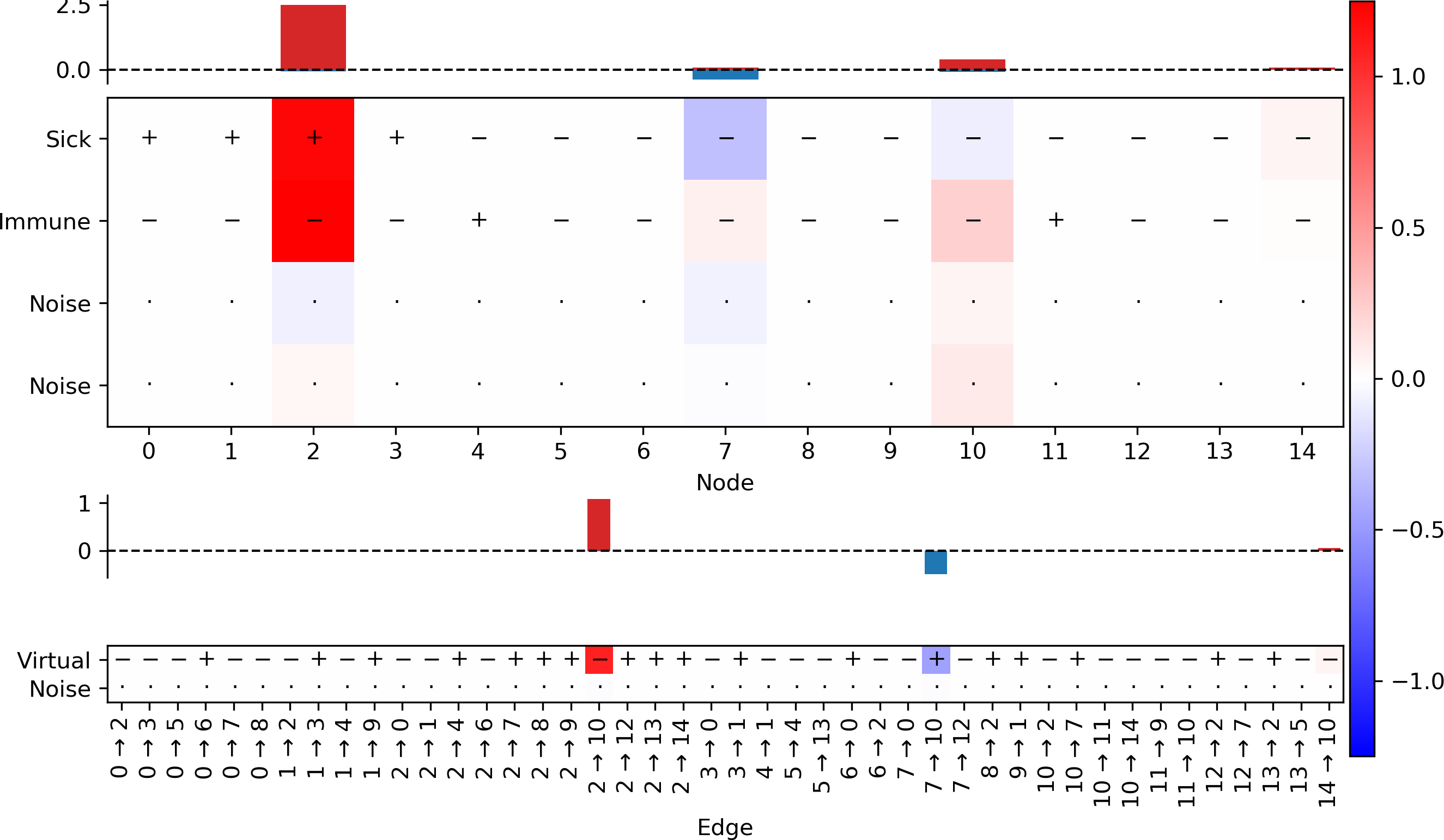}
\end{subfigure}%
\begin{subfigure}{0.3\linewidth}
    \includegraphics[width=1.\linewidth]{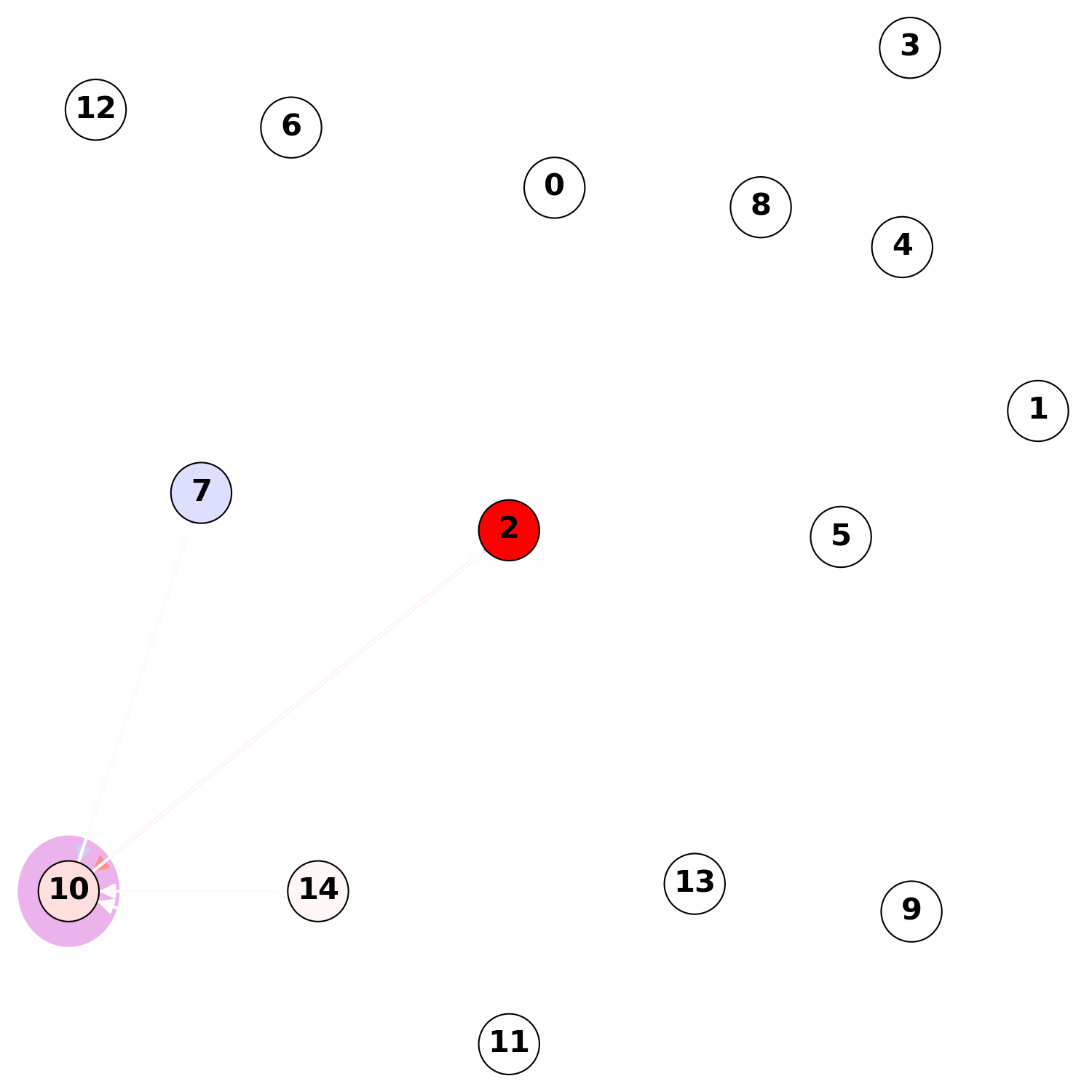}
    \caption*{Layer-wise Relevance Propagation (LRP)}
\end{subfigure}%
\caption{\textbf{Graph 1 - Explanation for node 10.} Node 10 is initially healthy and becomes infected due to the incoming edge from node 2. SA suggests that if node 2 was more sick and the edge $2\to 10$ was less virtual, then the prediction for node 10 would be even higher. GBP yield a similar explanation, but limited to the positive gradients. LRP decomposes the positive prediction into a positive contribution from the fact that node 2 is sick and that the edge $2\to 10$ is not virtual, and a negative contribution from the fact that node 7 is healty and its connection to node 10 is virtual.}
\label{fig:appendix-biggraph-10}
\end{figure*}

\begin{figure*}[h!]
\begin{subfigure}{0.7\linewidth}
    \centering
    \includegraphics[width=1.\linewidth]{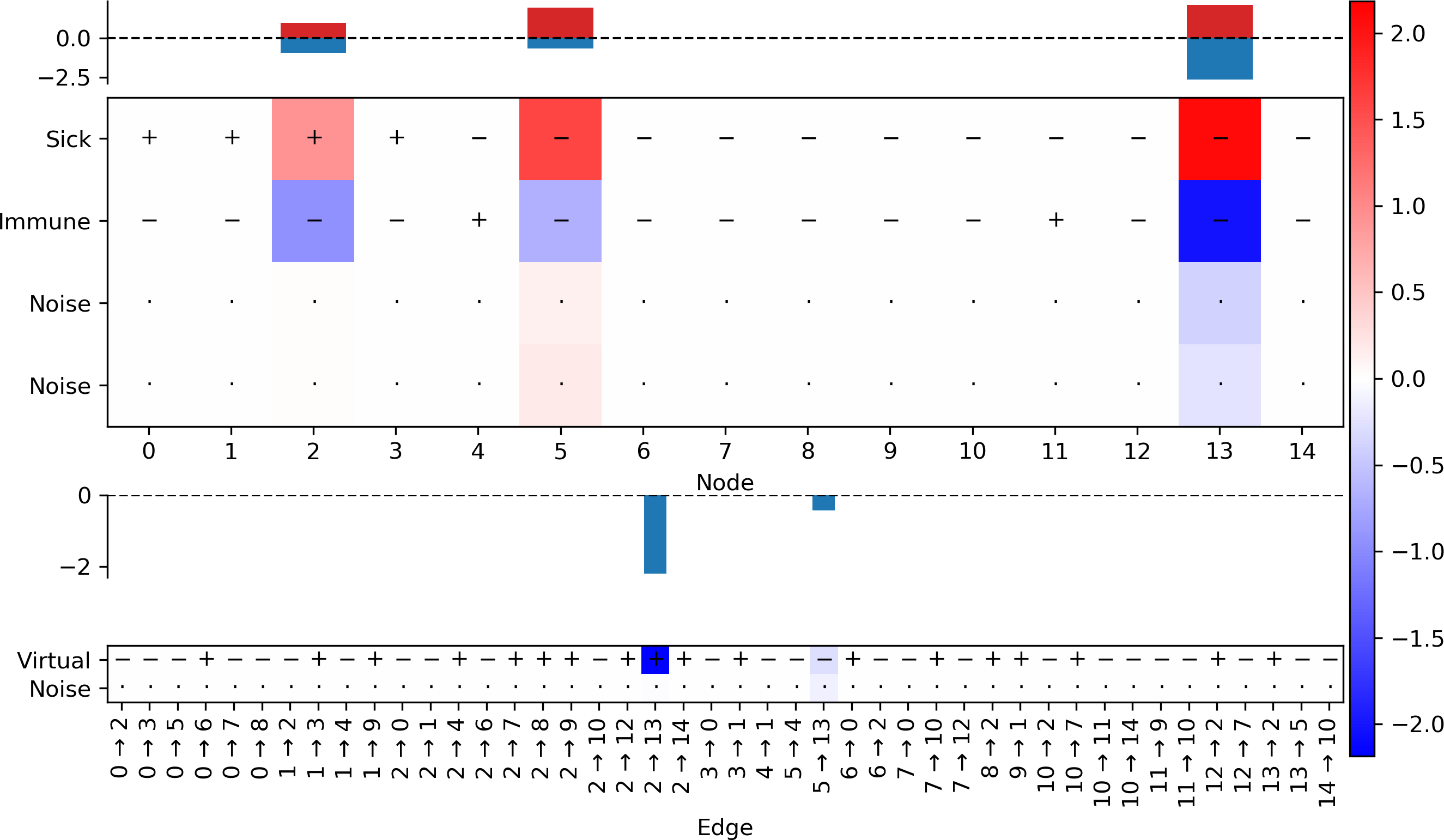}
\end{subfigure}%
\begin{subfigure}{0.3\linewidth}
    \includegraphics[width=1.\linewidth]{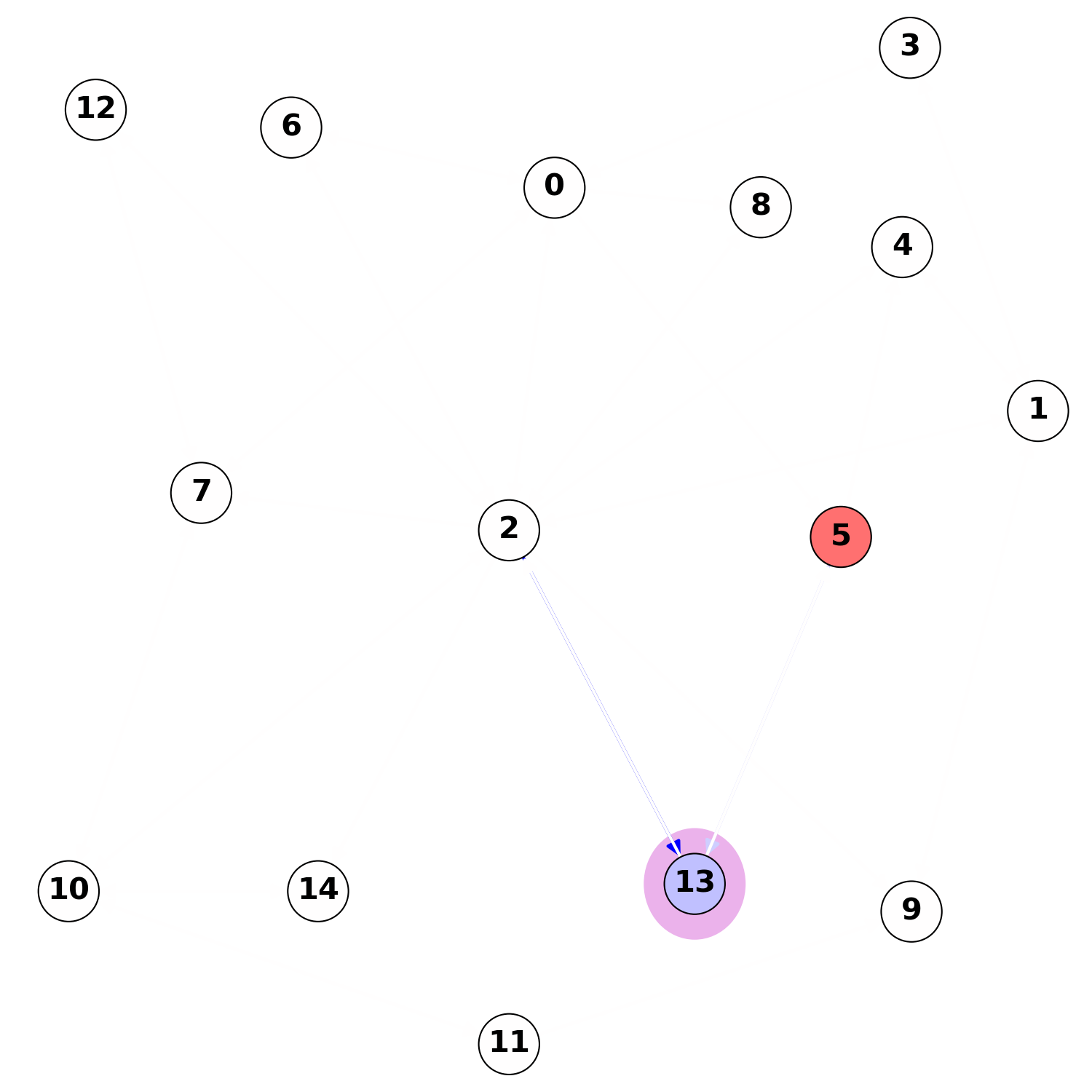}
    \caption*{Sensitivity Analysis (SA)}
\end{subfigure}\newline%
\begin{subfigure}{0.7\linewidth}
    \centering
    \includegraphics[width=1.\linewidth]{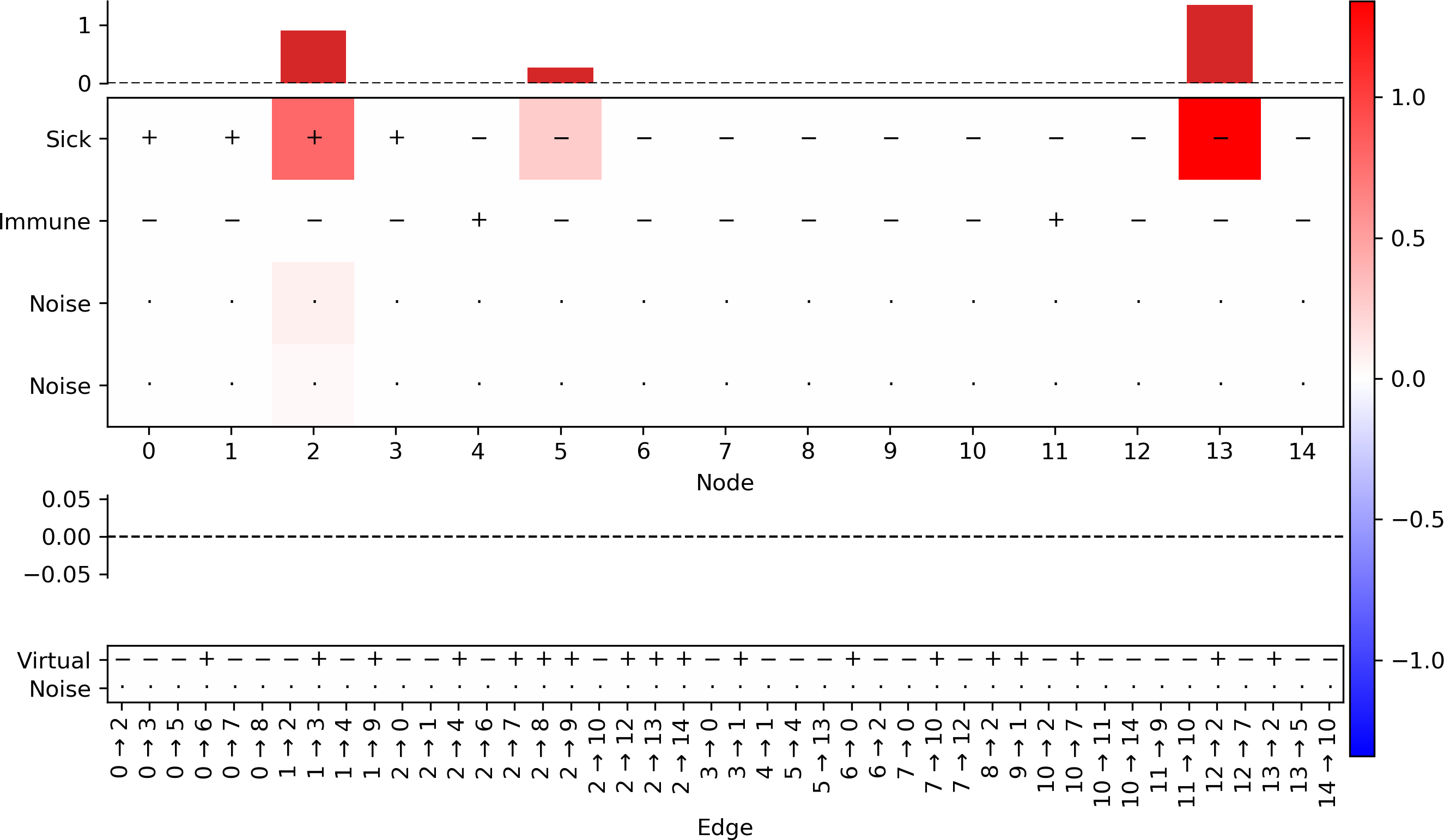}
\end{subfigure}%
\begin{subfigure}{0.3\linewidth}
    \includegraphics[width=1.\linewidth]{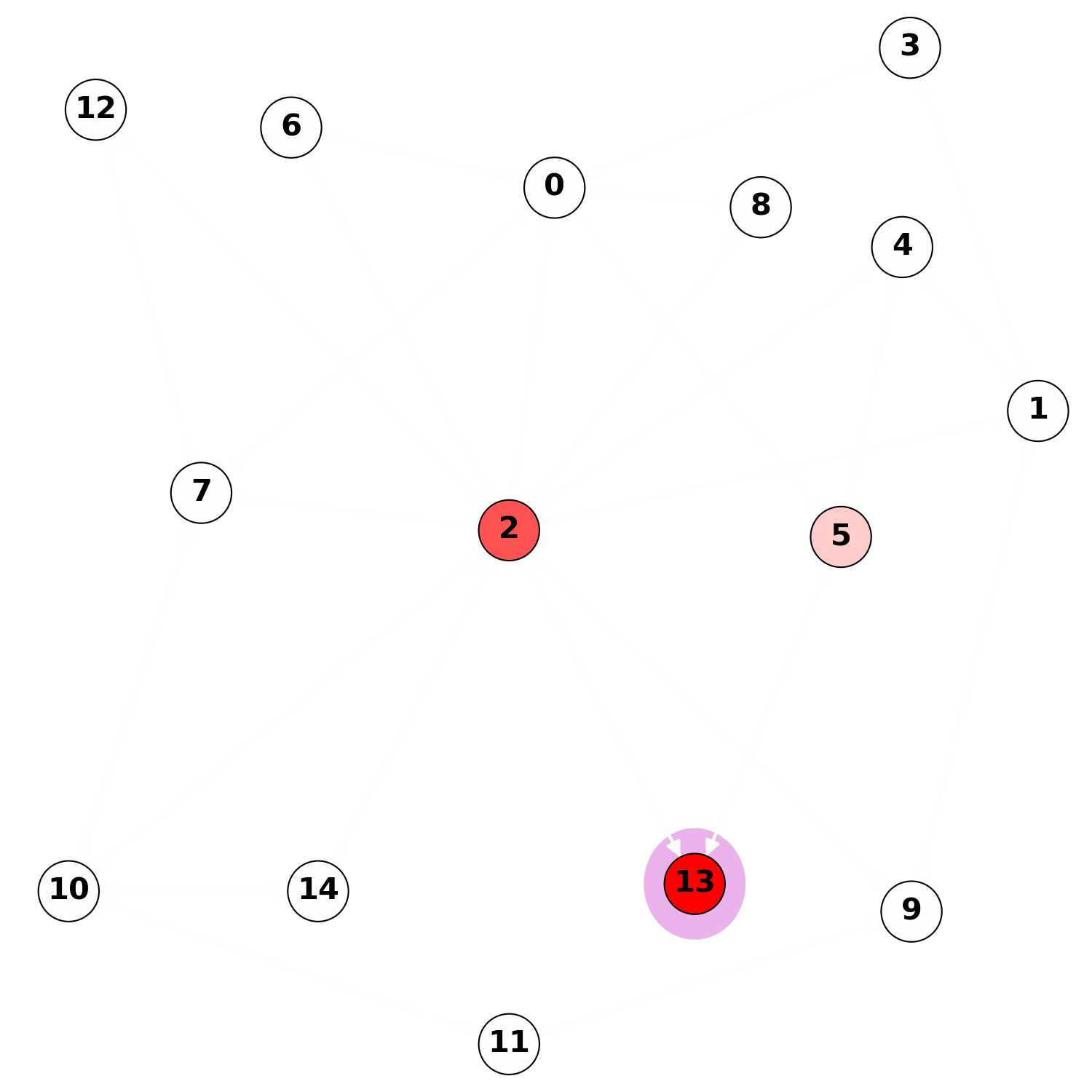}
    \caption*{Guided Backpropagation (GBP)}
\end{subfigure}\newline%
\begin{subfigure}{0.7\linewidth}
    \centering
    \includegraphics[width=1.\linewidth]{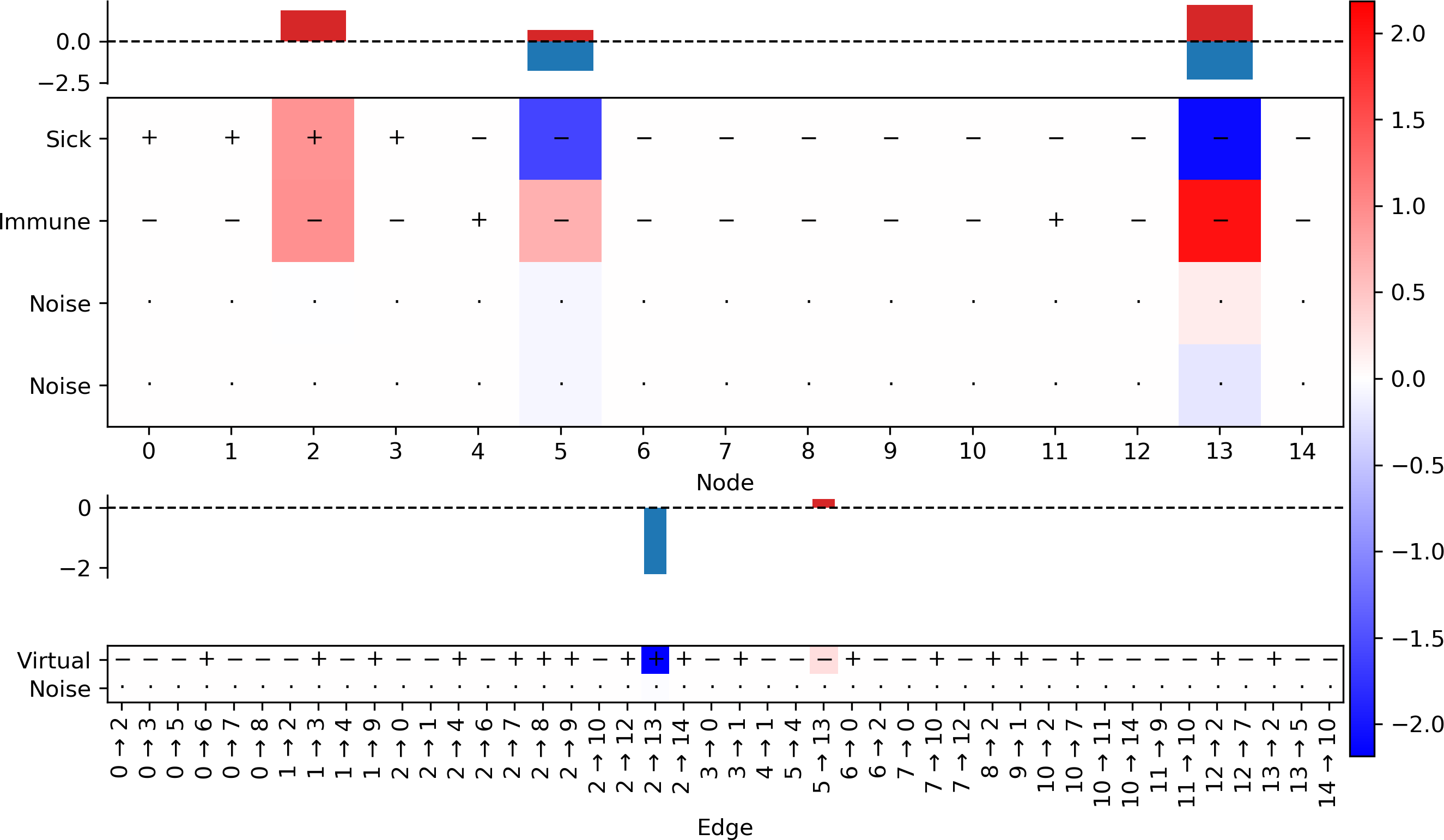}
\end{subfigure}%
\begin{subfigure}{0.3\linewidth}
    \includegraphics[width=1.\linewidth]{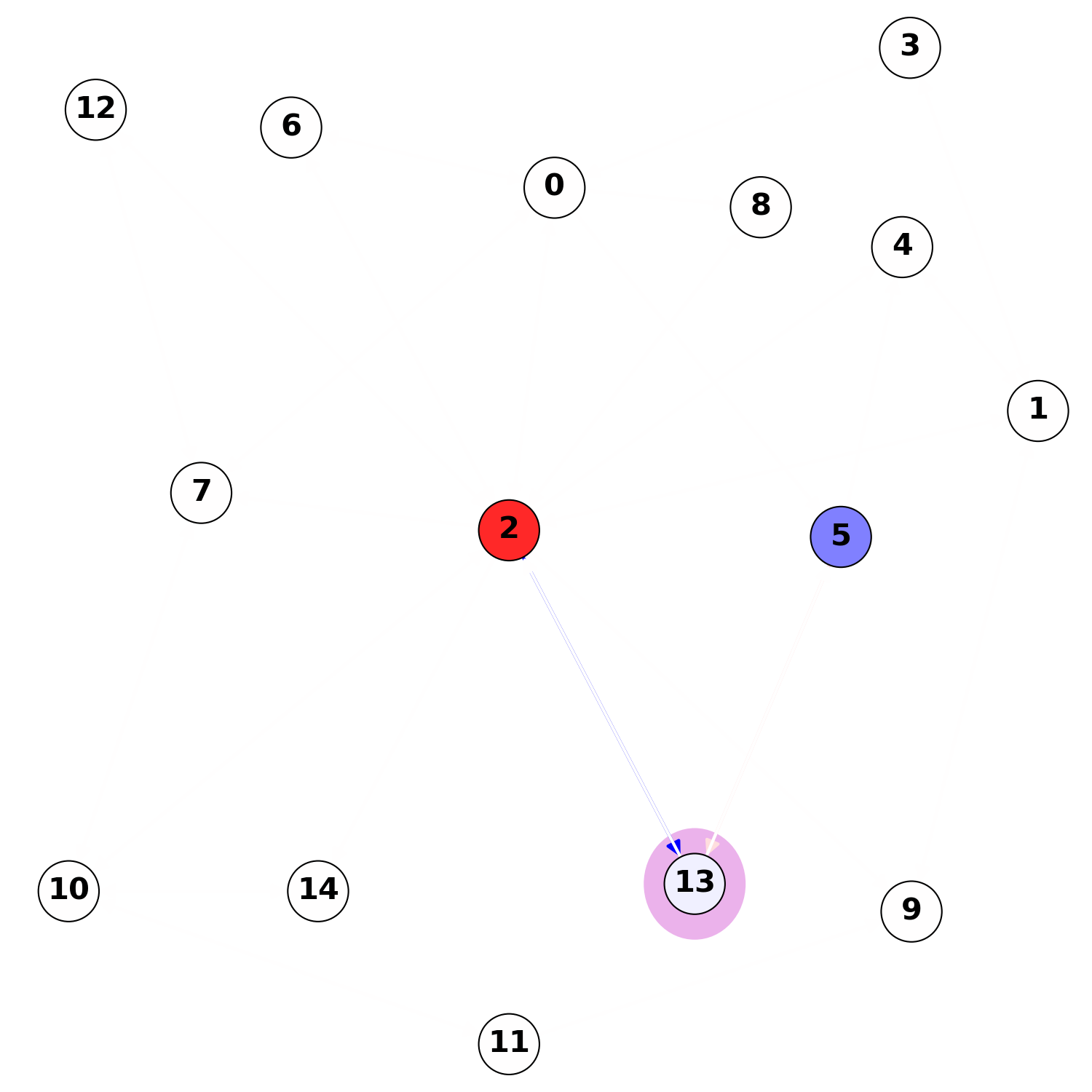}
    \caption*{Layer-wise Relevance Propagation (LRP)}
\end{subfigure}%
\caption{\textbf{Graph 1 - Explanation for node 13.} Node 13 remains healthy after one propagation step and the network correctly predicts a negative output (before activation). SA explanation suggests that for the prediction to be higher, the edge $2\to 13$ should be less virtual, or banally that node 13 should be more sick and less immune. GBP produces a similar explanation, but only captures the positive gradient on the "sick" feature of node 13 itself. For LRP the fact that node 13 is initially healthy is a negative contribution to the output, together with the fact that its neighbor 5 is healthy as well; a positive contribution comes from the fact that node 2 is sick, which is counteracted by the negative contribution of the edge $2\to 13$.}
\label{fig:appendix-biggraph-13}
\end{figure*}

\begin{figure*}[h!]
\begin{subfigure}{0.7\linewidth}
    \centering
    \includegraphics[width=1.\linewidth]{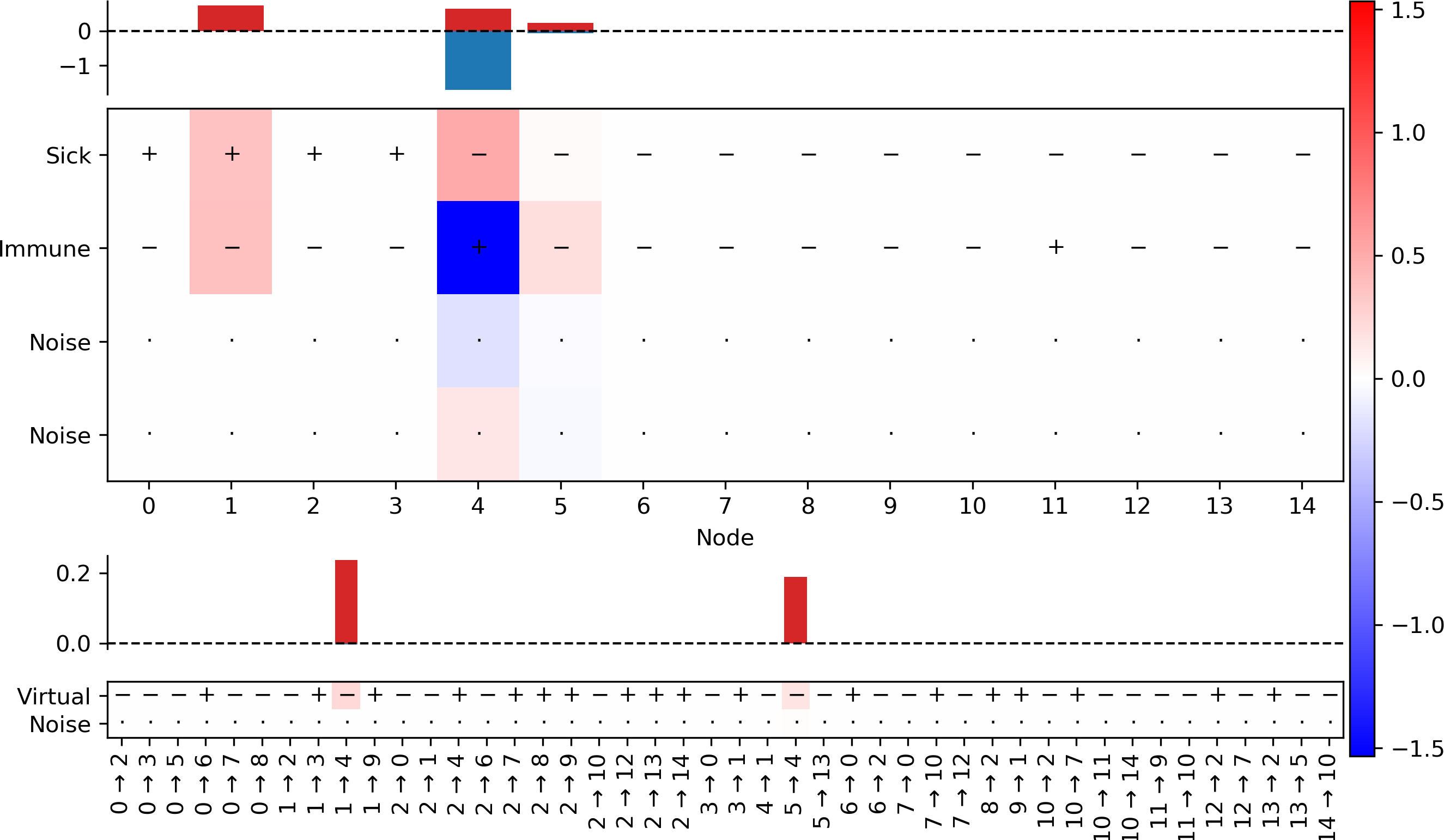}
\end{subfigure}%
\begin{subfigure}{0.3\linewidth}
    \includegraphics[width=1.\linewidth]{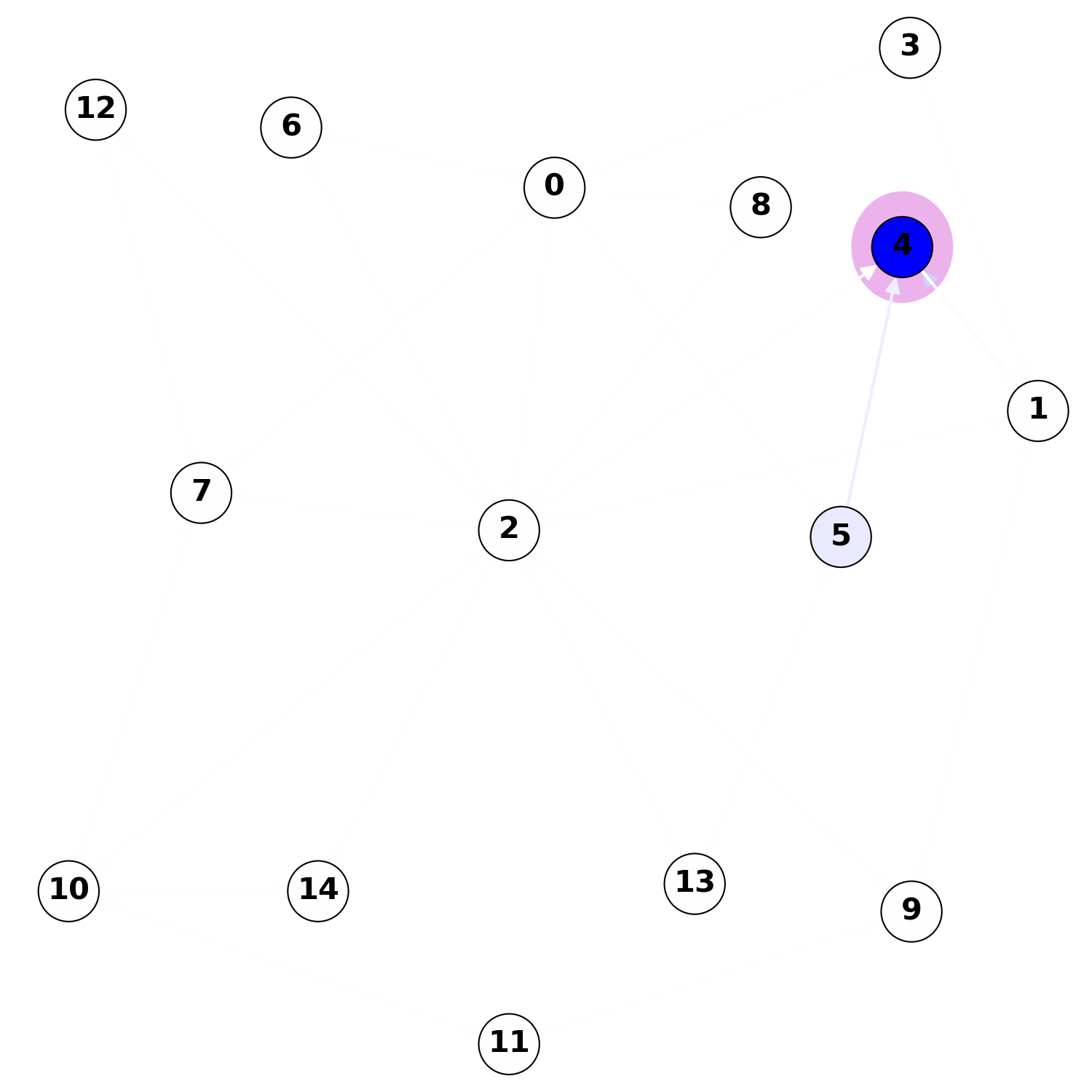}
    \caption*{Sensitivity Analysis (SA)}
\end{subfigure}\newline%
\begin{subfigure}{0.7\linewidth}
    \centering
    \includegraphics[width=1.\linewidth]{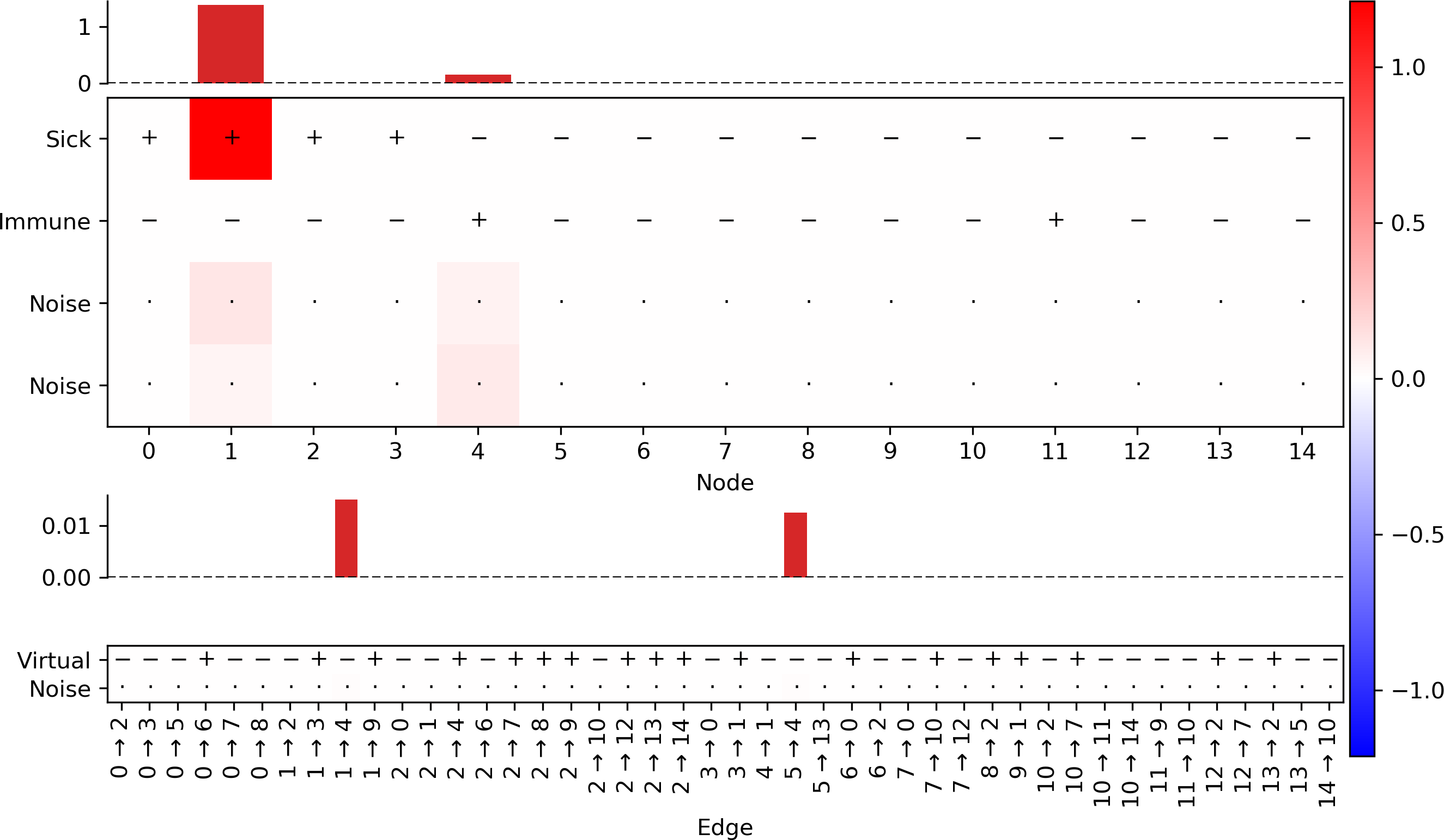}
\end{subfigure}%
\begin{subfigure}{0.3\linewidth}
    \includegraphics[width=1.\linewidth]{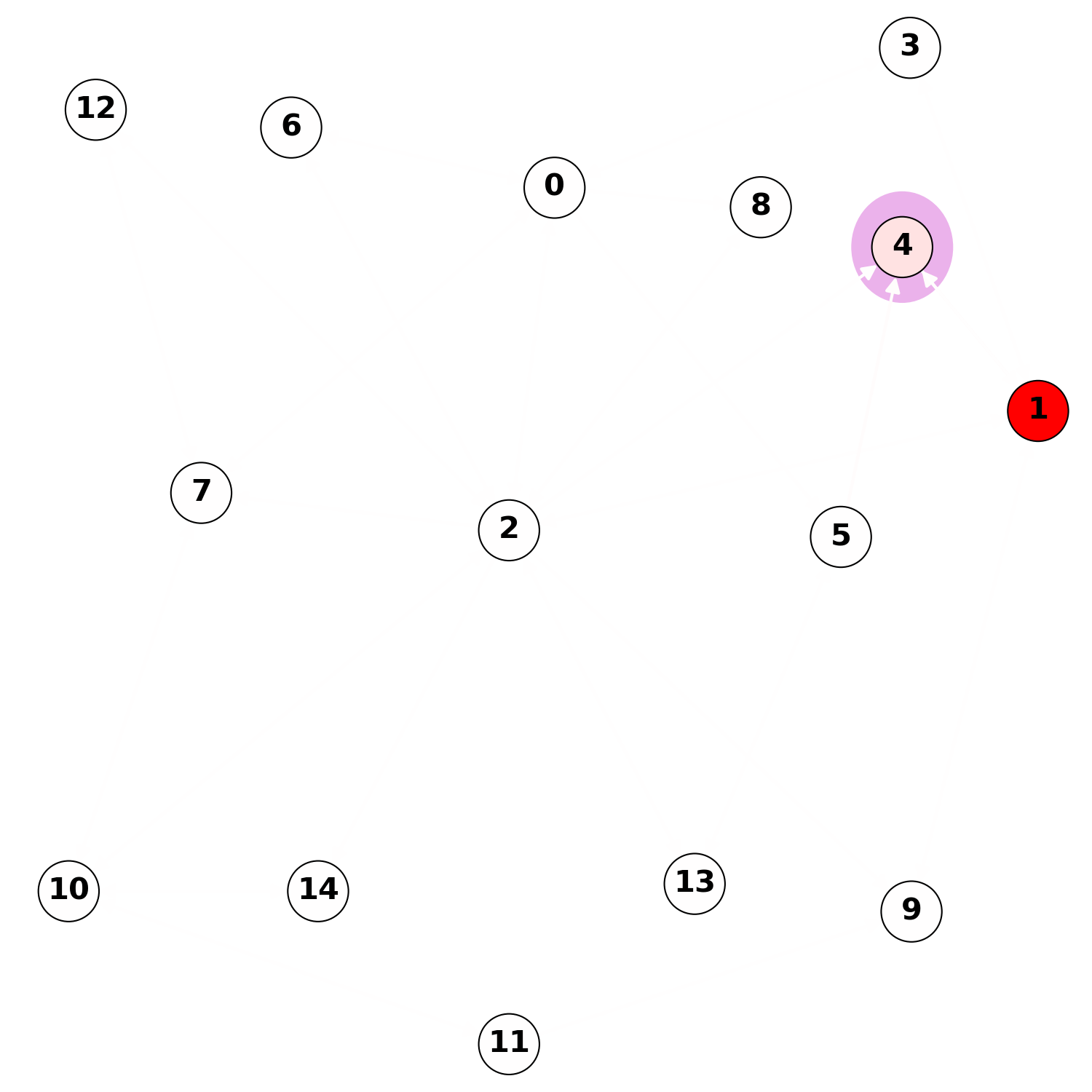}
    \caption*{Guided Backpropagation (GBP)}
\end{subfigure}\newline%
\begin{subfigure}{0.7\linewidth}
    \centering
    \includegraphics[width=1.\linewidth]{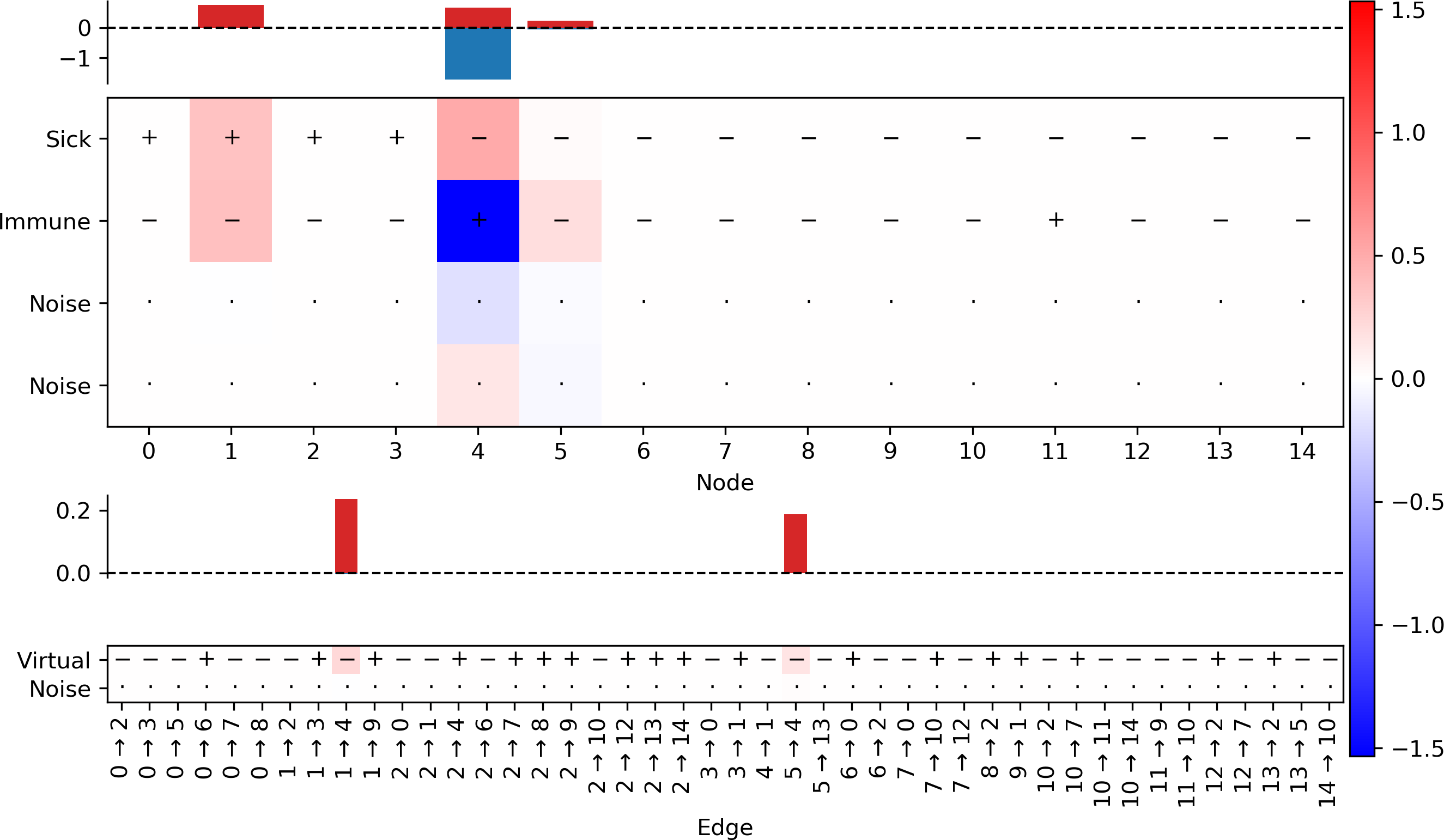}
\end{subfigure}%
\begin{subfigure}{0.3\linewidth}
    \includegraphics[width=1.\linewidth]{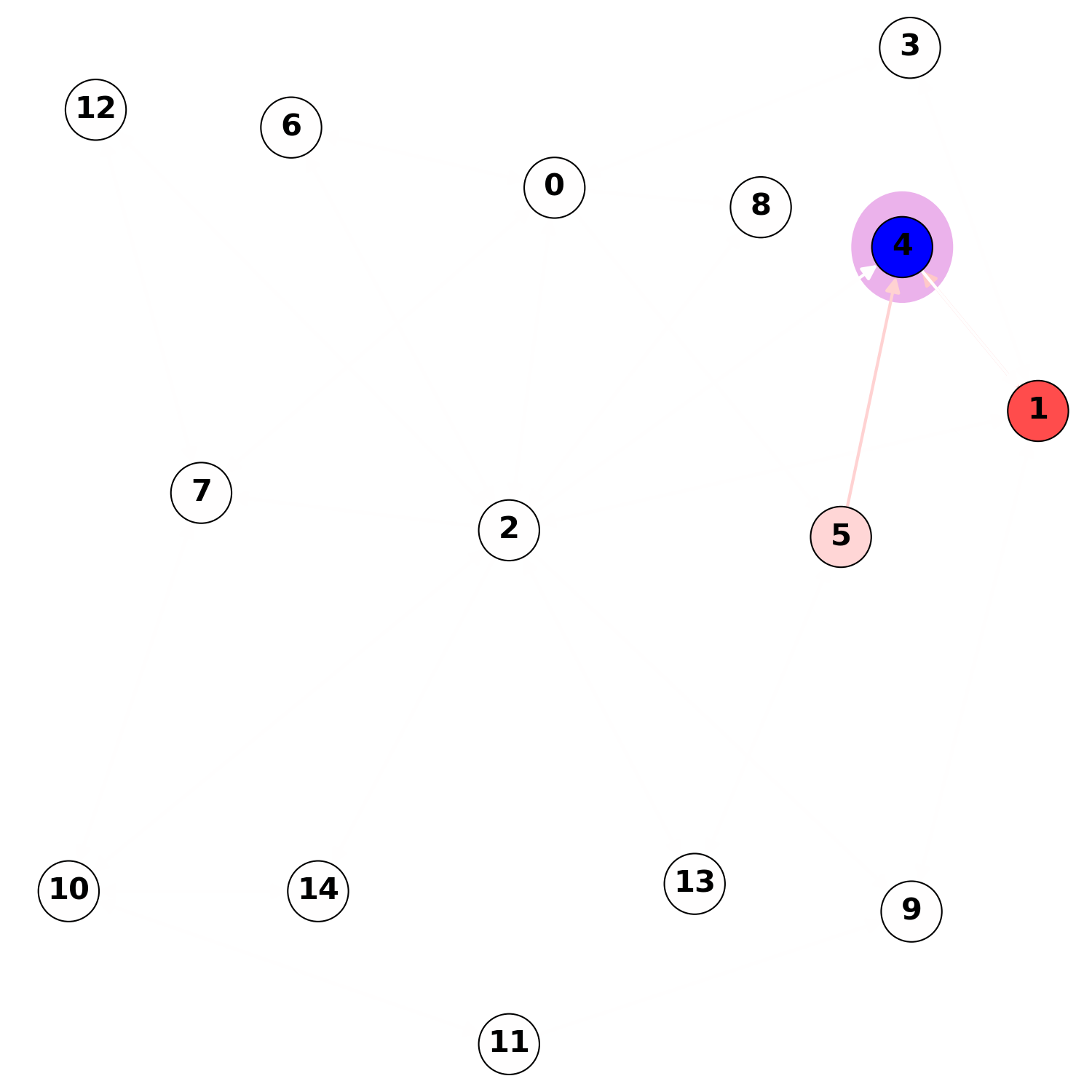}
    \caption*{Layer-wise Relevance Propagation (LRP)}
\end{subfigure}%
\caption{\textbf{Graph 1 - Explanation for node 4.} The node is immune and therefore remains healthy after one step of infection propagation. SA places a negative gradient on the "immune" feature of node 4, meaning that if it wasn't immune it would become infected. GBP focuses on node 1, which is a sick neighbor of node 4. LRP decompose most of the negative prediction into a negative contribution from the node itself, weakly counteracted by positive contributions from its neighbors.}
\label{fig:appendix-biggraph-4}
\end{figure*}

In the following pages we present an in-depth comparison between the three explainability methods we experimented with. We consider a graph with multiple sources of infection and immune nodes. The network, that uses max pooling to aggregate information from the incoming edges, correctly predicts the state of every node after one step of infection propagation (Fig.~\ref{fig:appendix-biggraph-output}). In the figures that follow, we present a visualization of the explanations produced for three nodes of the graph: one that becomes infected (Fig.~\ref{fig:appendix-biggraph-10}), one that receives no infection from its neighbors (Fig.~\ref{fig:appendix-biggraph-13}) and one that is immune (Fig.~\ref{fig:appendix-biggraph-4}). Refer to the captions for observations specific to every example.

\subsubsection*{Aggregation: max vs. sum comparison}
It then follows an overview of explanations obtained for smaller graphs. For every input graph, we show two predictions: one made by a GN that uses max pooling and one made by a GN that uses sum pooling. The predictions are followed by a visualization of the explanations produced with Sensitivity Analysis, Guided Backpropagation and Layer-wise Relevance Propagation (in this order). For each explainability method, we represent the values of the gradient/relevance as a heatmap over the individual features of every node/edge, as well as with a graphical representation of the graph. Refer to the captions for observations specific to every example.

\subsection{Solubility}
\label{sec:app-solubility-results}

\subsubsection*{Progressive alteration of a molecule}

\begin{figure}[b!]
\begin{subfigure}{0.33\linewidth}
    \centering
    \includegraphics[width=.8\linewidth]{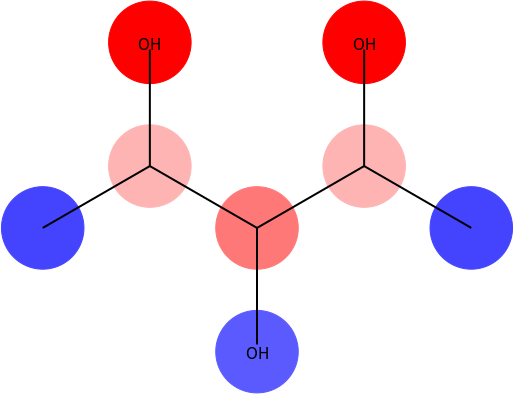}
    \vspace{-5pt}
    \caption*{0.66 \vspace{10pt}}
\end{subfigure}%
\begin{subfigure}{0.33\linewidth}
    \centering
    \includegraphics[width=.8\linewidth]{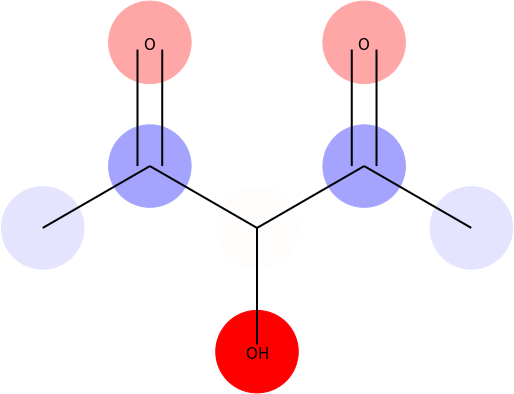}
    \vspace{-5pt}
    \caption*{0.23 \vspace{10pt}}
\end{subfigure}%
\begin{subfigure}{0.33\linewidth}
    \centering
    \includegraphics[width=.8\linewidth]{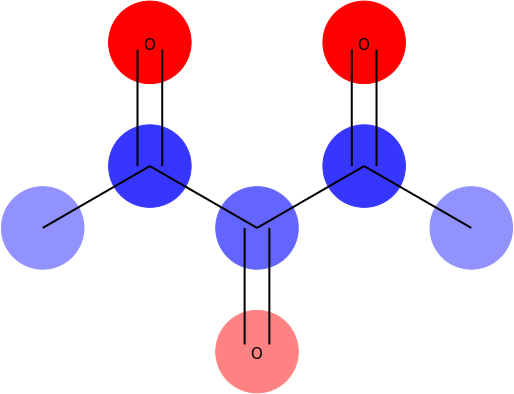}
    \vspace{-5pt}
    \caption*{-0.23 \vspace{10pt}}
\end{subfigure}\newline%
\begin{subfigure}{0.33\linewidth}
    \centering
    \includegraphics[width=.8\linewidth]{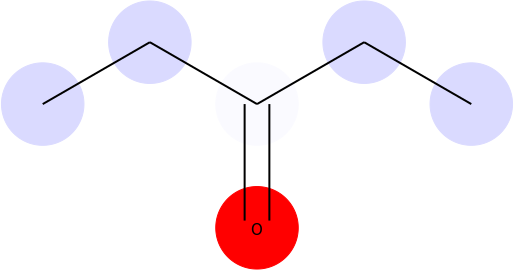}
    \vspace{-5pt}
    \caption*{-0.63 \vspace{10pt}}
\end{subfigure}%
\begin{subfigure}{0.33\linewidth}
    \centering
    \includegraphics[width=.8\linewidth]{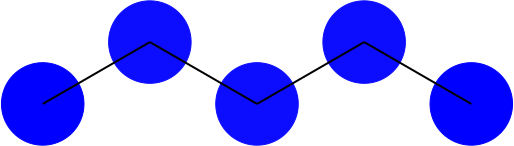}
    \vspace{-5pt}
    \caption*{-2.4 \vspace{10pt}}
\end{subfigure}\hfill%
\caption{\textbf{Progressively removing molecular elements that are important for solubility.} {\small The top-left molecule is rendered progressively less soluble by a) turning single C-O bonds into double C=O bonds and b) removing oxygen atoms altogether (left-to-right, top-to-bottom). Such modifications result in chemically valid molecules and are taken according the importance attributed via LRP (red = positive relevance). The predicted solubility in log mol/L is reported under each molecule.}}
\label{fig:appendix-molecule-progressive}
\end{figure}

As mentioned in the discussion (Sec.~\ref{sec:discussions}), choosing to model molecules as graphs yields a  non-redundant and highly structured representation. As a consequence, the ability to slightly alter a molecule by performing small steps in Euclidean space is lost. This makes it hard to verify that explanations correspond to how the trained network predicts solubility. In fact, it is not possible to automatically alter a molecule according to the importance of its atoms/bonds and still obtain a valid molecule.
In this case, it is necessary to apply domain-specific knowledge and identify which changes are viable in the space of valid molecules. In Figure \ref{fig:appendix-molecule-progressive} we show a trivial example where we use LRP to identify important atoms/bonds of a molecule and progressively remove them to reduce the predicted solubility.

\subsubsection*{Solubility explanations}
\begin{figure*}[h!]
\begin{subfigure}{\linewidth}
    \centering
    \includegraphics[width=.25\linewidth]{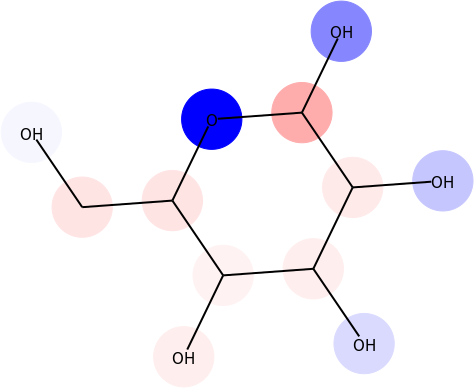}
\end{subfigure}\hfill\newline%
\begin{subfigure}{\linewidth}
    \centering
    \includegraphics[width=1.\linewidth]{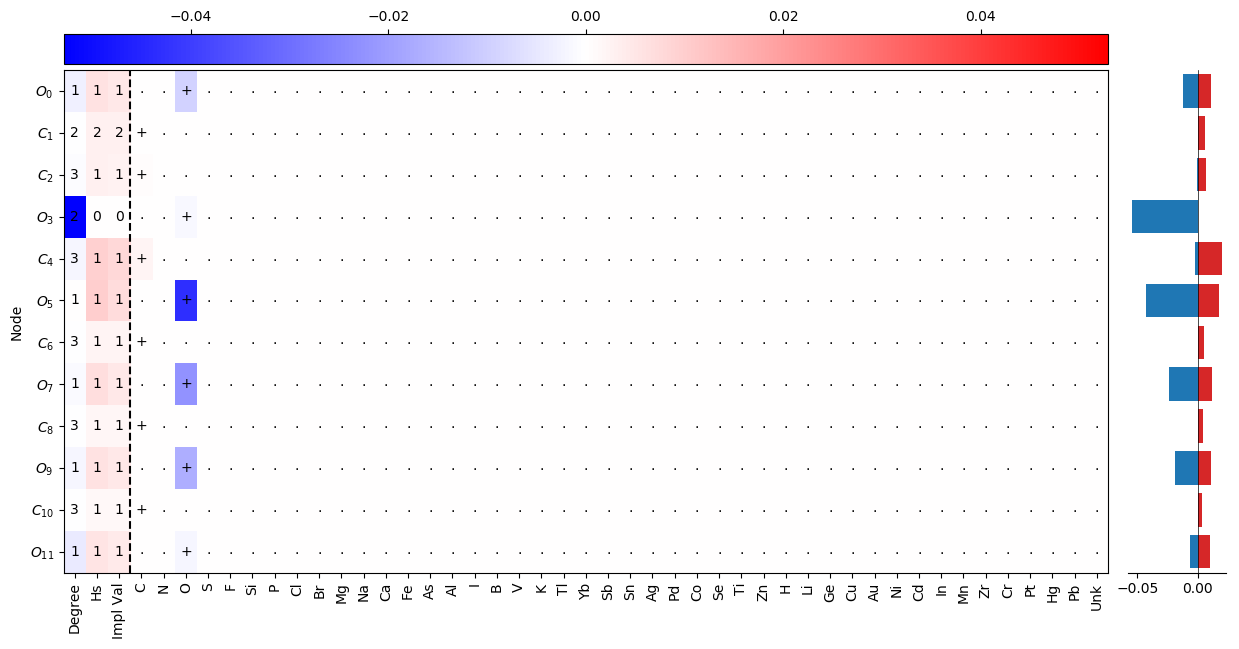}
\end{subfigure}\newline%
\begin{subfigure}{\linewidth}
    \centering
    \includegraphics[width=1.\linewidth]{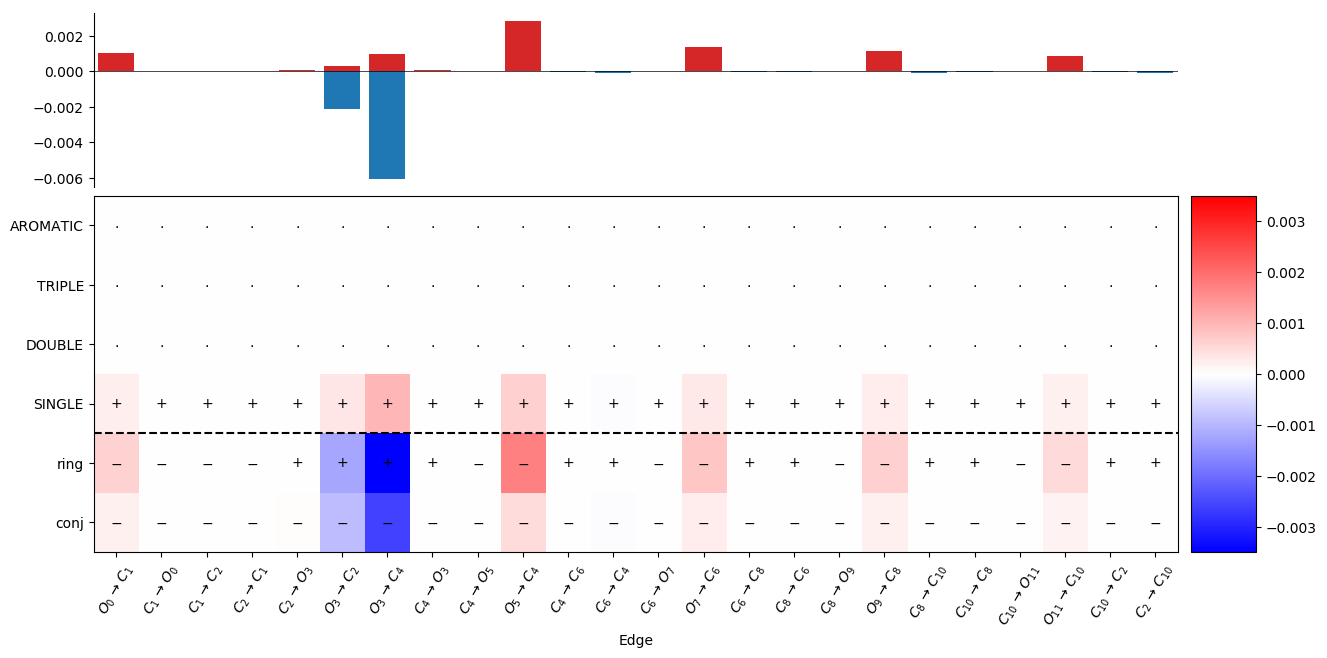}
\end{subfigure}%
\caption{\textbf{Solubility of glucose:} predicted 0.74 log mol/L, measured 0.67 log mol/L, explanation over the atom and bond features using Layer-wise Relevance Propagation. Note how the main negative contribution is assigned to the oxygen atom that renders the ring heterocyclic.}
\label{fig:appendix-glucose}
\end{figure*}

\begin{figure*}[h!]
\begin{subfigure}{\linewidth}
    \centering
    \includegraphics[width=.4\linewidth]{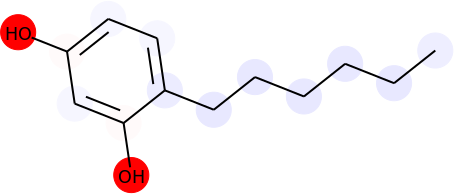}
\end{subfigure}\hfill\newline%
\begin{subfigure}{\linewidth}
    \centering
    \includegraphics[width=1.\linewidth]{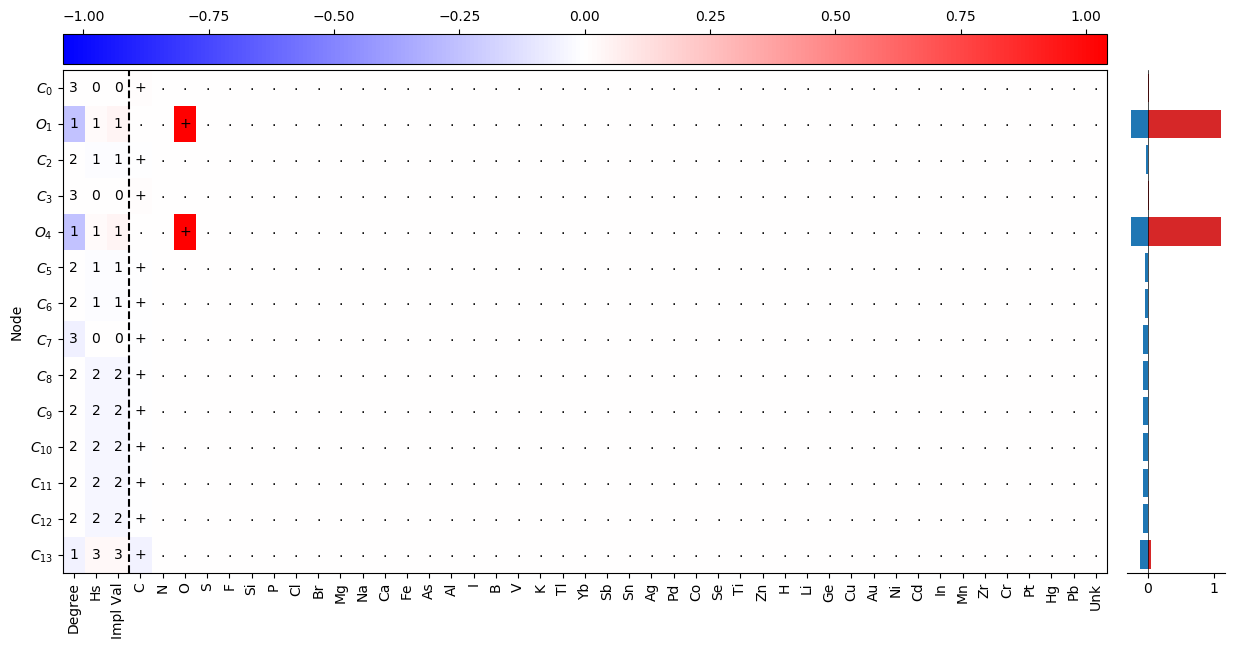}
\end{subfigure}\newline%
\begin{subfigure}{\linewidth}
    \centering
    \includegraphics[width=1.\linewidth]{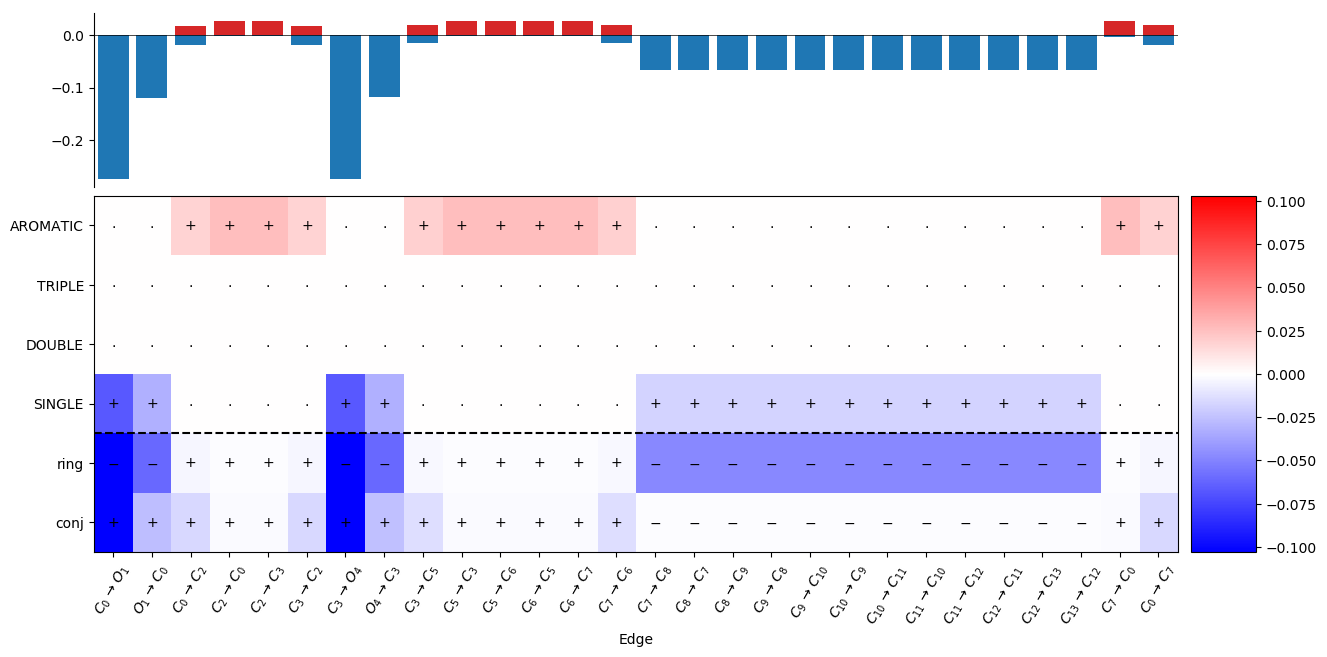}
\end{subfigure}%
\caption{\textbf{Solubility of 4-hexylresorcinol:} predicted -2.59 log mol/L, measured -2.54log mol/L, explanation over the atom and bond features using Layer-wise Relevance Propagation. The explanation attributes negative relevance to the long hydrocarbon chain and positive relevance to the OH groups linked to the aromatic ring, which is in agreement with common knowledge in organic chemistry.}
\label{fig:appendix-4-hexylresorcinol}
\end{figure*}

In figures \ref{fig:appendix-glucose} and \ref{fig:appendix-4-hexylresorcinol} we present a visualization of the explanation for the predicted solubility of glucose (moderately soluble) and 4-hexylresorcinol (moderately insoluble). The explanations are produced by applying LRP and propagating positive and negative relevance to the individual features of the atoms/bonds. 

\clearpage

\FloatBarrier
\clearpage

\clearpage

\clearpage

\clearpage

\begin{figure*}[h!]
\begin{subfigure}{.6\linewidth}
    \includegraphics[width=1.\linewidth]{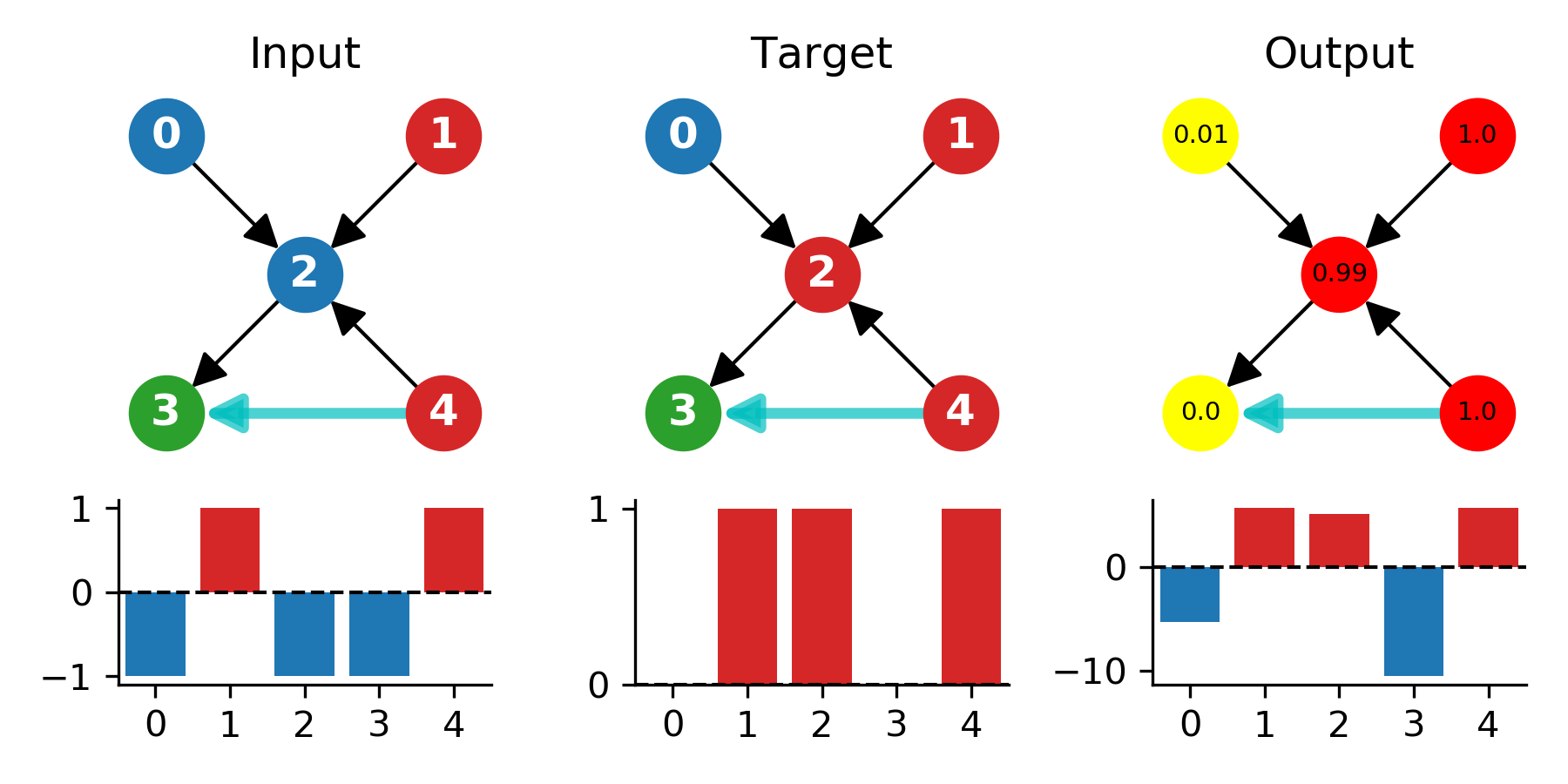}
\end{subfigure}\hfill\newline%
\begin{subfigure}{0.44\linewidth}
    \centering
    \includegraphics[width=1.\linewidth]{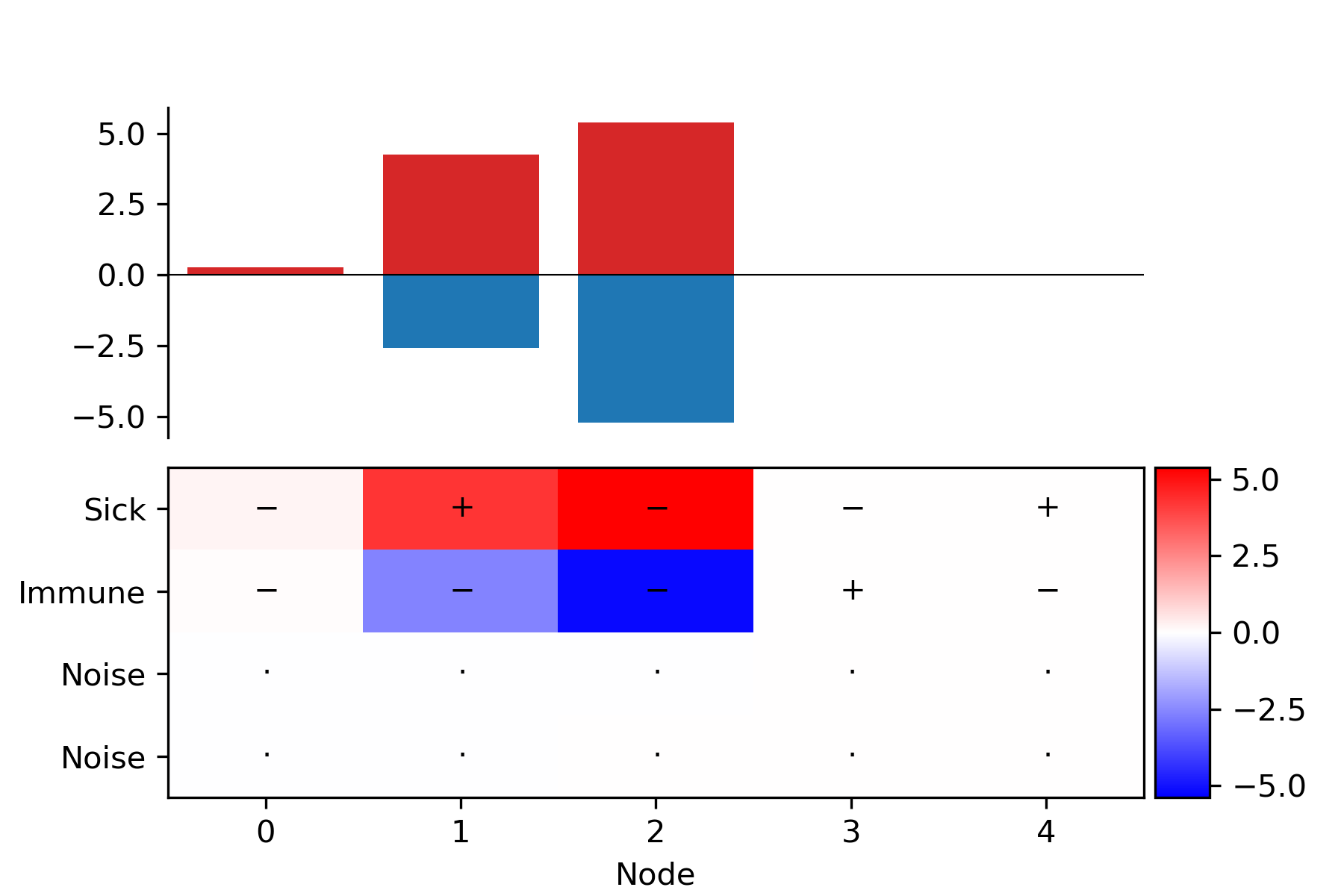}
\end{subfigure}%
\begin{subfigure}{0.44\linewidth}
    \centering
    \includegraphics[width=1.\linewidth]{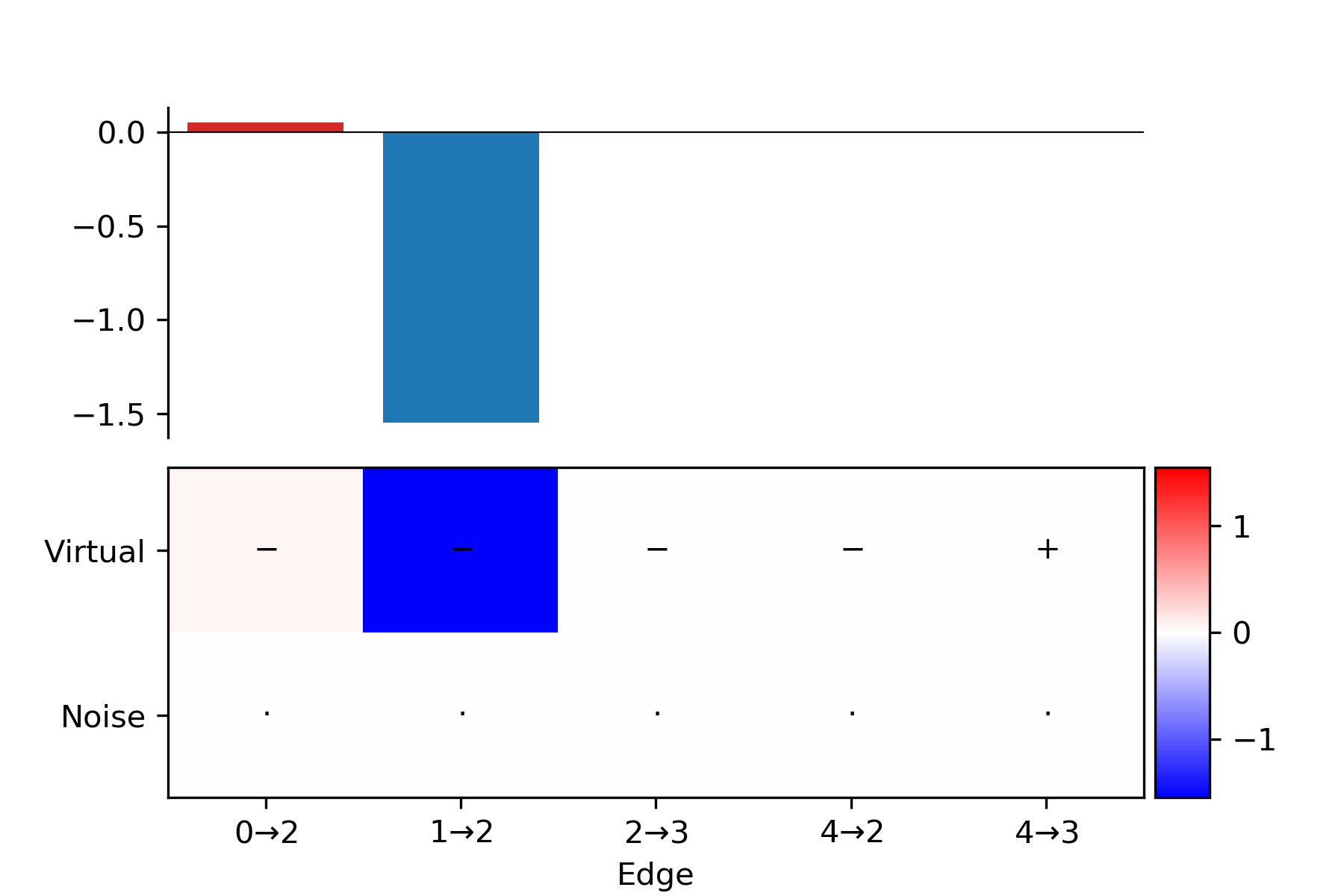}
\end{subfigure}%
\begin{subfigure}{0.11\linewidth}
    \includegraphics[width=1.\linewidth]{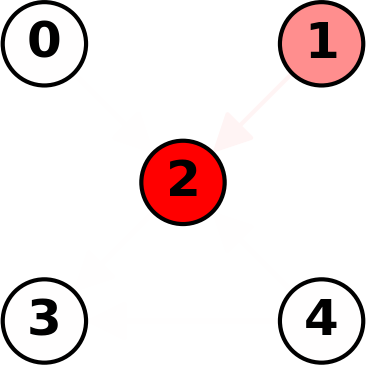}
    \caption*{SA}
\end{subfigure}\newline%
\begin{subfigure}{0.44\linewidth}
    \centering
    \includegraphics[width=1.\linewidth]{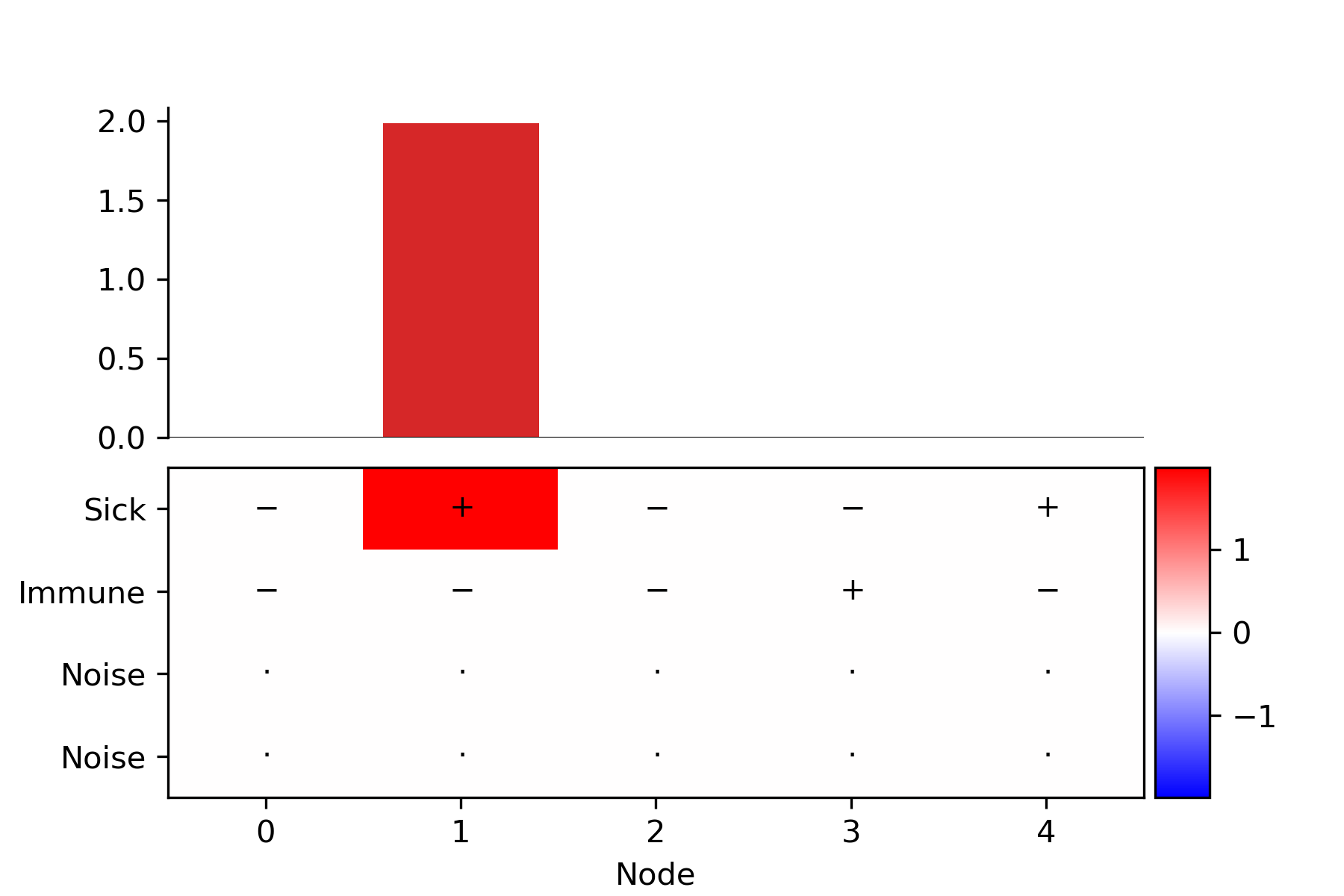}
\end{subfigure}%
\begin{subfigure}{0.44\linewidth}
    \centering
    \includegraphics[width=1.\linewidth]{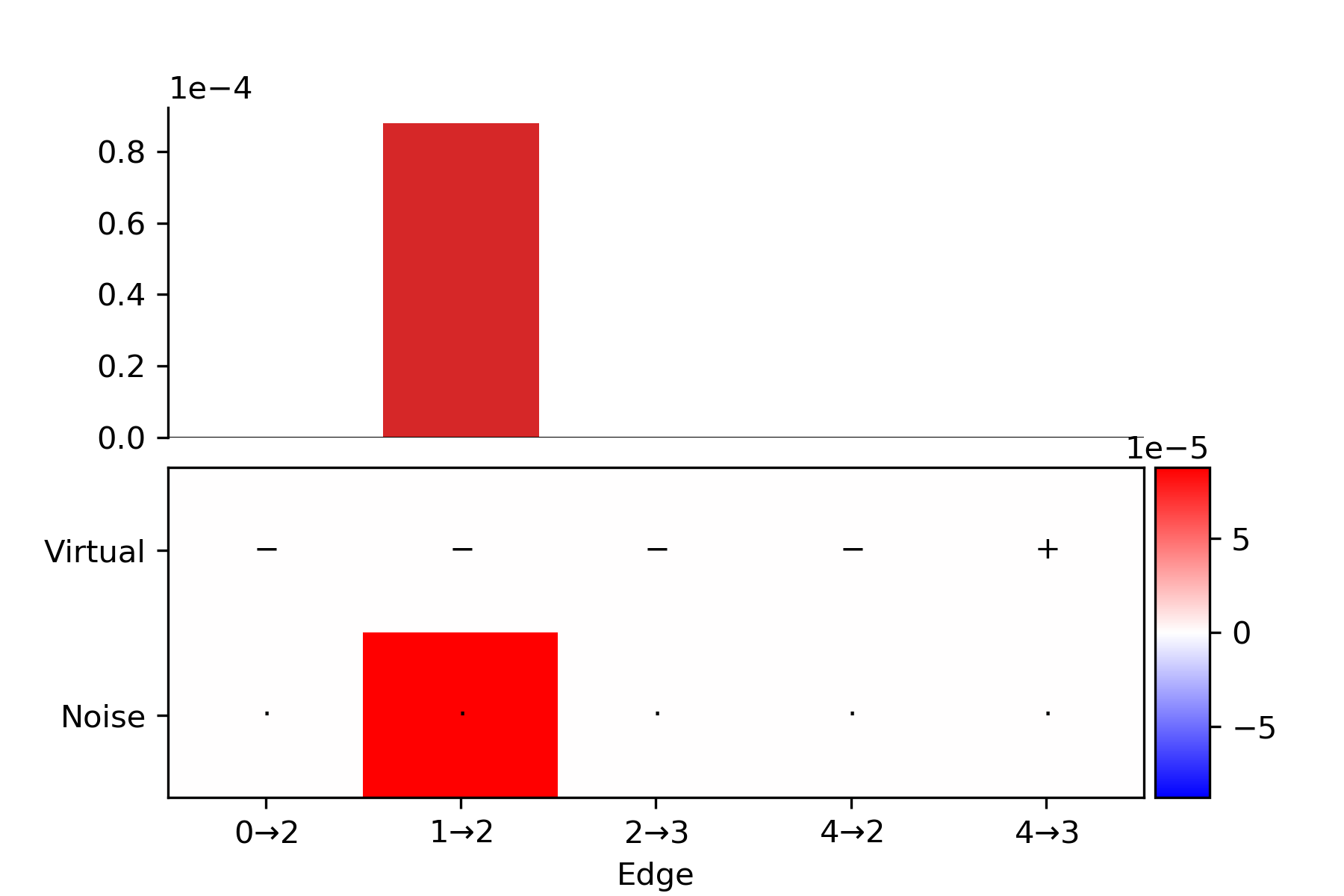}
\end{subfigure}%
\begin{subfigure}{0.11\linewidth}
    \includegraphics[width=1.\linewidth]{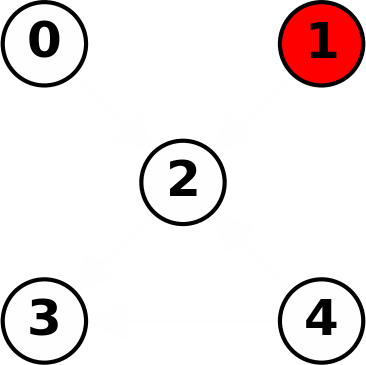}
    \caption*{GBP}
\end{subfigure}\newline%
\begin{subfigure}{0.44\linewidth}
    \centering
    \includegraphics[width=1.\linewidth]{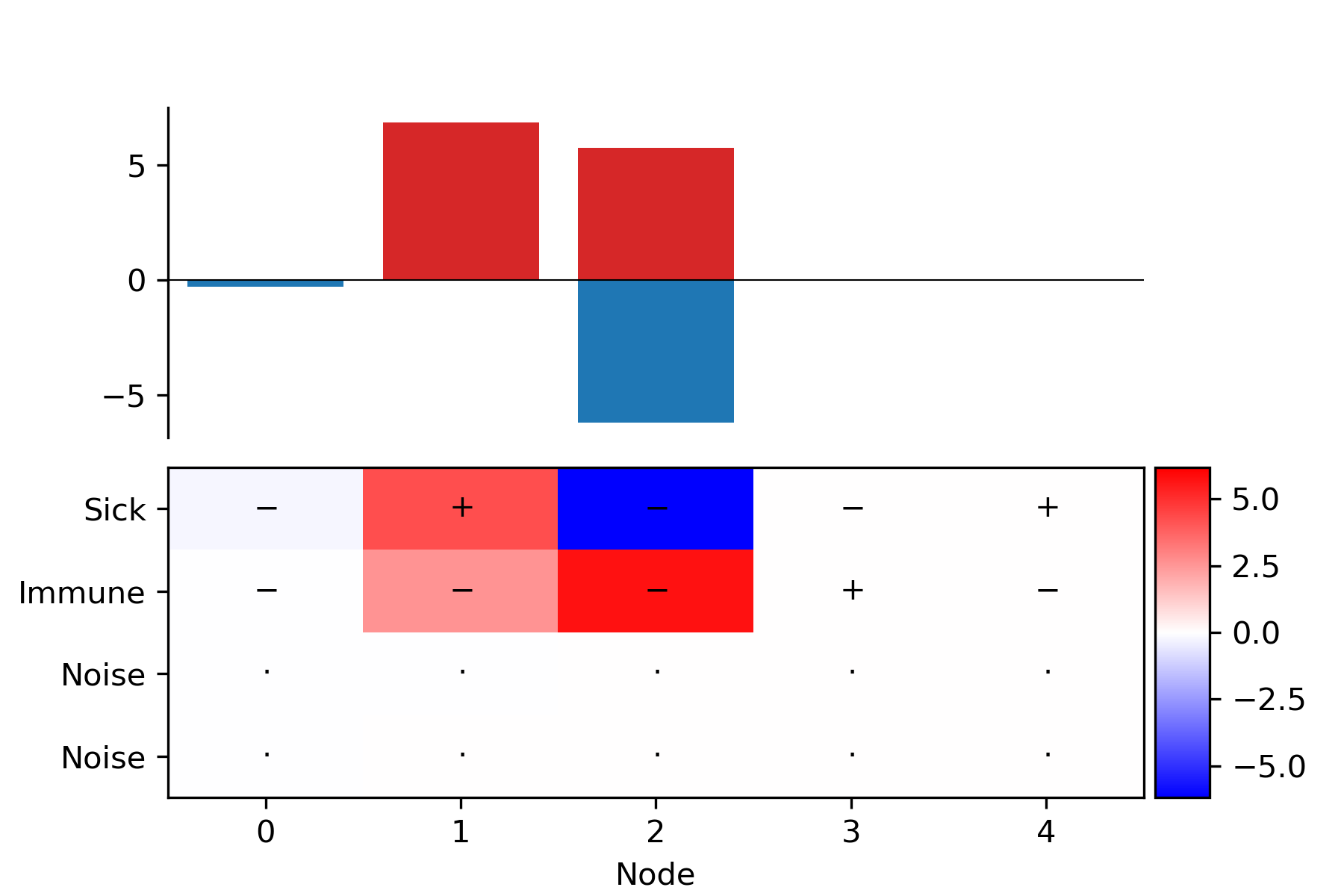}
\end{subfigure}%
\begin{subfigure}{0.44\linewidth}
    \centering
    \includegraphics[width=1.\linewidth]{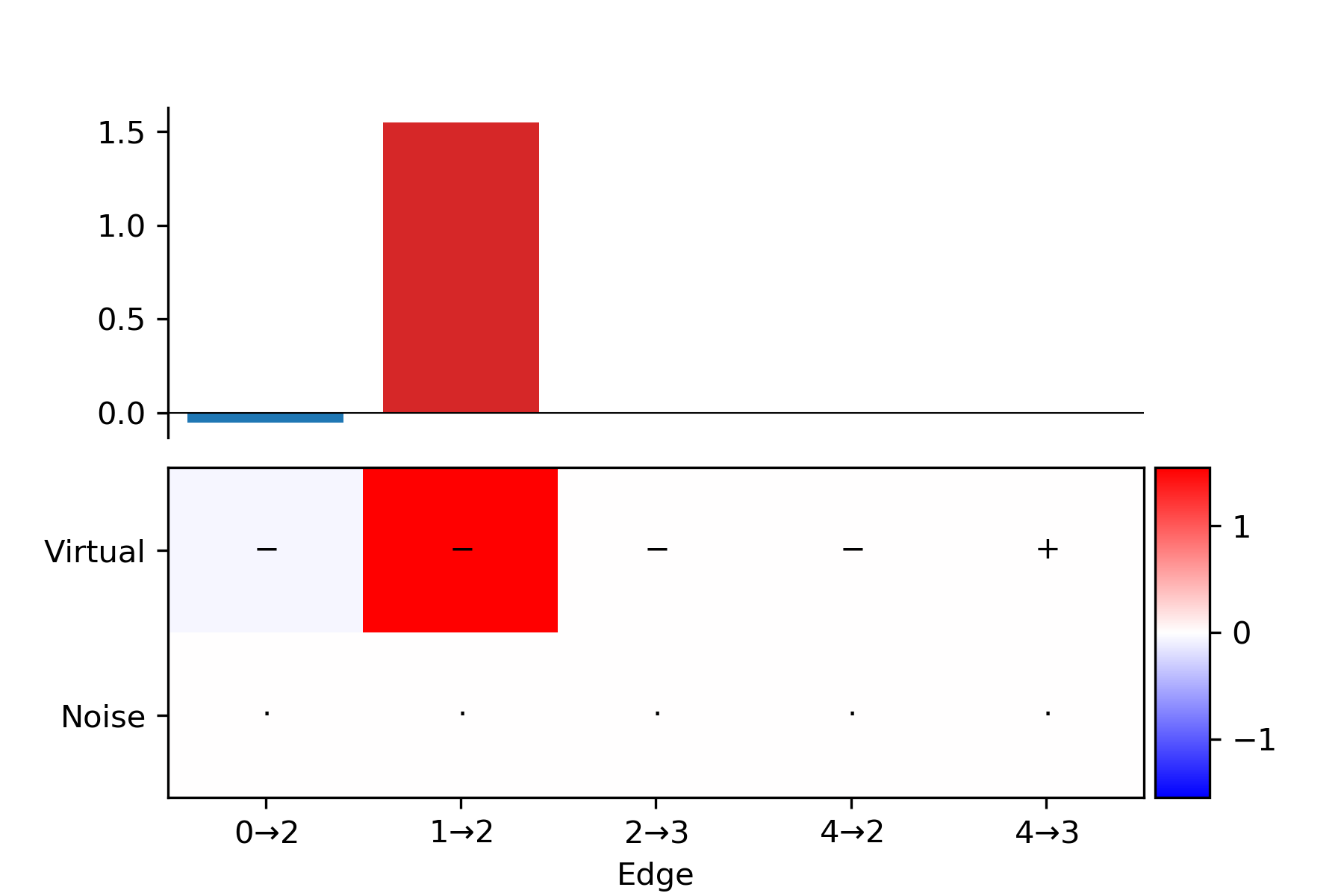}
\end{subfigure}%
\begin{subfigure}{0.11\linewidth}
    \includegraphics[width=1.\linewidth]{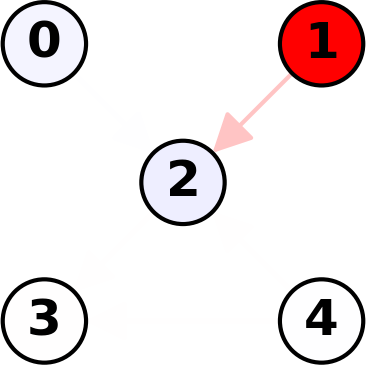}
    \caption*{LRP}
\end{subfigure}%
\caption{\textbf{Graph 2 - Max pooling.} The healthy central node, on which the explanation is focused, has two neighbors that are sick. All methods are only only able to identify one neighbor as a possible explanation for the network's prediction, due to the fact that max pooling is used to aggregate incoming features. This example is similar to Fig.~\ref{fig:max-vs-sum} in the main text.}
\end{figure*}\clearpage

\begin{figure*}[h!]
\begin{subfigure}{.6\linewidth}
    \includegraphics[width=1.\linewidth]{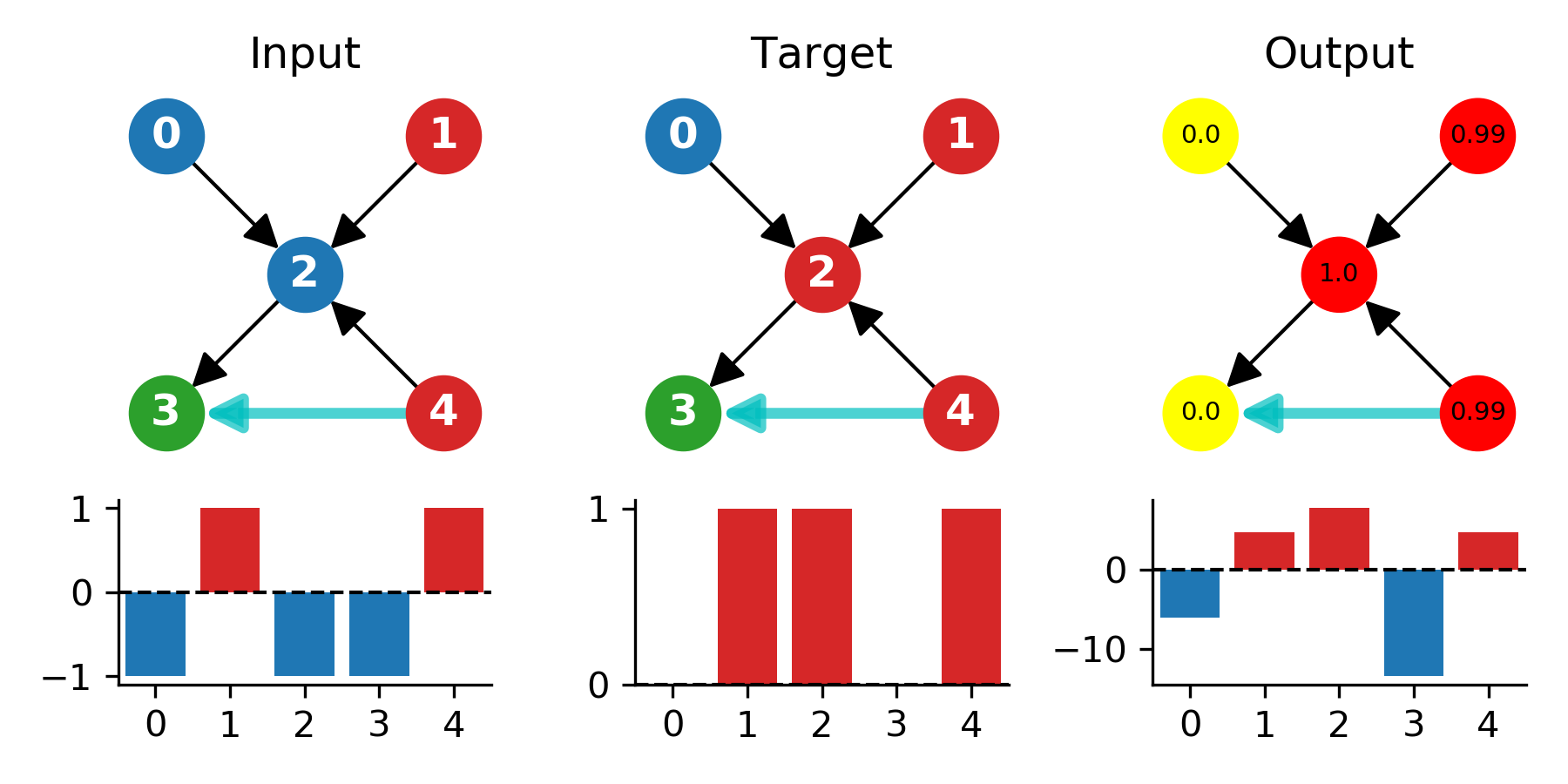}
\end{subfigure}\hfill\newline%
\begin{subfigure}{0.44\linewidth}
    \centering
    \includegraphics[width=1.\linewidth]{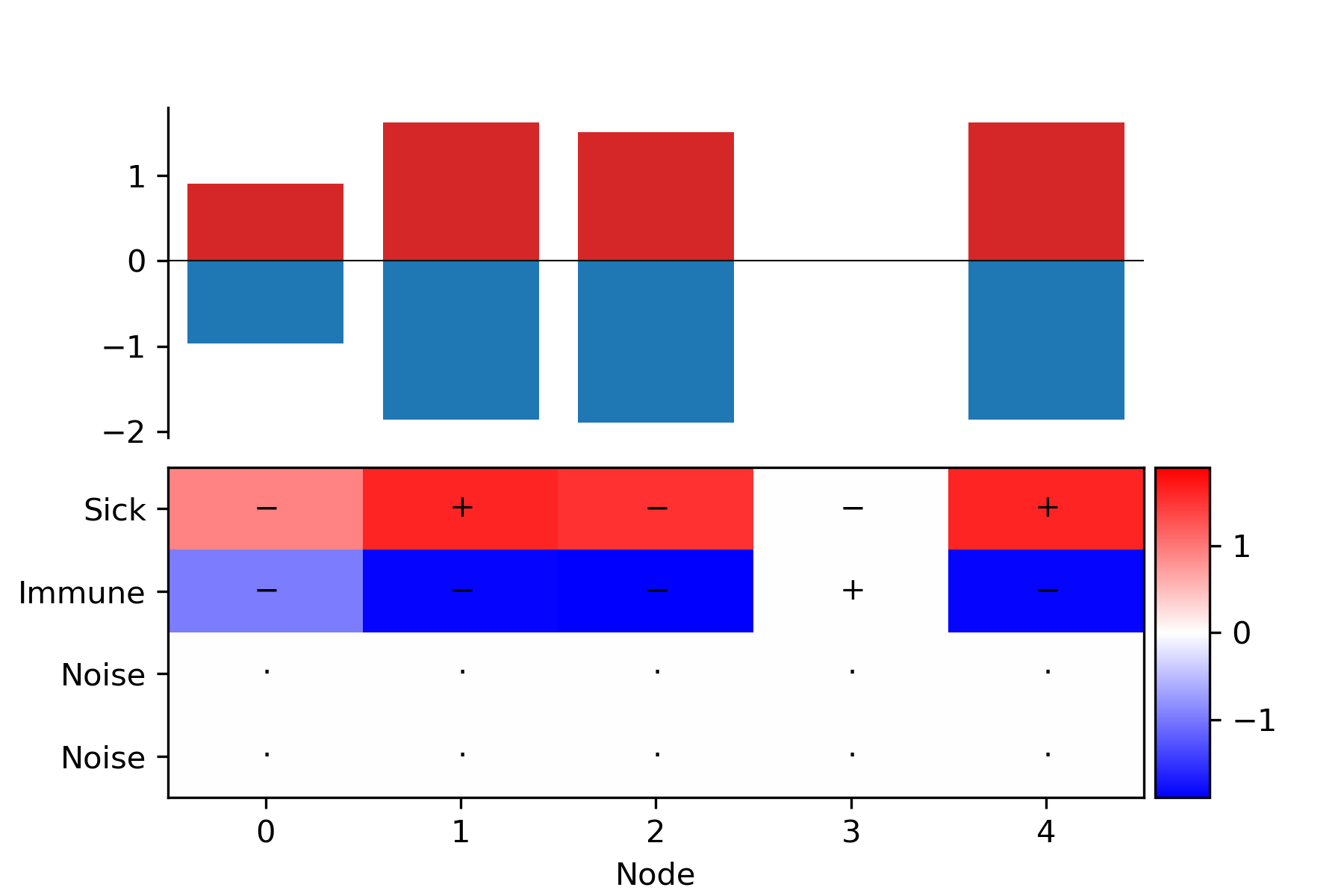}
\end{subfigure}%
\begin{subfigure}{0.44\linewidth}
    \centering
    \includegraphics[width=1.\linewidth]{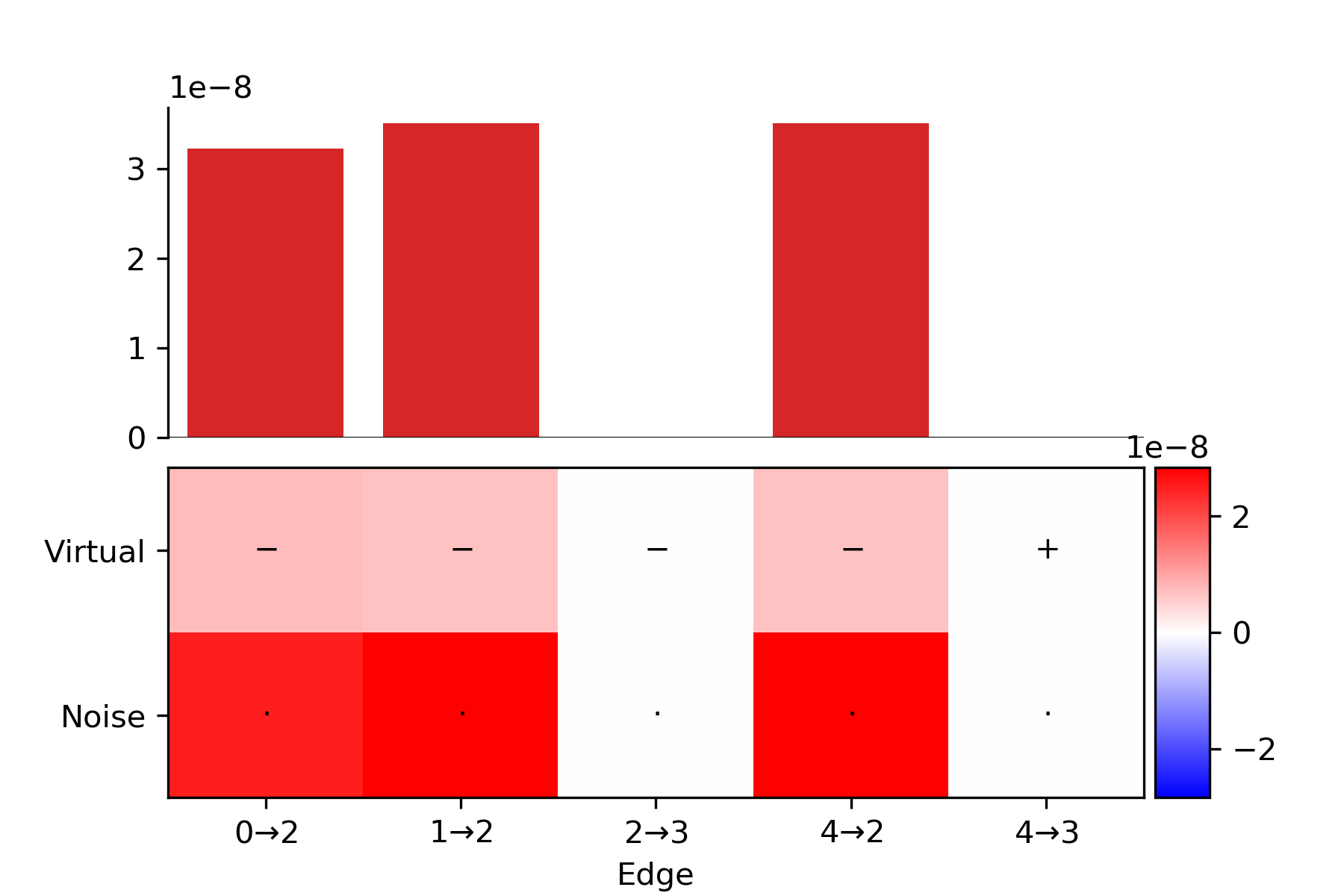}
\end{subfigure}%
\begin{subfigure}{0.11\linewidth}
    \includegraphics[width=1.\linewidth]{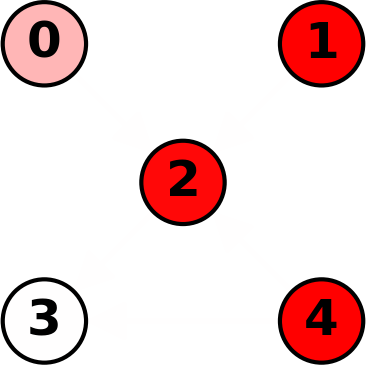}
    \caption*{SA}
\end{subfigure}\newline%
\begin{subfigure}{0.44\linewidth}
    \centering
    \includegraphics[width=1.\linewidth]{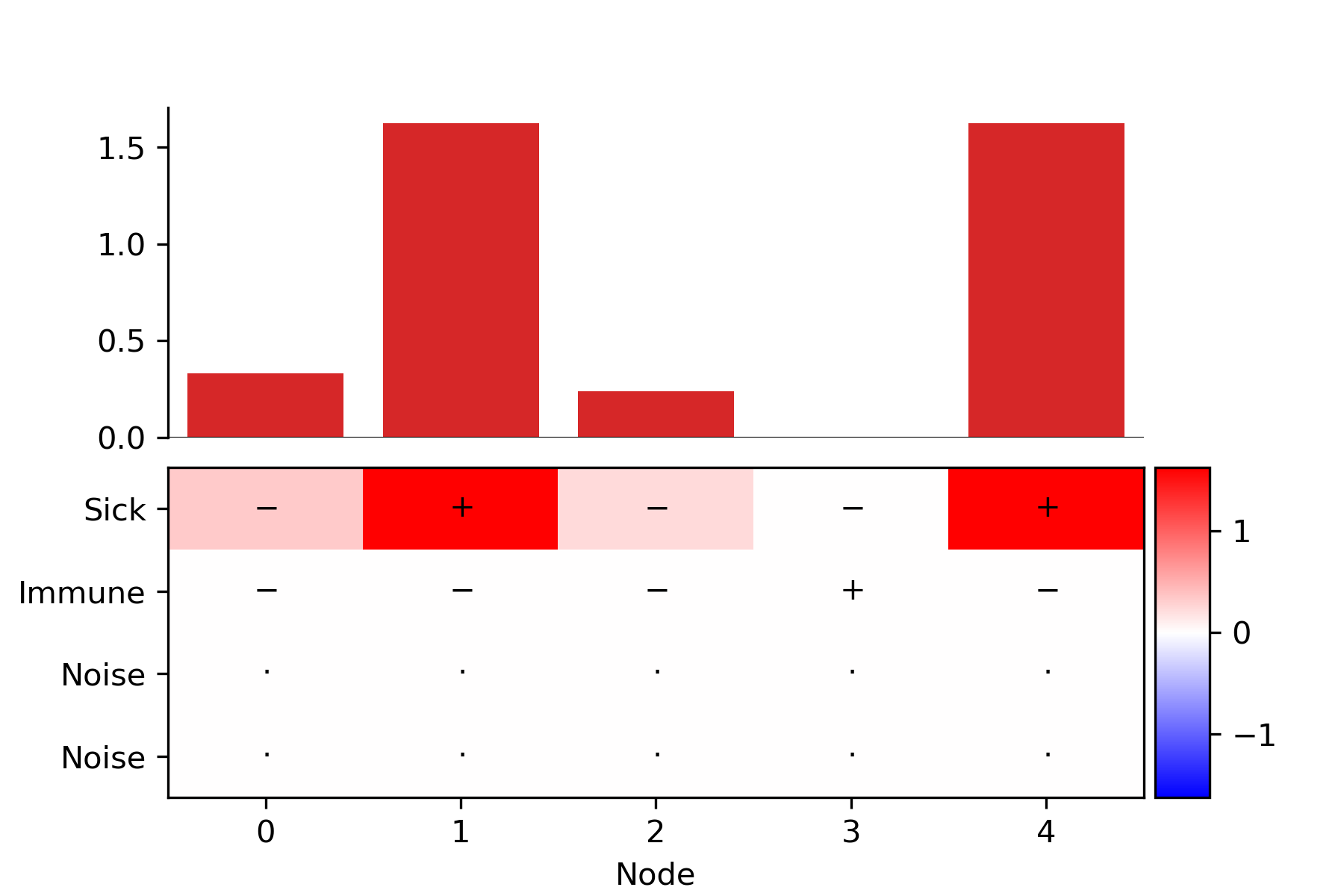}
\end{subfigure}%
\begin{subfigure}{0.44\linewidth}
    \centering
    \includegraphics[width=1.\linewidth]{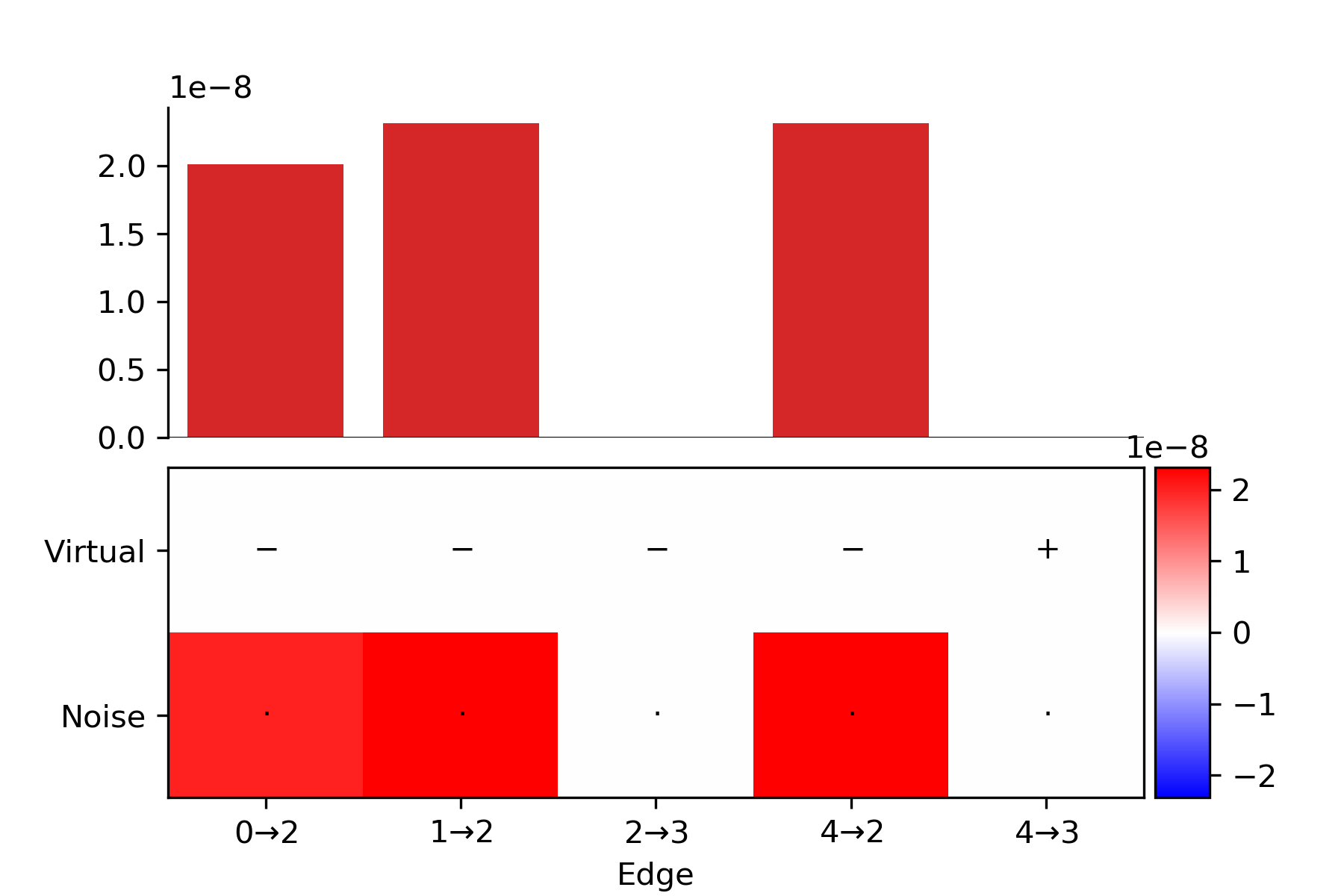}
\end{subfigure}%
\begin{subfigure}{0.11\linewidth}
    \includegraphics[width=1.\linewidth]{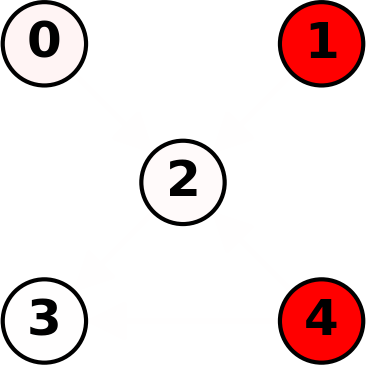}
    \caption*{GBP}
\end{subfigure}\newline%
\begin{subfigure}{0.44\linewidth}
    \centering
    \includegraphics[width=1.\linewidth]{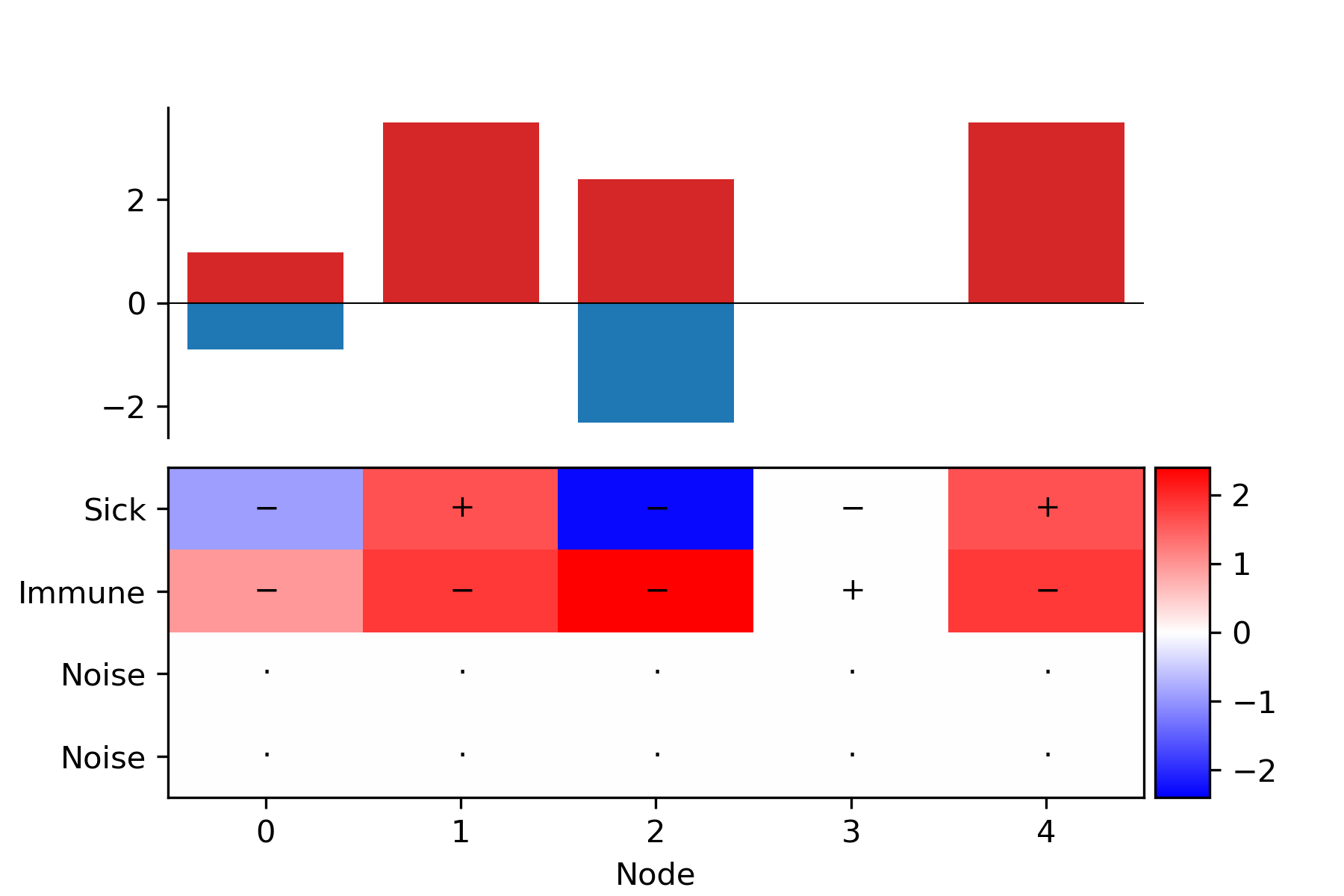}
\end{subfigure}%
\begin{subfigure}{0.44\linewidth}
    \centering
    \includegraphics[width=1.\linewidth]{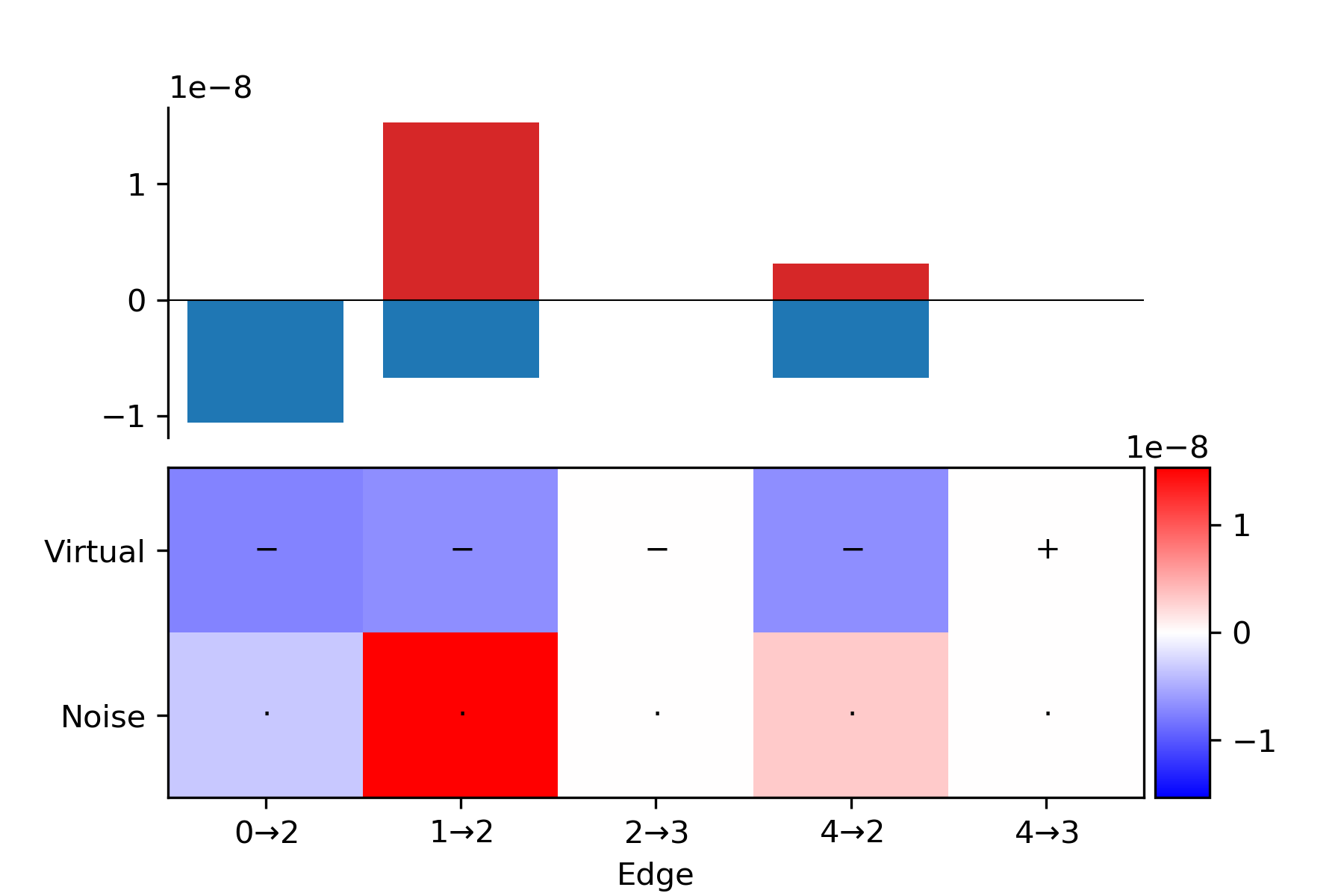}
\end{subfigure}%
\begin{subfigure}{0.11\linewidth}
    \includegraphics[width=1.\linewidth]{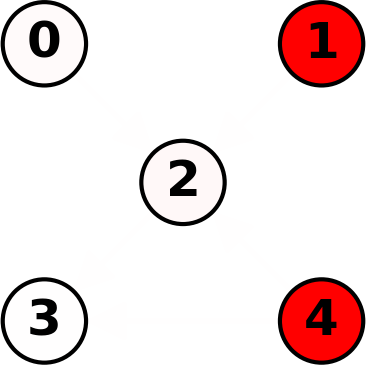}
    \caption*{LRP}
\end{subfigure}%
\caption{\textbf{Graph 2 - Sum pooling.} Compared to the max pooling version, the network that uses sum pooling produces explanations that include all possible sources of infection. However, sum pooling might result in incorrect predictions, as demonstrated in Fig.~\ref{fig:appendix-graph-5-sum}. This example is similar to Fig.~\ref{fig:max-vs-sum} in the main text.}
\end{figure*}\clearpage

\begin{figure*}[h!]
\begin{subfigure}{.6\linewidth}
    \includegraphics[width=1.\linewidth]{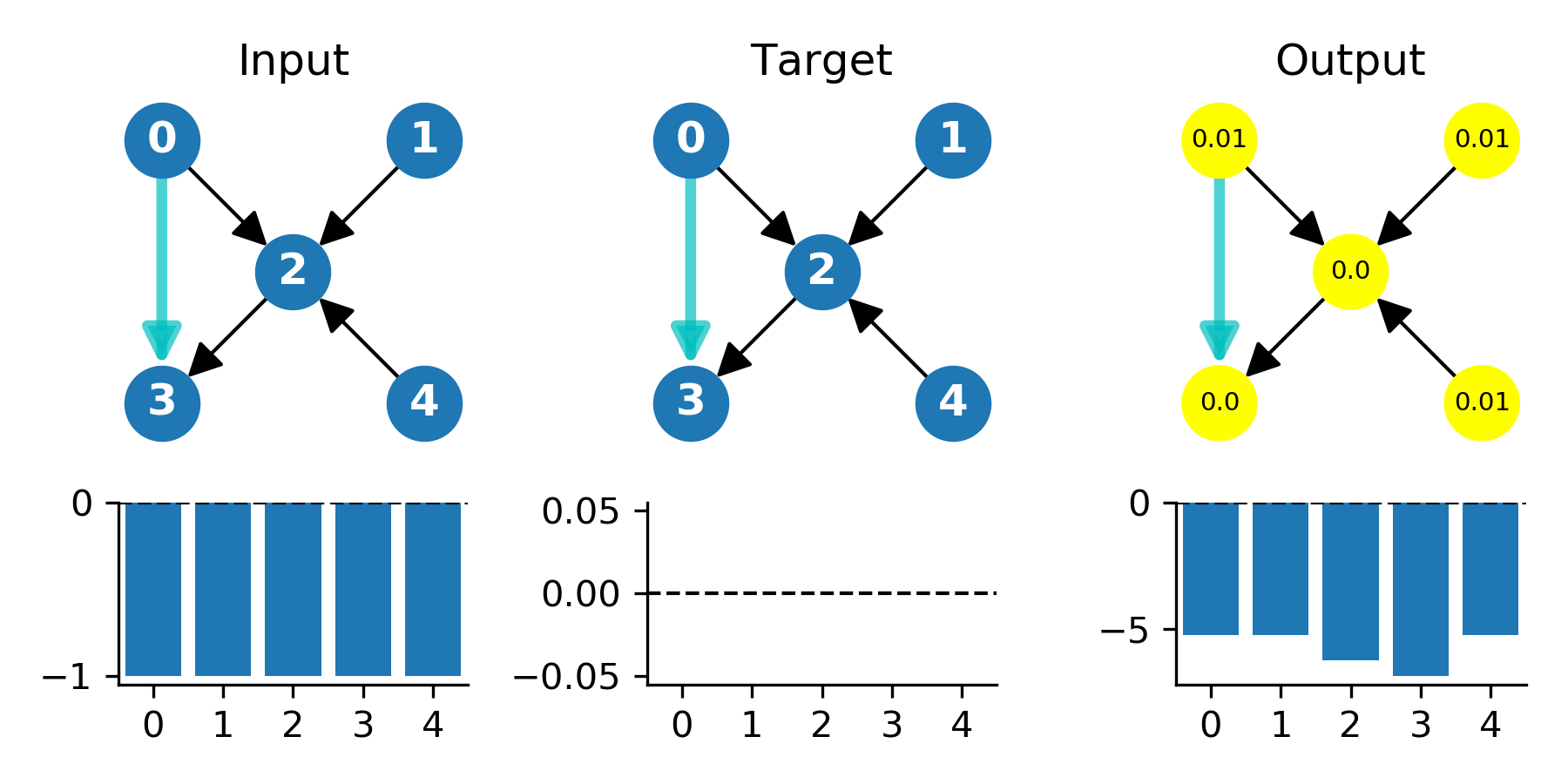}
\end{subfigure}\hfill\newline%
\begin{subfigure}{0.44\linewidth}
    \centering
    \includegraphics[width=1.\linewidth]{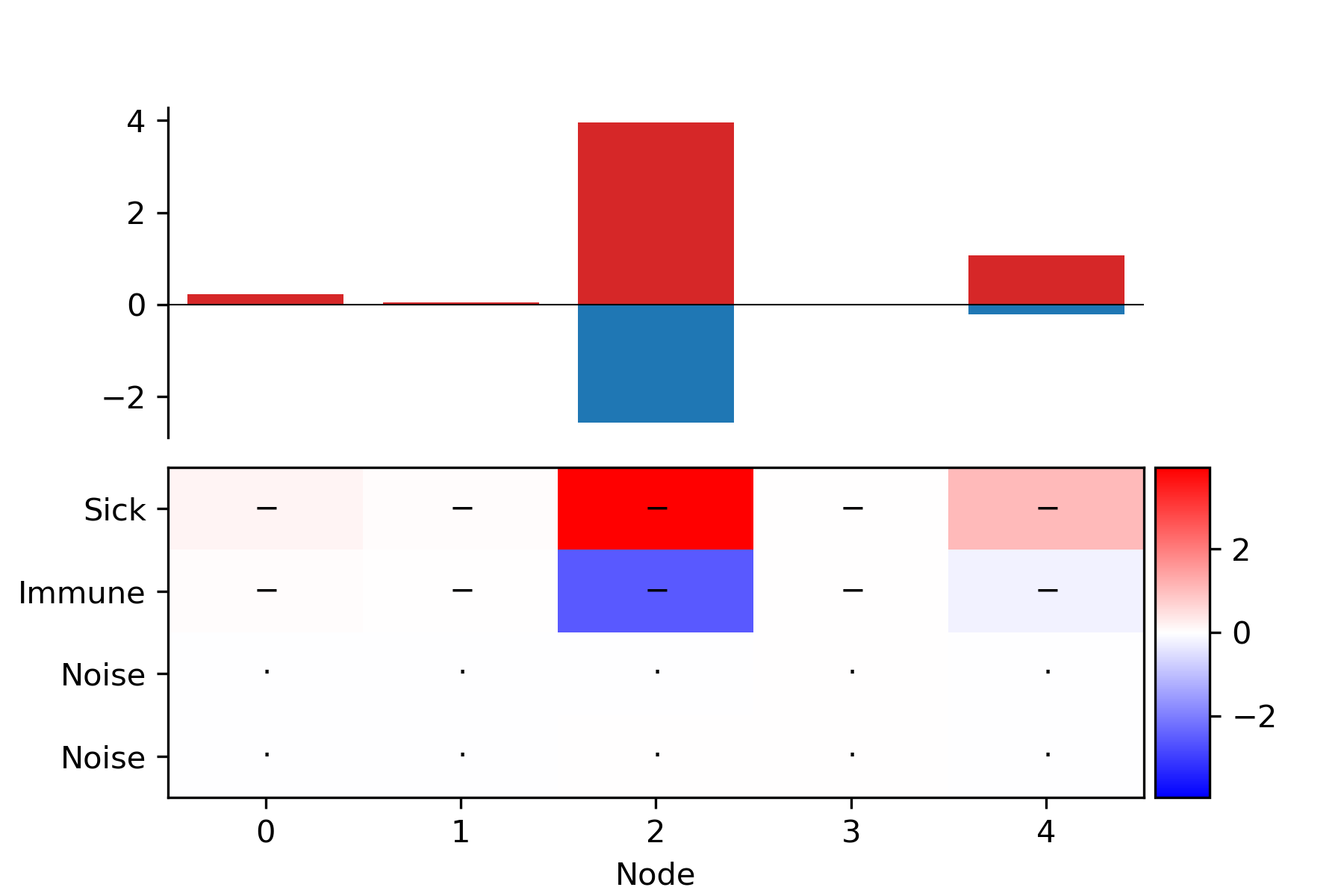}
\end{subfigure}%
\begin{subfigure}{0.44\linewidth}
    \centering
    \includegraphics[width=1.\linewidth]{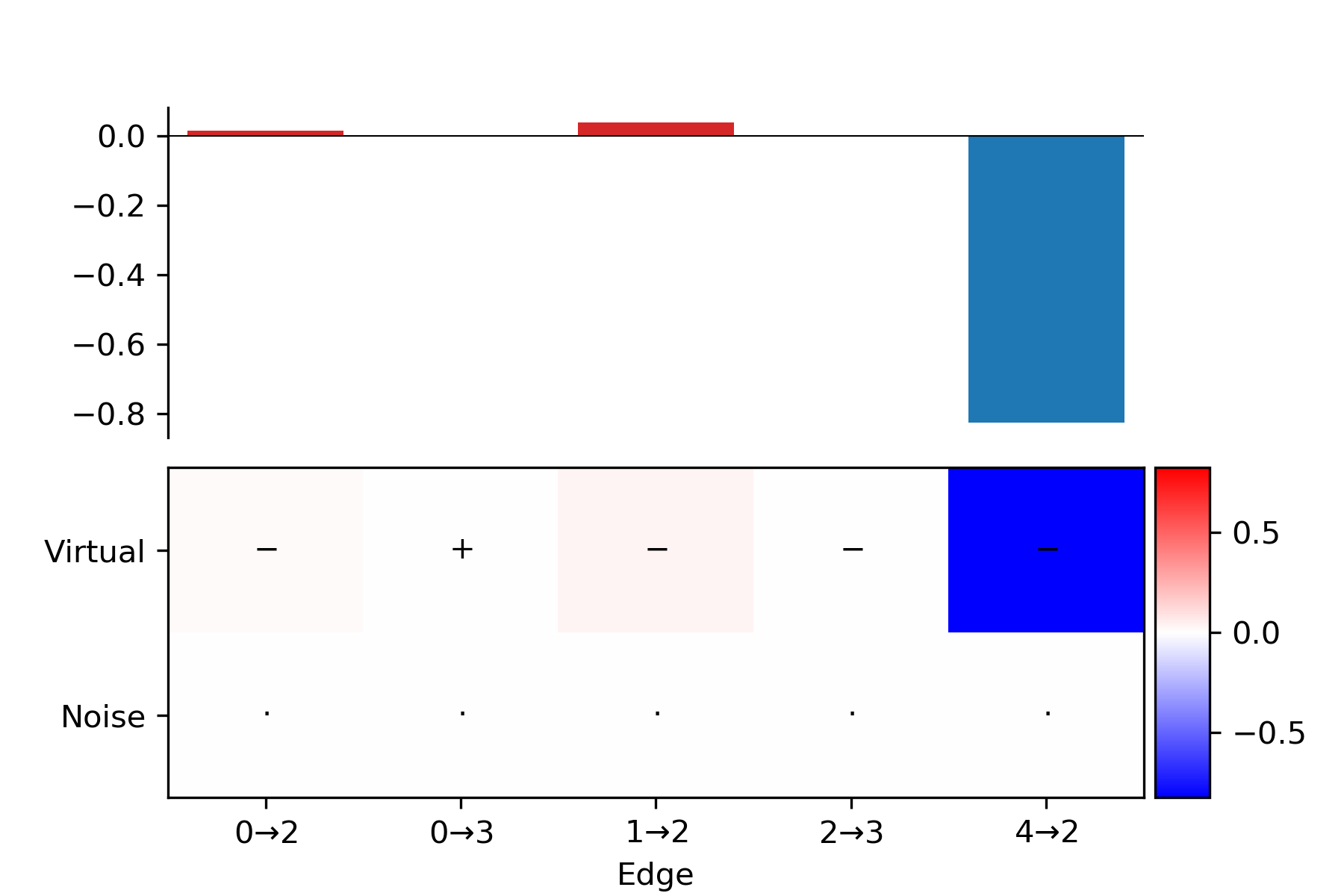}
\end{subfigure}%
\begin{subfigure}{0.11\linewidth}
    \includegraphics[width=1.\linewidth]{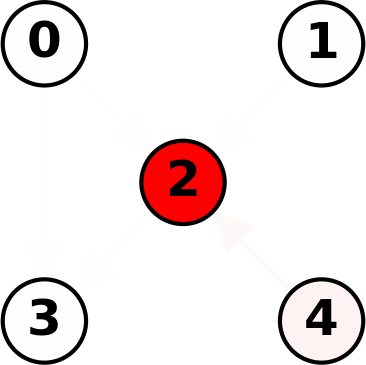}
    \caption*{SA}
\end{subfigure}\newline%
\begin{subfigure}{0.44\linewidth}
    \centering
    \includegraphics[width=1.\linewidth]{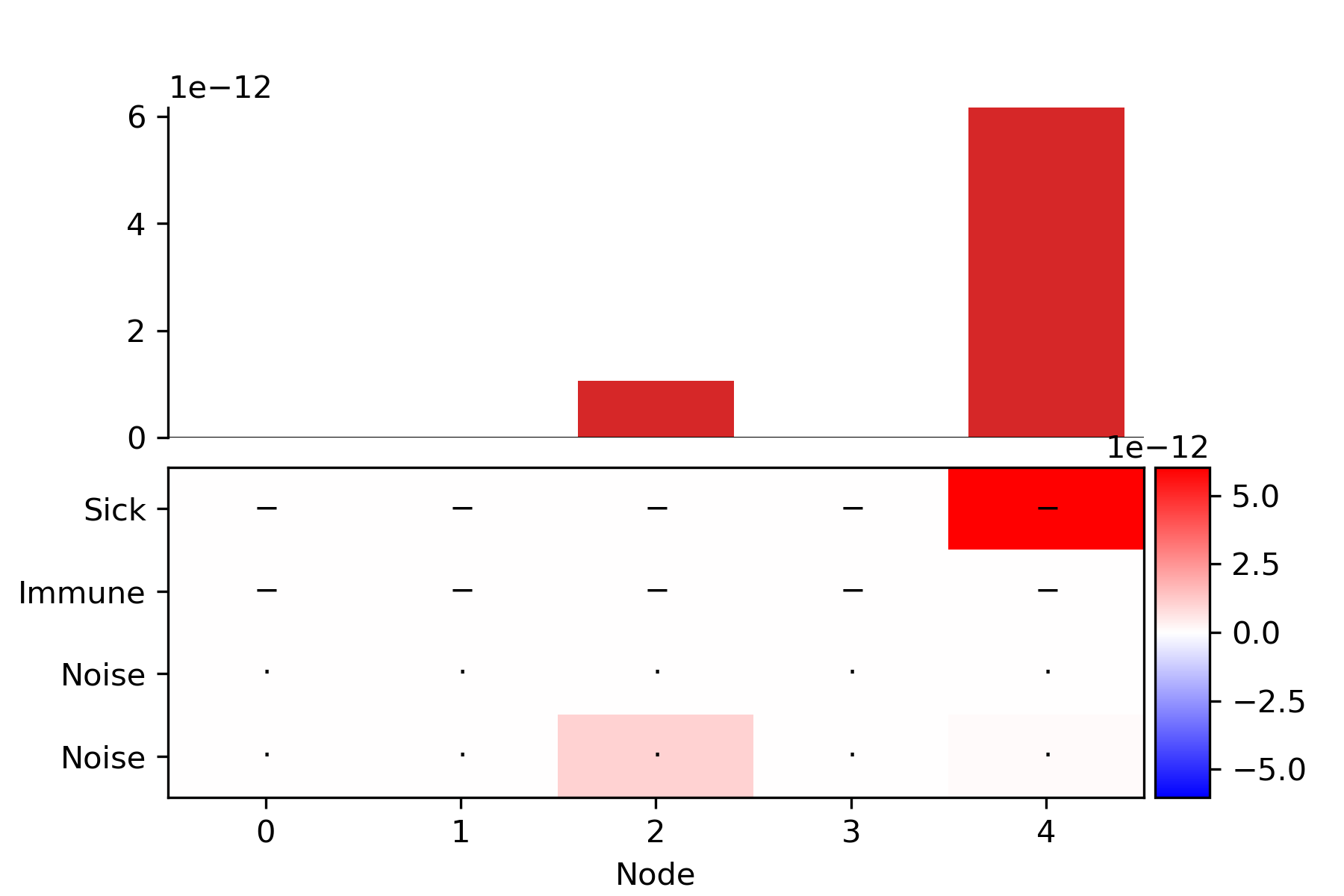}
\end{subfigure}%
\begin{subfigure}{0.44\linewidth}
    \centering
    \includegraphics[width=1.\linewidth]{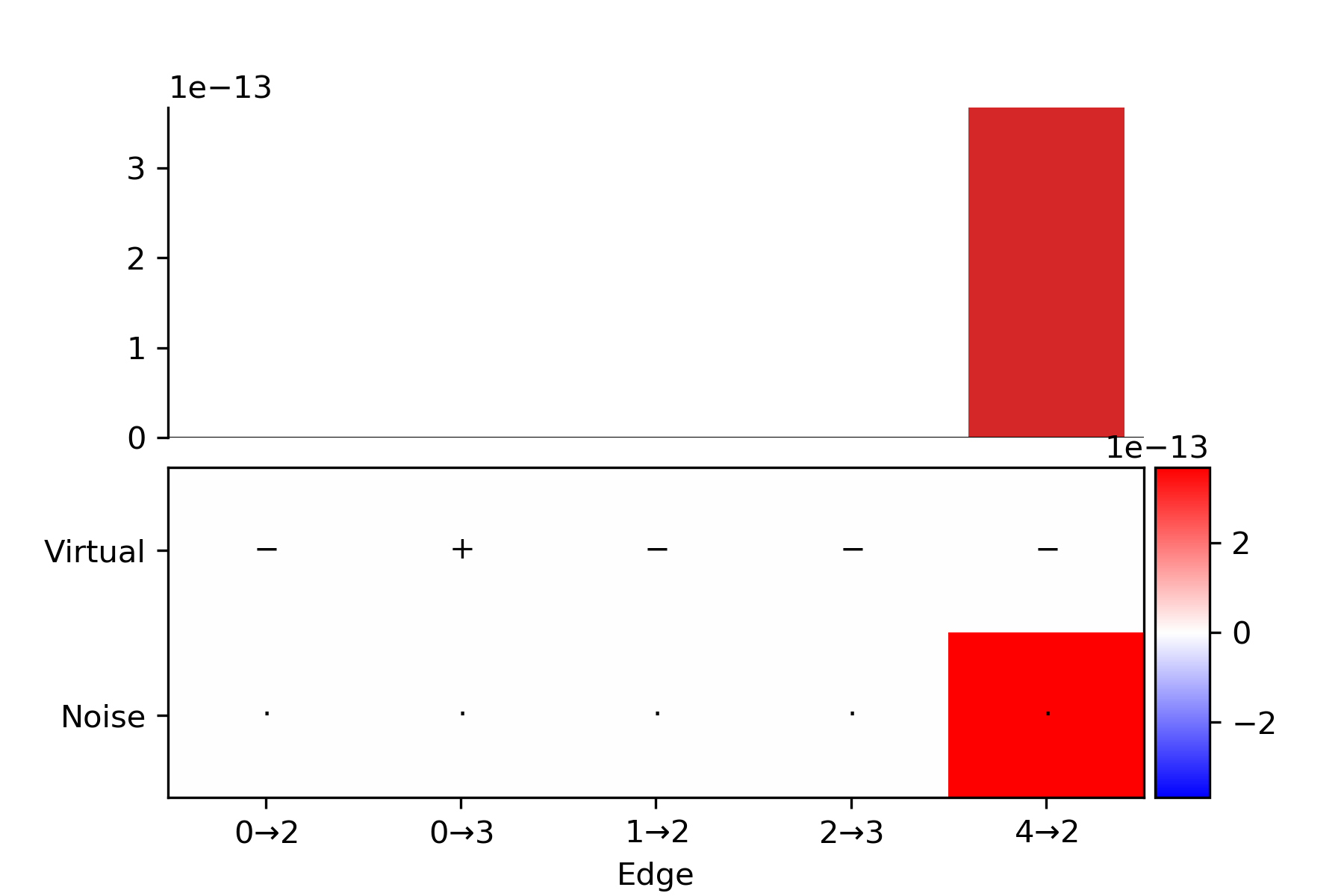}
\end{subfigure}%
\begin{subfigure}{0.11\linewidth}
    \includegraphics[width=1.\linewidth]{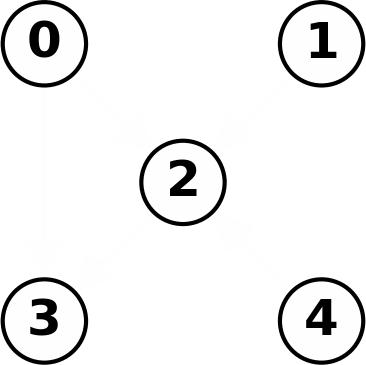}
    \caption*{GBP}
\end{subfigure}\newline%
\begin{subfigure}{0.44\linewidth}
    \centering
    \includegraphics[width=1.\linewidth]{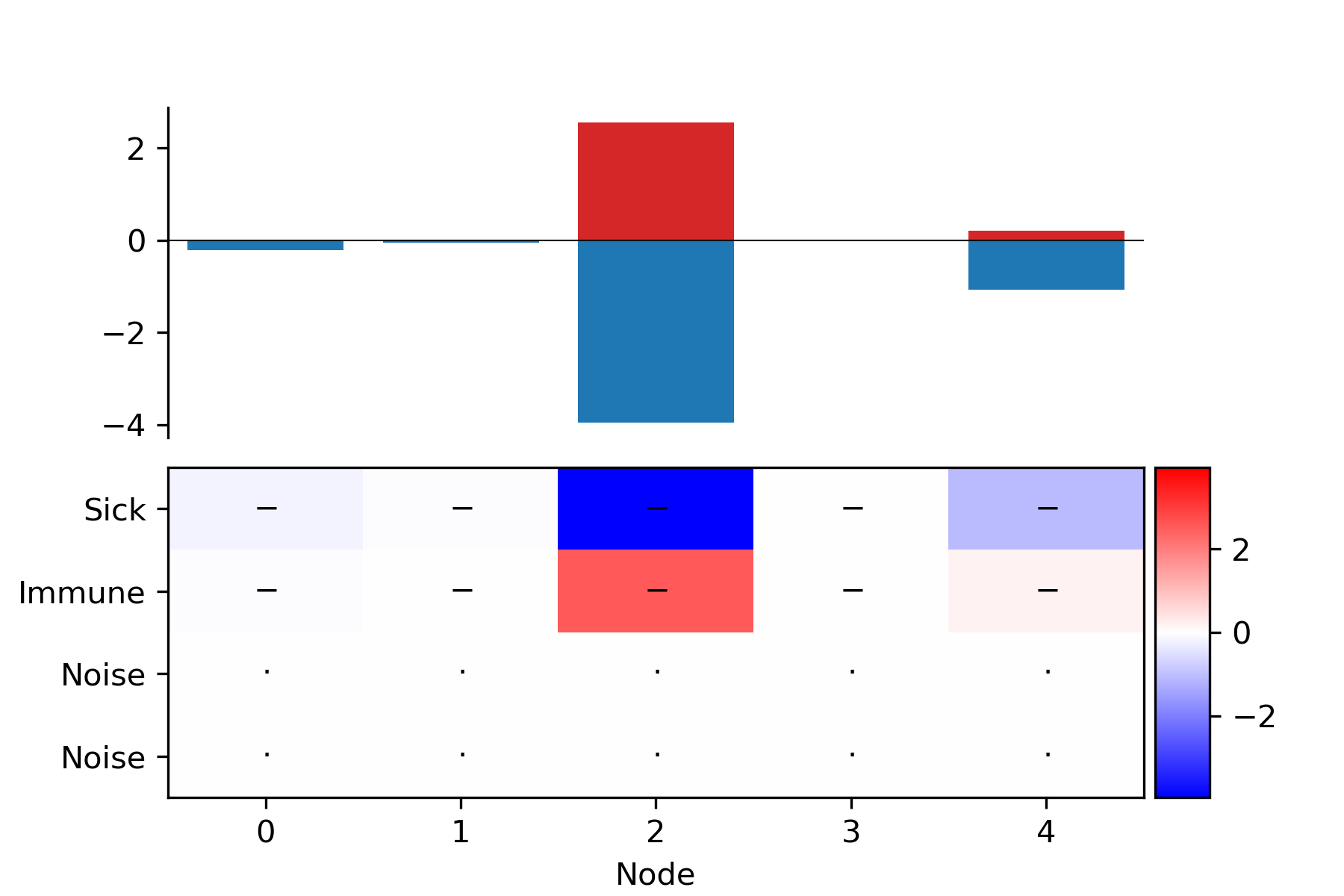}
\end{subfigure}%
\begin{subfigure}{0.44\linewidth}
    \centering
    \includegraphics[width=1.\linewidth]{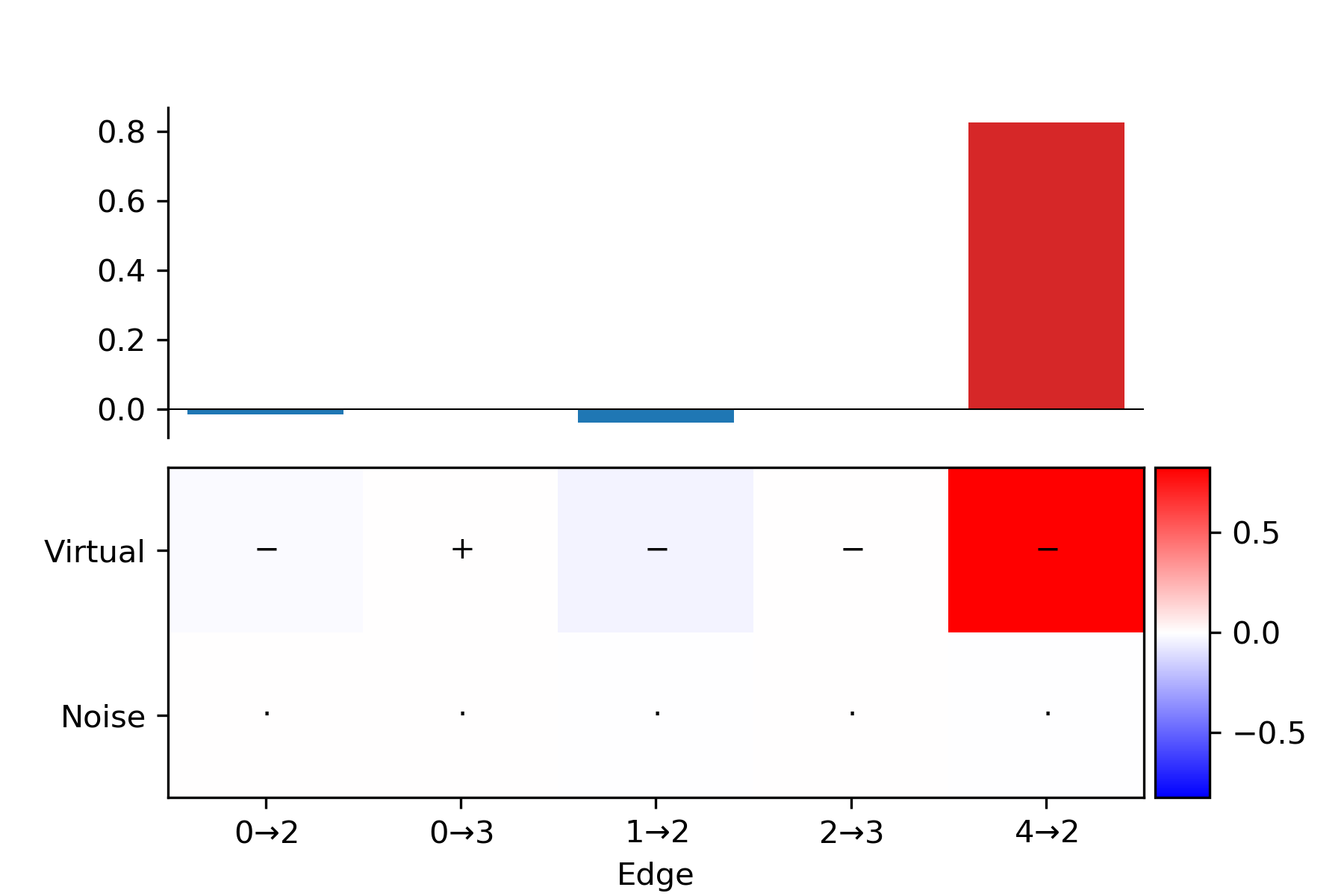}
\end{subfigure}%
\begin{subfigure}{0.11\linewidth}
    \includegraphics[width=1.\linewidth]{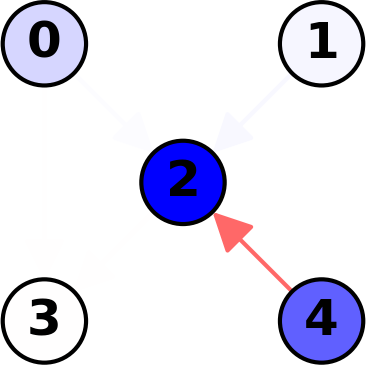}
    \caption*{LRP}
\end{subfigure}%
\caption{\textbf{Graph 3 - Max pooling.} All nodes are healthy, the explanation focuses on node 2, trying to explain why the network predicts it will remain healthy. SA assigns a positive gradient to the "sick" feature of node 2 itself, indicating that if the node was more sick the prediction would also shift towards mode positive values, i.e. sick. LRP decomposes most of the negative prediction onto the "sick" feature of node 2 itself, which is similar to SA, but more intuitive from a user perspective.}
\end{figure*}\clearpage

\begin{figure*}[h!]
\begin{subfigure}{.6\linewidth}
    \includegraphics[width=1.\linewidth]{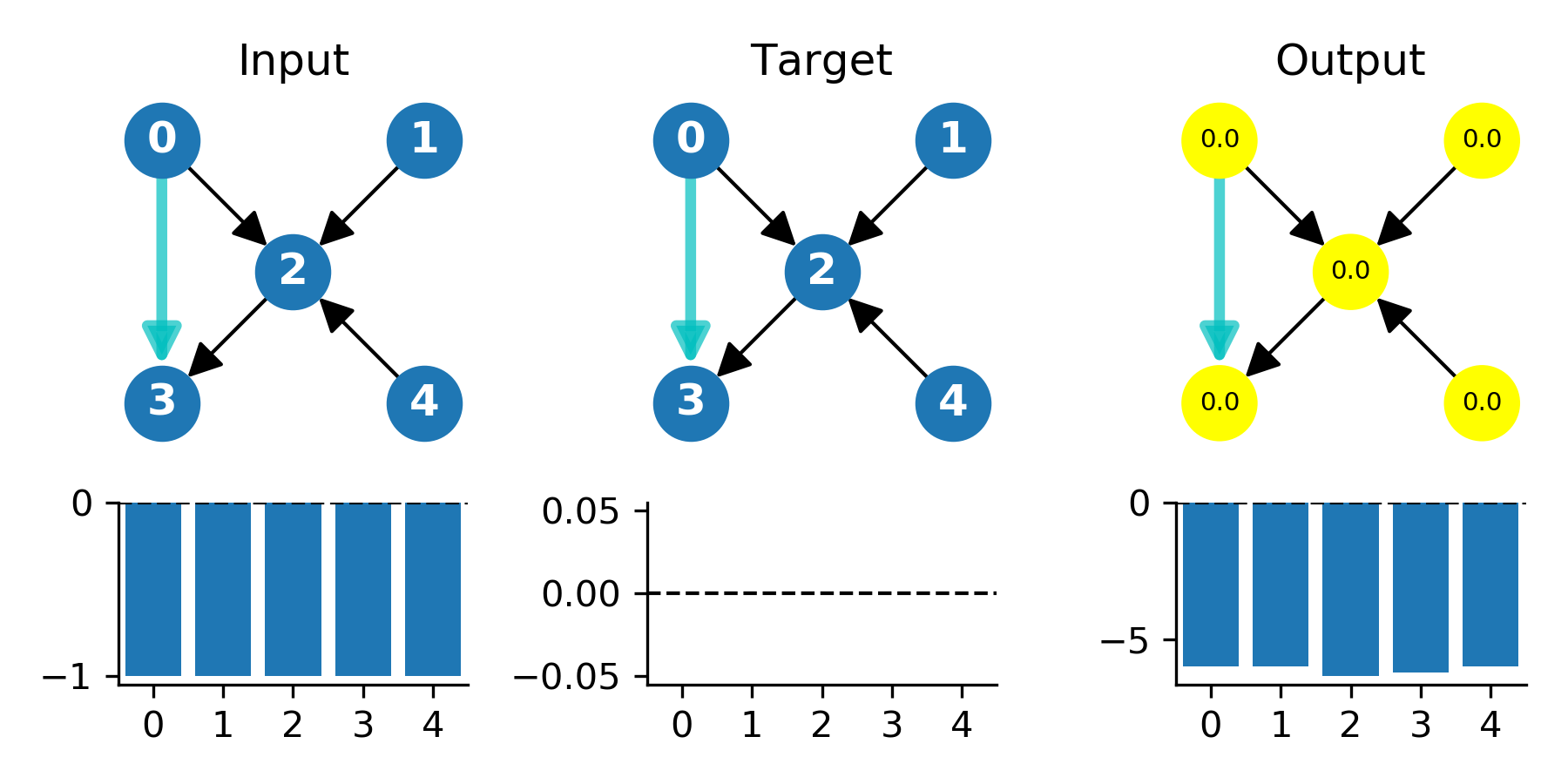}
\end{subfigure}\hfill\newline%
\begin{subfigure}{0.44\linewidth}
    \centering
    \includegraphics[width=1.\linewidth]{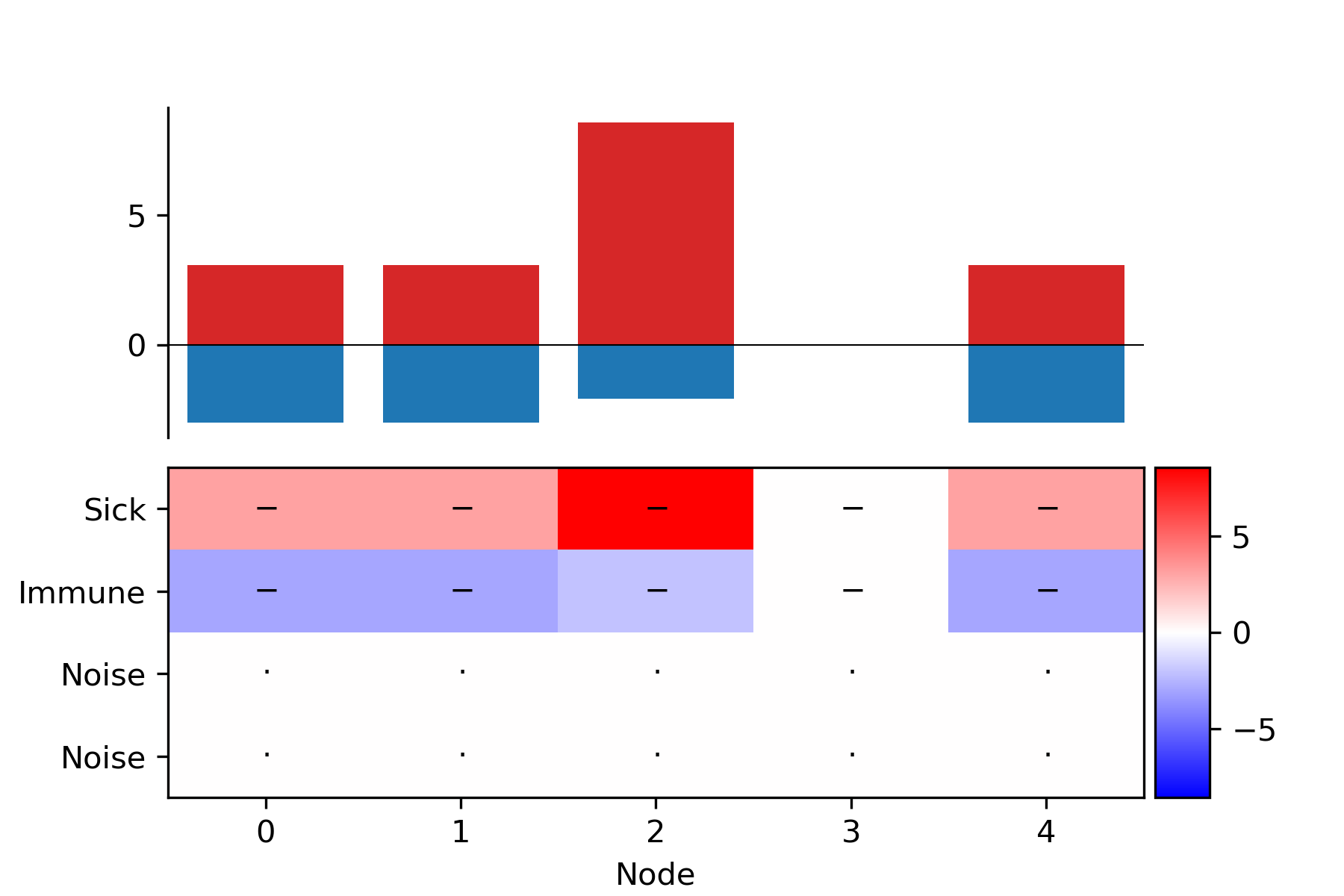}
\end{subfigure}%
\begin{subfigure}{0.44\linewidth}
    \centering
    \includegraphics[width=1.\linewidth]{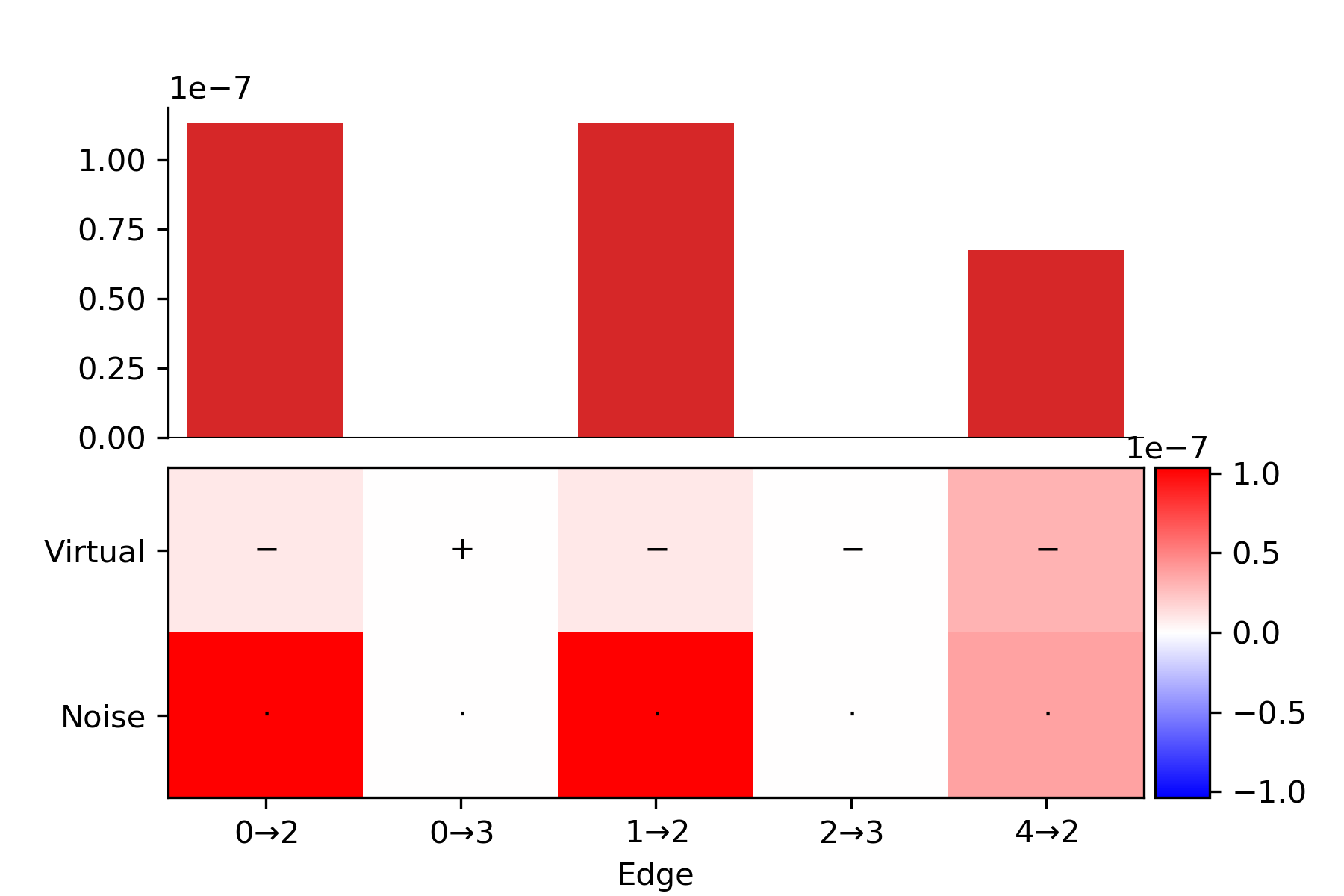}
\end{subfigure}%
\begin{subfigure}{0.11\linewidth}
    \includegraphics[width=1.\linewidth]{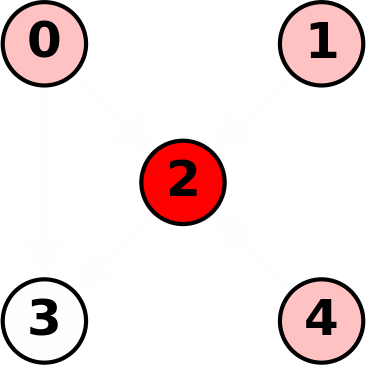}
    \caption*{SA}
\end{subfigure}\newline%
\begin{subfigure}{0.44\linewidth}
    \centering
    \includegraphics[width=1.\linewidth]{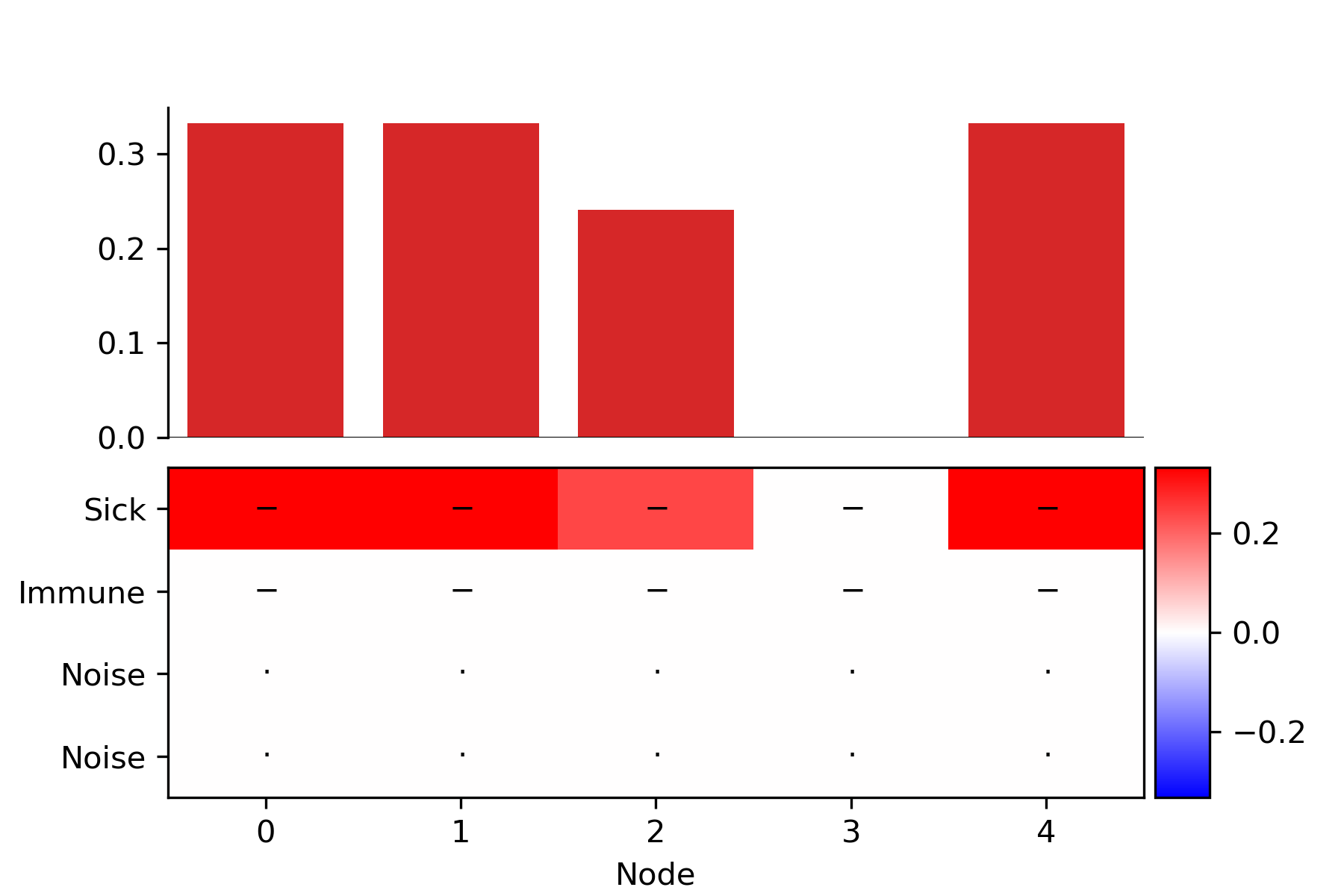}
\end{subfigure}%
\begin{subfigure}{0.44\linewidth}
    \centering
    \includegraphics[width=1.\linewidth]{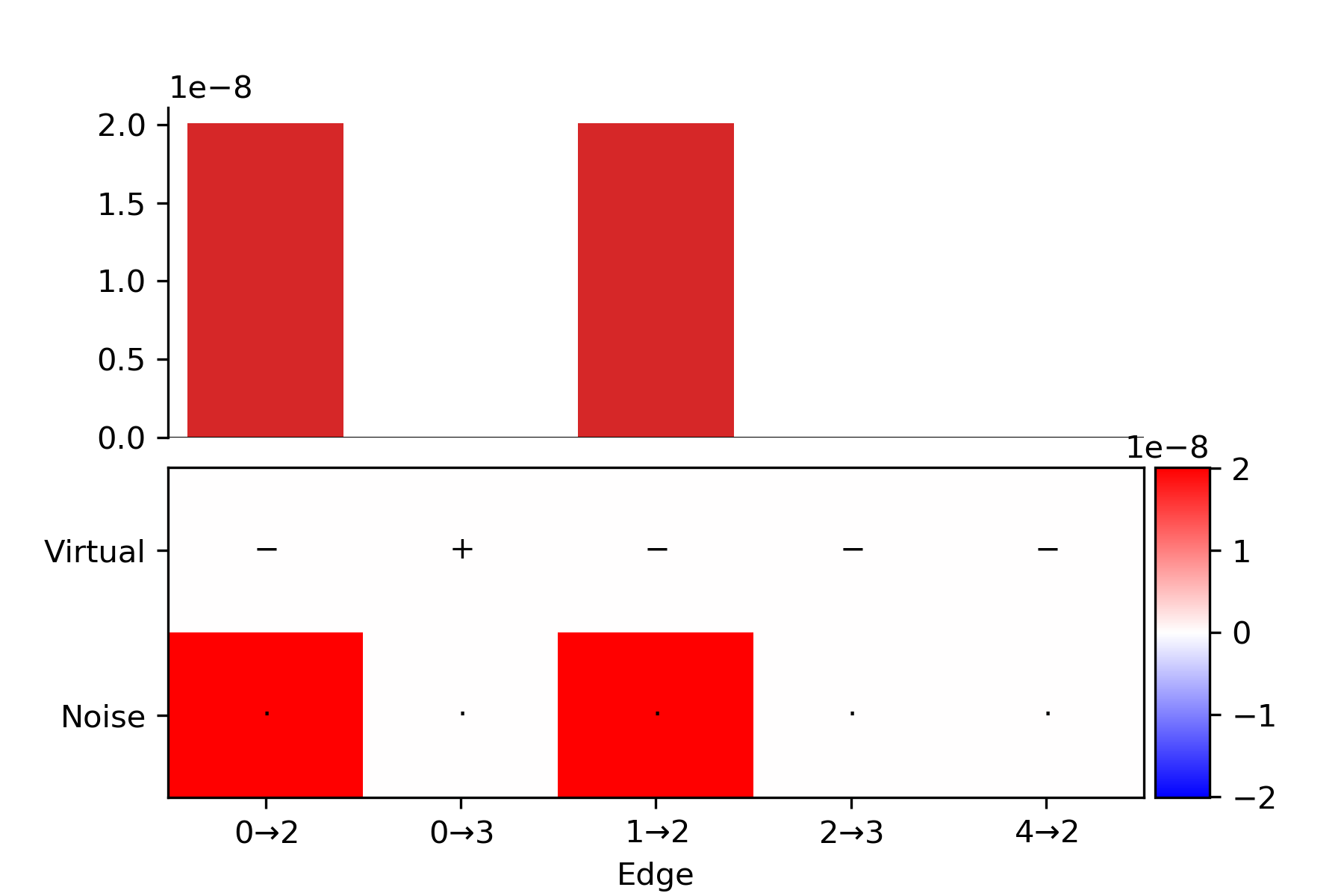}
\end{subfigure}%
\begin{subfigure}{0.11\linewidth}
    \includegraphics[width=1.\linewidth]{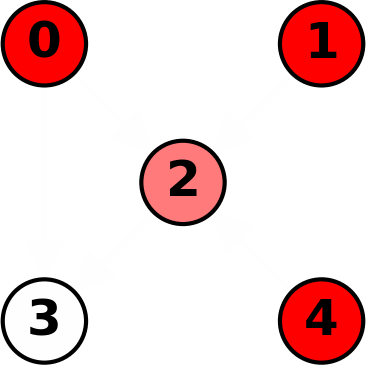}
    \caption*{GBP}
\end{subfigure}\newline%
\begin{subfigure}{0.44\linewidth}
    \centering
    \includegraphics[width=1.\linewidth]{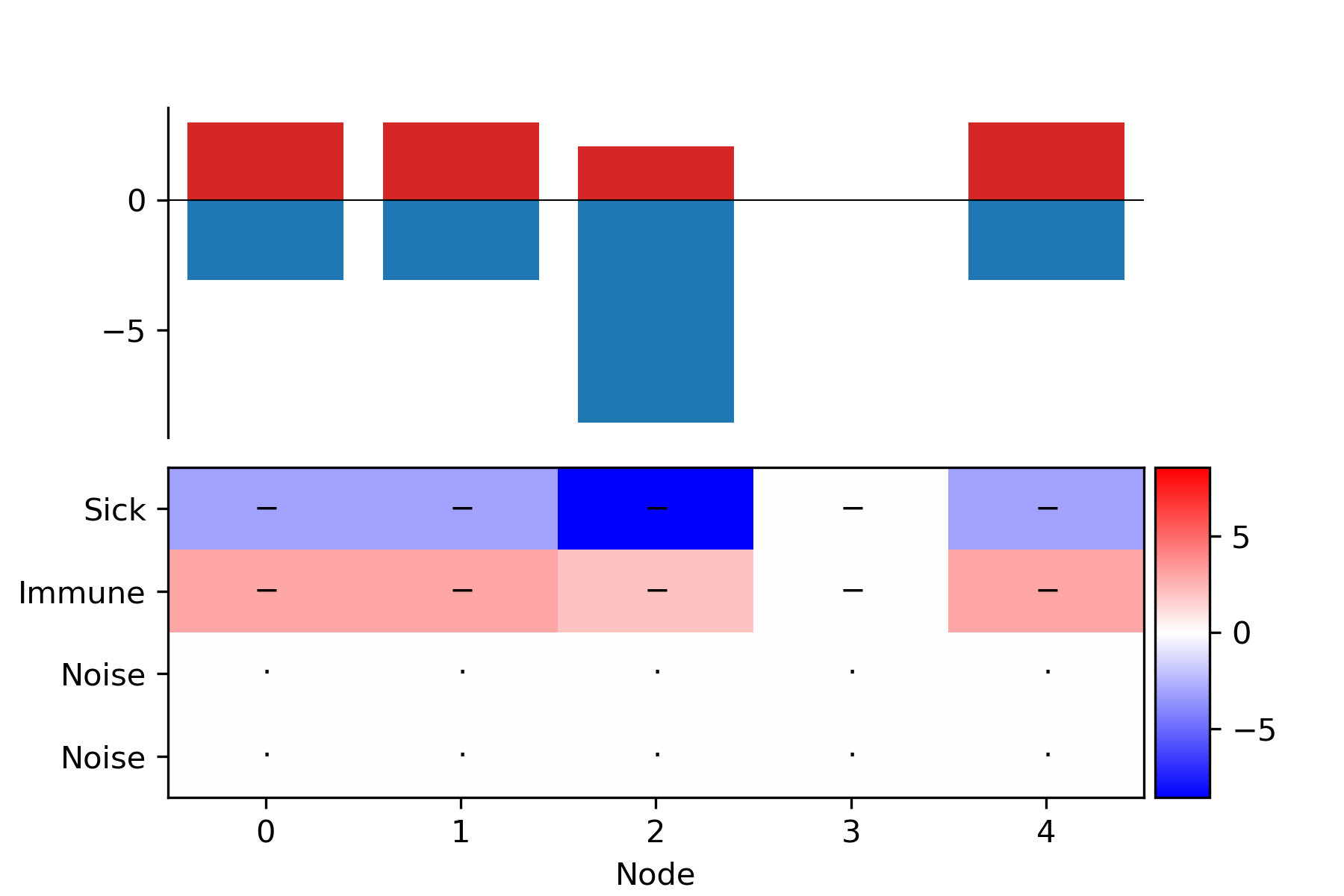}
\end{subfigure}%
\begin{subfigure}{0.44\linewidth}
    \centering
    \includegraphics[width=1.\linewidth]{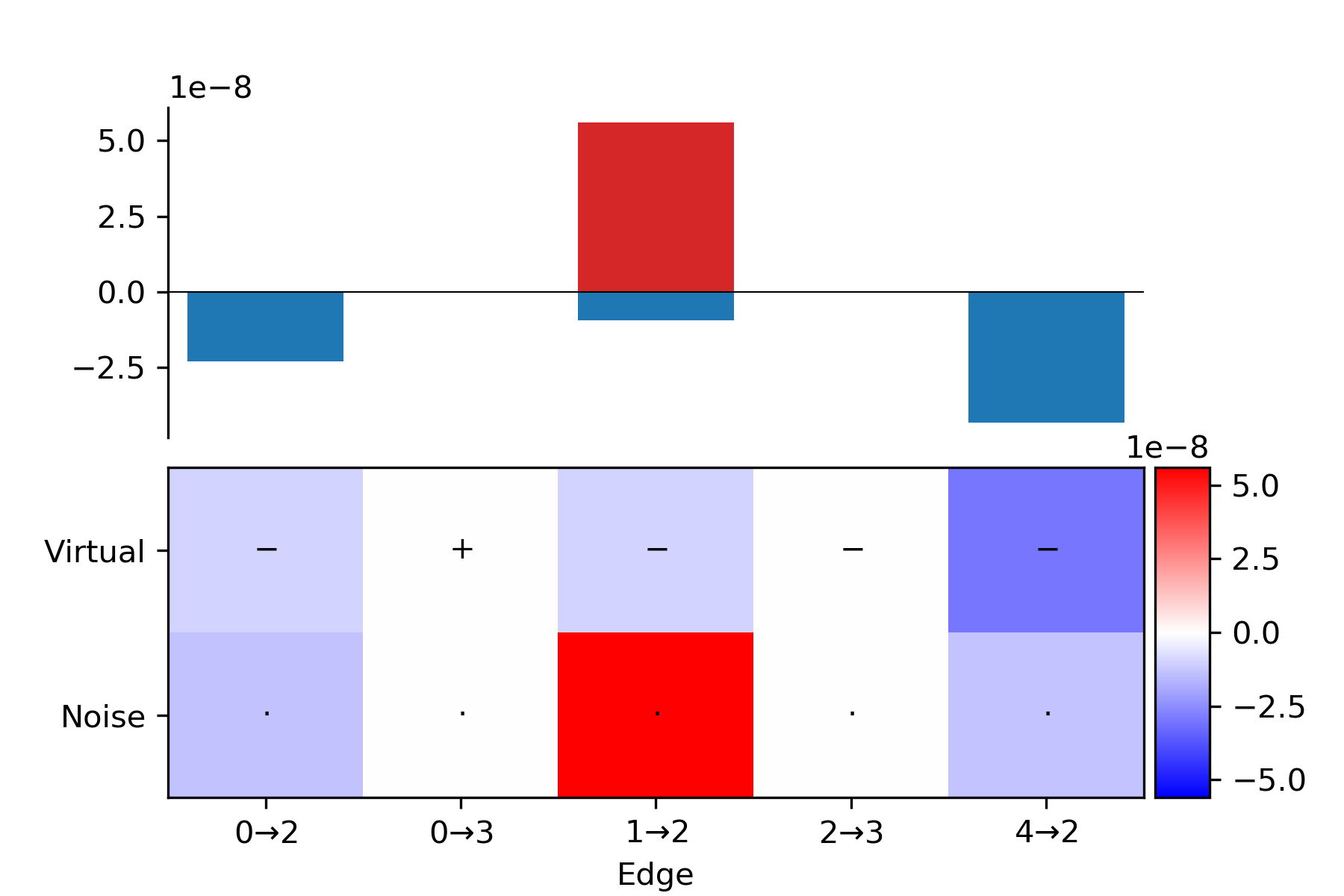}
\end{subfigure}%
\begin{subfigure}{0.11\linewidth}
    \includegraphics[width=1.\linewidth]{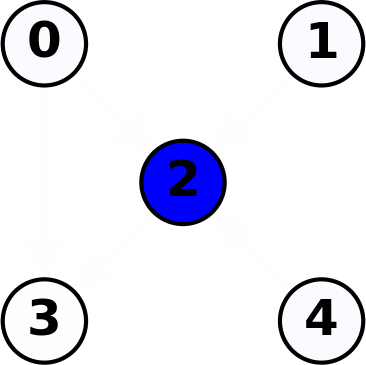}
    \caption*{LRP}
\end{subfigure}%
\caption{\textbf{Graph 3 - Sum pooling.} All nodes are healthy and the network outputs the correct prediction. The explanation focuses on node 2. Both SA and GBP focus on what would need to happen for the prediction to change, e.g. the surrounding nodes should be more sick and less immune for the central node to become infected. LRP instead decomposes the negative prediction into several negative contributions, the largest of which is attributed to the central node's feature of not being sick.}
\end{figure*}\clearpage

\begin{figure*}[h!]
\begin{subfigure}{.6\linewidth}
    \includegraphics[width=1.\linewidth]{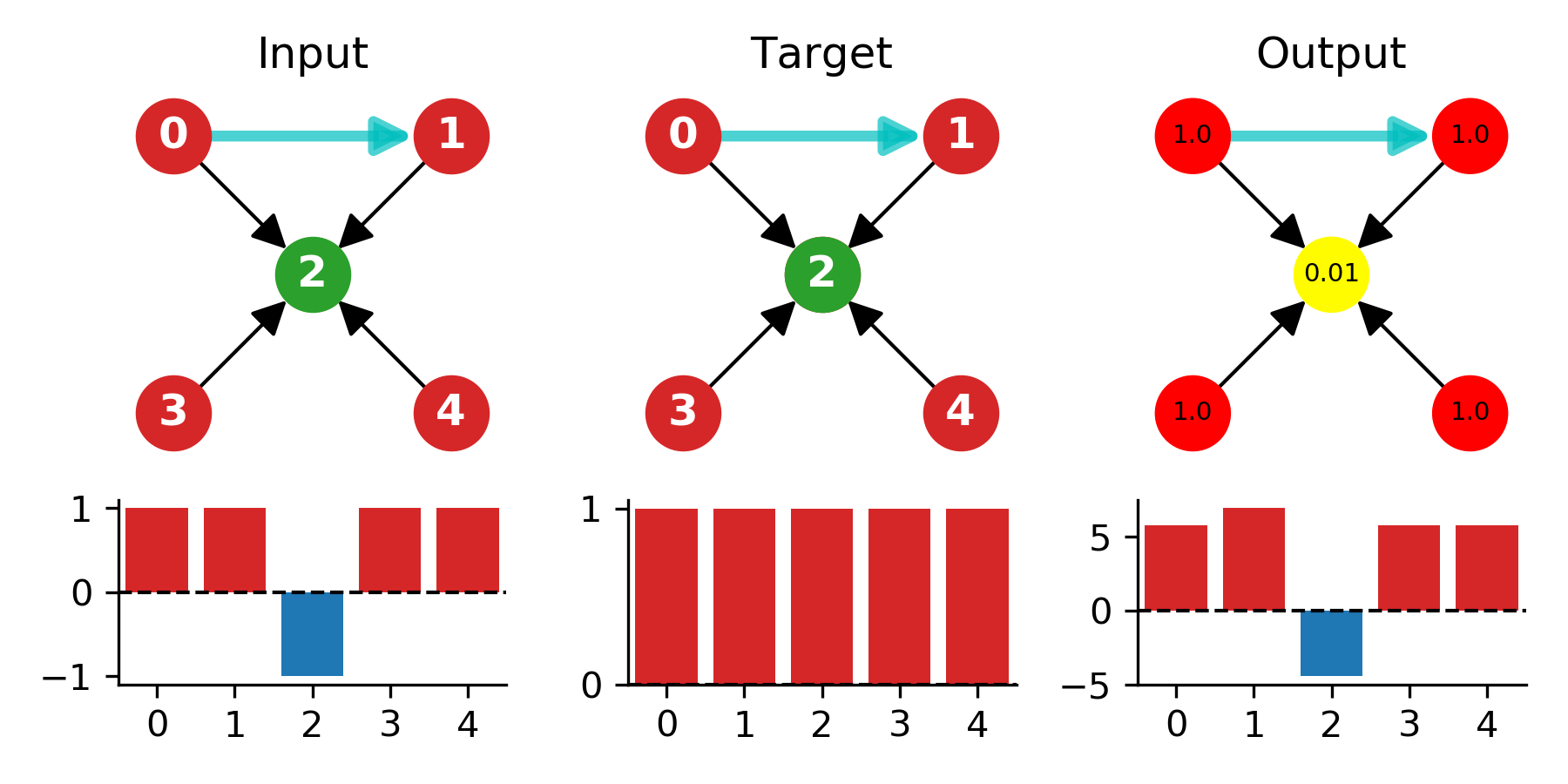}
\end{subfigure}\hfill\newline%
\begin{subfigure}{0.44\linewidth}
    \centering
    \includegraphics[width=1.\linewidth]{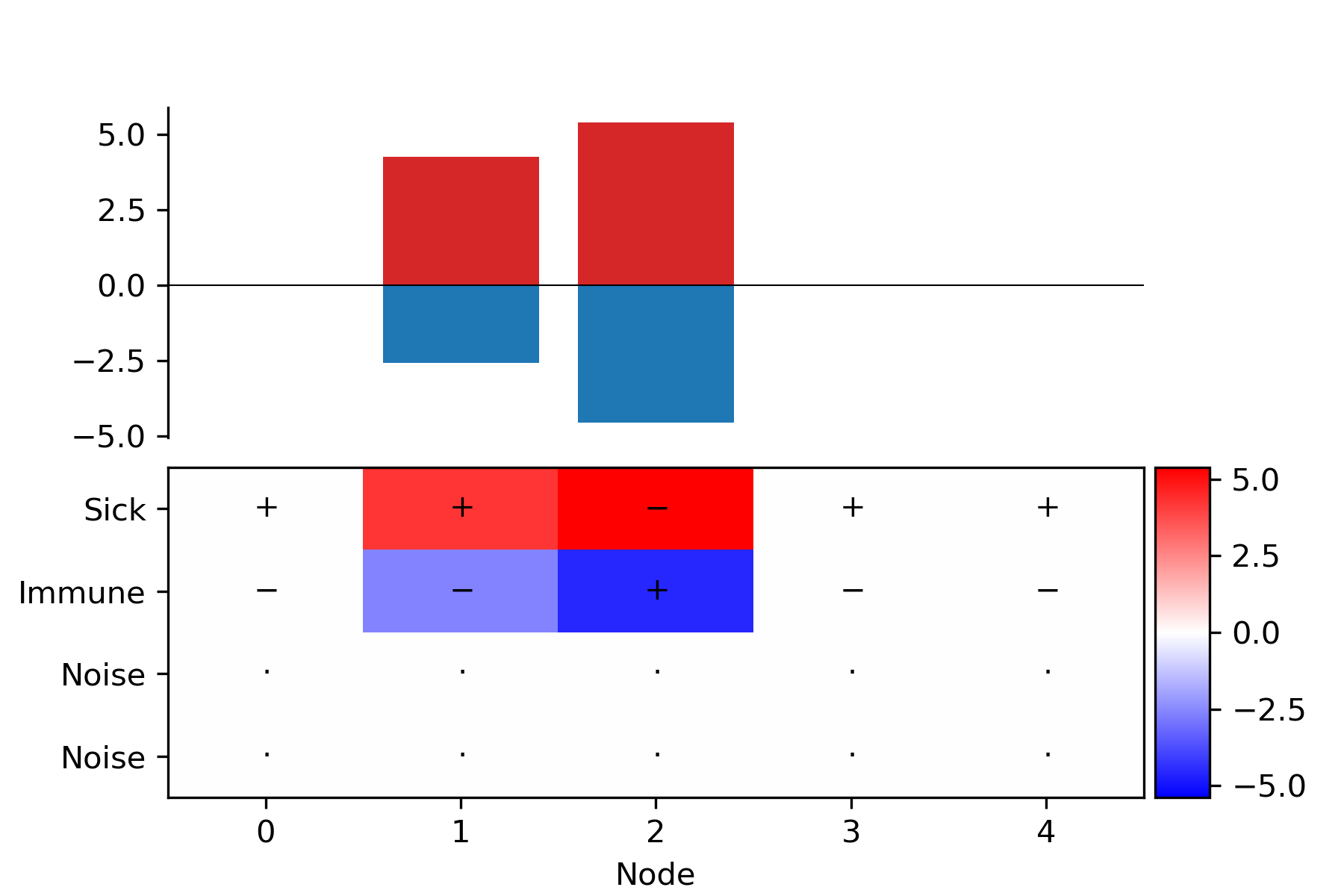}
\end{subfigure}%
\begin{subfigure}{0.44\linewidth}
    \centering
    \includegraphics[width=1.\linewidth]{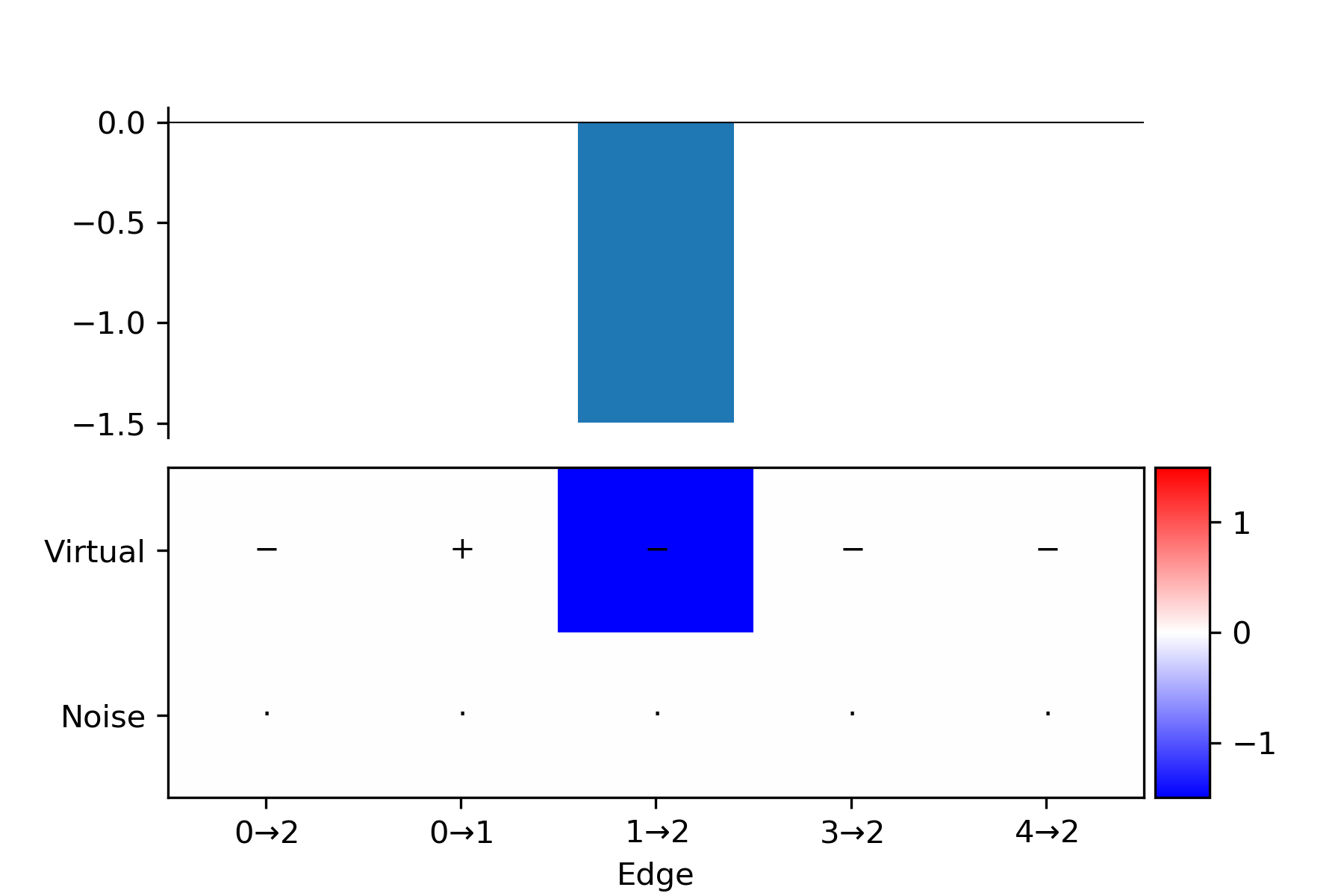}
\end{subfigure}%
\begin{subfigure}{0.11\linewidth}
    \includegraphics[width=1.\linewidth]{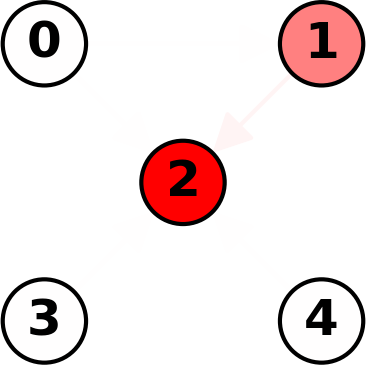}
    \caption*{SA}
\end{subfigure}\newline%
\begin{subfigure}{0.44\linewidth}
    \centering
    \includegraphics[width=1.\linewidth]{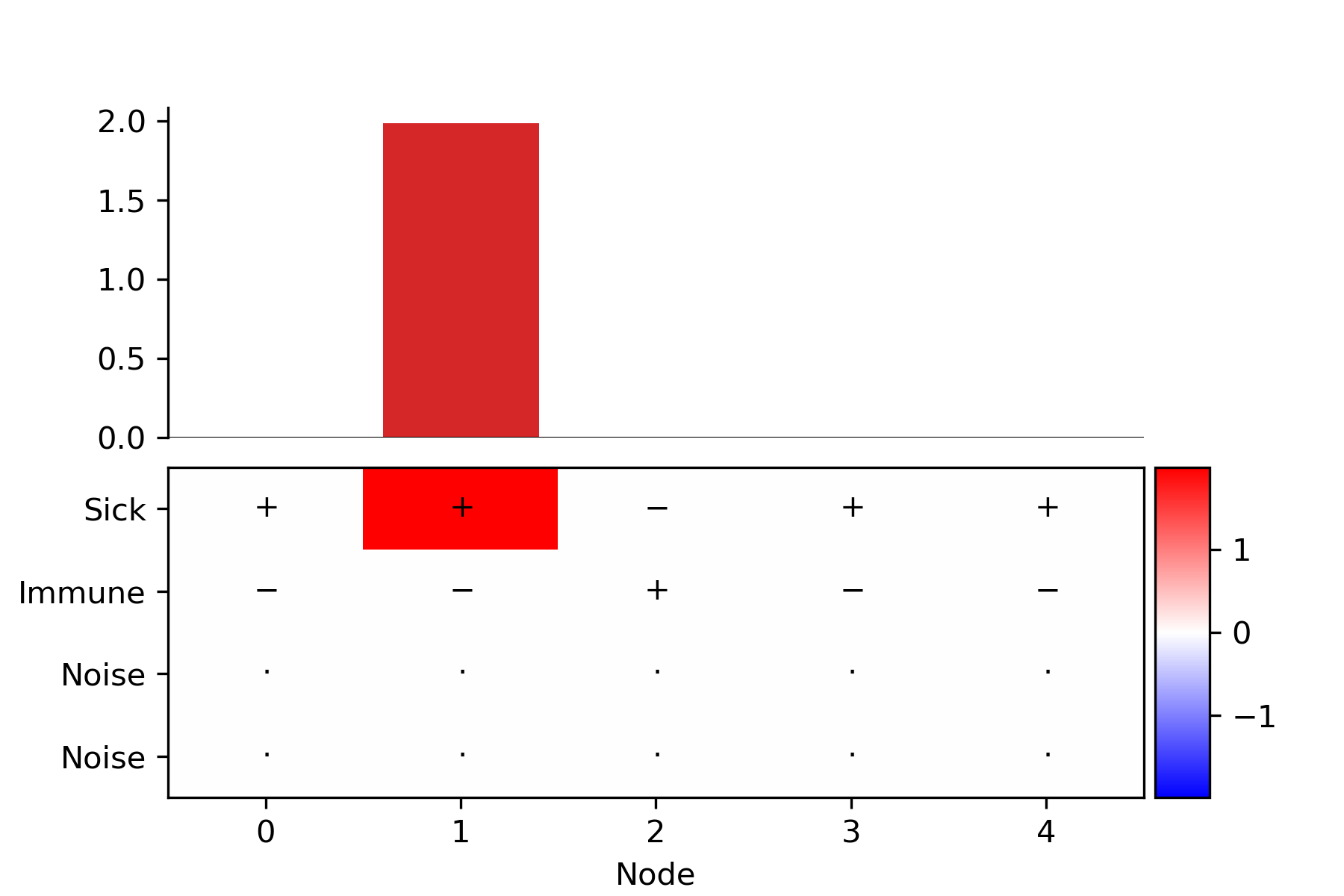}
\end{subfigure}%
\begin{subfigure}{0.44\linewidth}
    \centering
    \includegraphics[width=1.\linewidth]{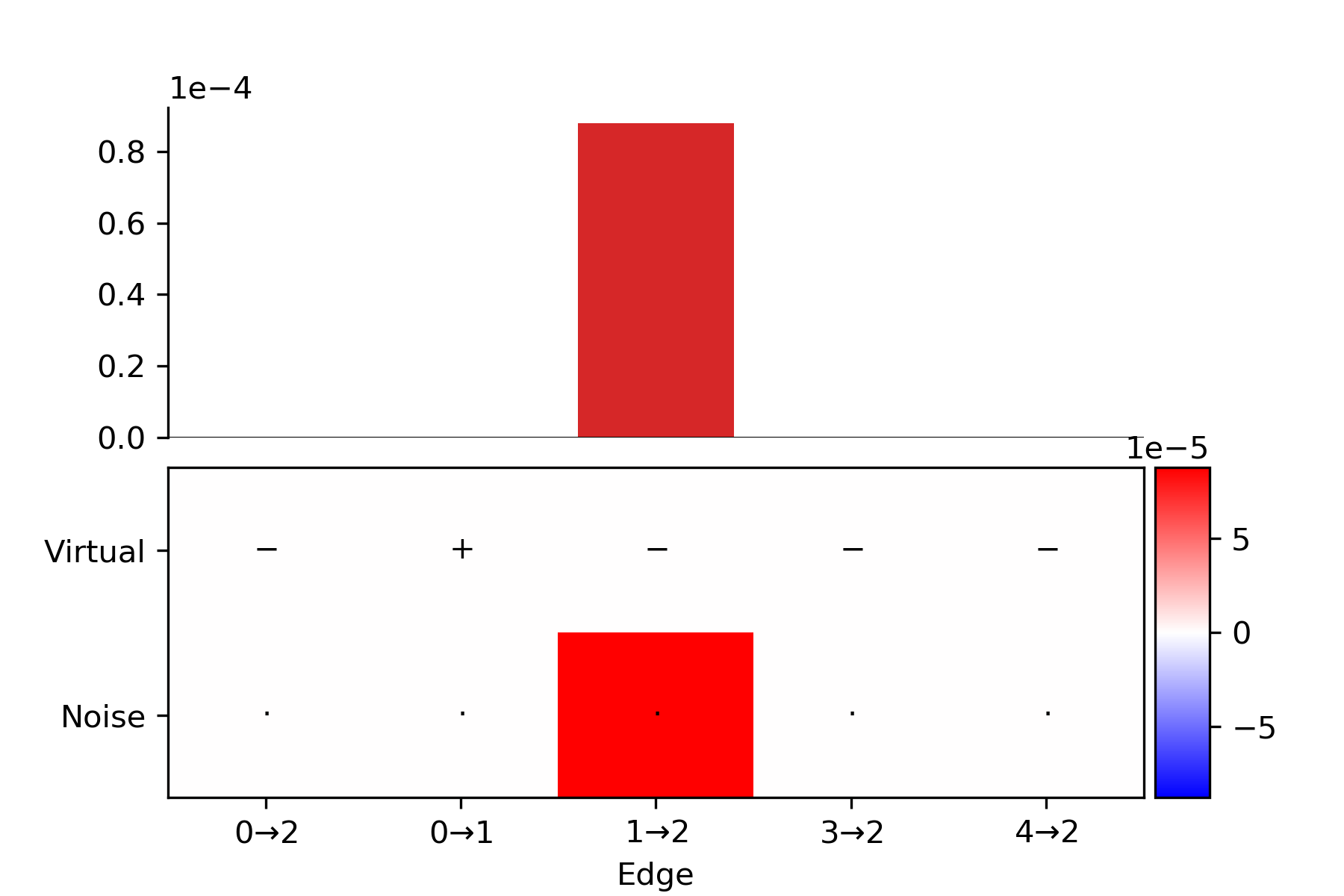}
\end{subfigure}%
\begin{subfigure}{0.11\linewidth}
    \includegraphics[width=1.\linewidth]{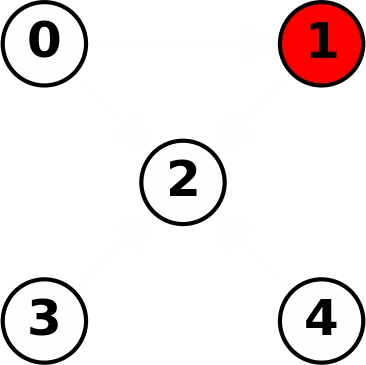}
    \caption*{GBP}
\end{subfigure}\newline%
\begin{subfigure}{0.44\linewidth}
    \centering
    \includegraphics[width=1.\linewidth]{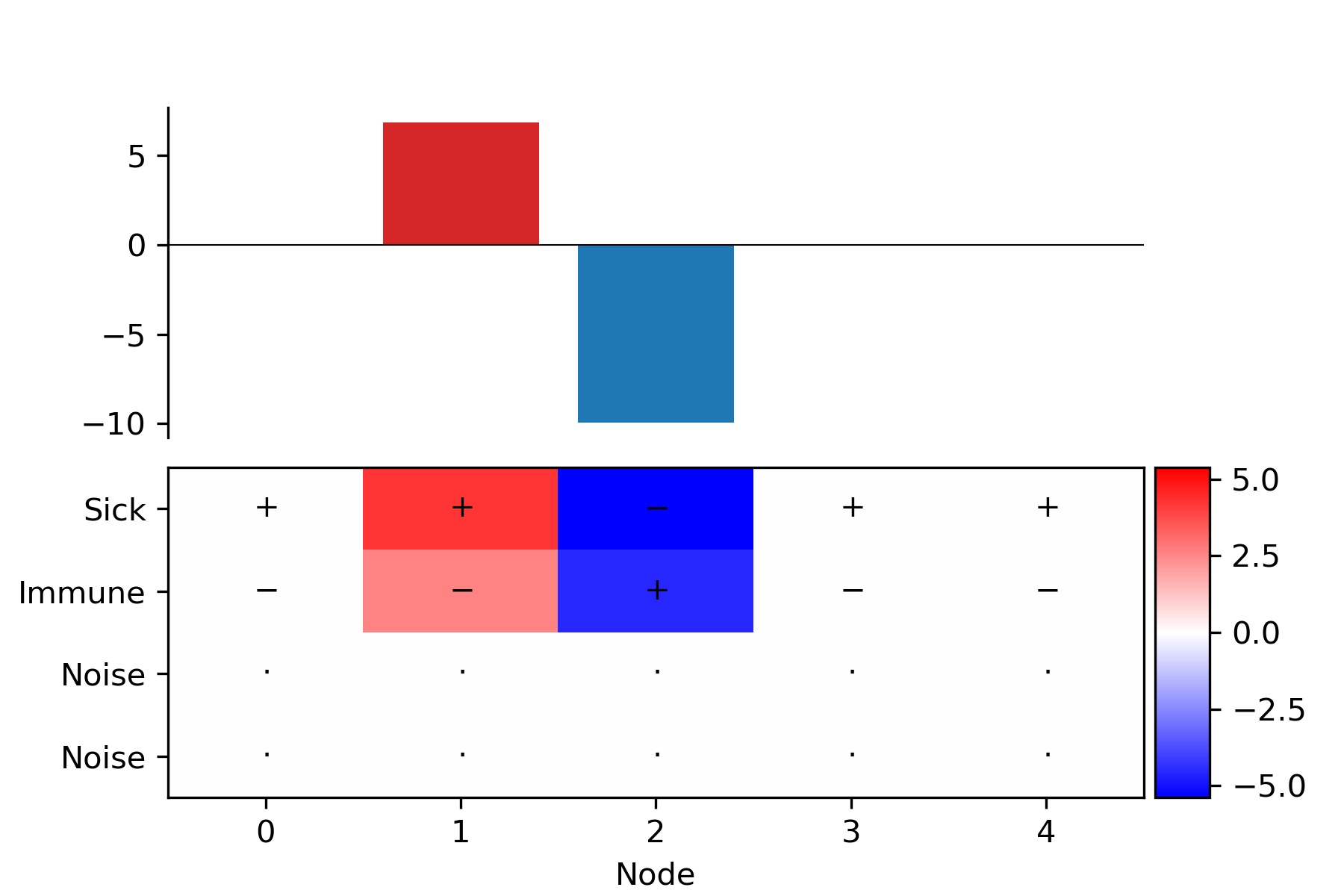}
\end{subfigure}%
\begin{subfigure}{0.44\linewidth}
    \centering
    \includegraphics[width=1.\linewidth]{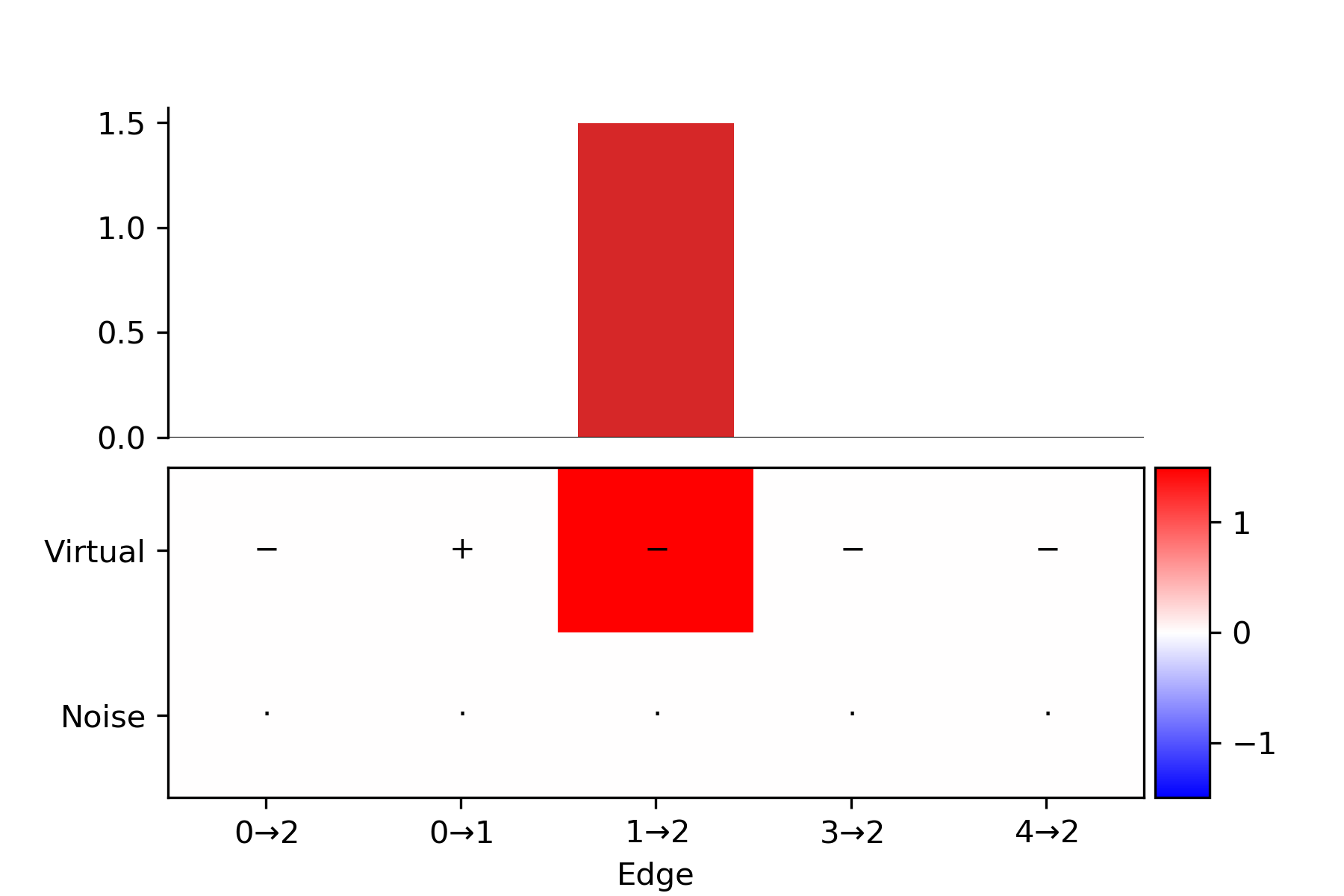}
\end{subfigure}%
\begin{subfigure}{0.11\linewidth}
    \includegraphics[width=1.\linewidth]{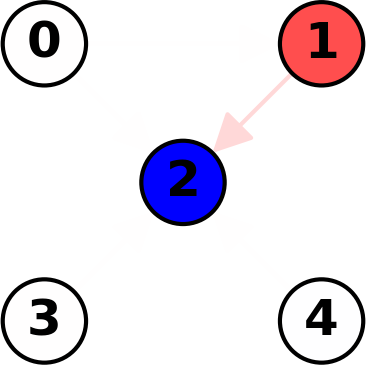}
    \caption*{LRP}
\end{subfigure}%
\caption{\textbf{Graph 4 - Max pooling.} Node 2 is immune and surrounded by sick nodes, the network is able to produce the correct answer, since max pooling is invariant w.r.t. the number of edges connected to node 2. Both SA and LRP identify the feature of being immune as important, while GBP does not consider it as its gradient would be negative.}
\end{figure*}\clearpage

\clearpage

\clearpage

\clearpage

\end{document}